\documentclass{article}

\usepackage[preprint]{neurips_2026}

\usepackage[utf8]{inputenc} %
\usepackage[T1]{fontenc}    %
\usepackage{hyperref}       %
\usepackage{url}            %
\usepackage{booktabs}       %
\usepackage{amsfonts}       %
\usepackage{nicefrac}       %
\usepackage{microtype}      %
\usepackage{xcolor}         %
\usepackage{bbm}
\usepackage{algorithm}
\usepackage{amsmath,amssymb,amsthm}
\usepackage[capitalize]{cleveref}
\usepackage{graphicx}
\usepackage{enumitem}
\usepackage{algpseudocode}
\usepackage{wrapfig}
\usepackage[dvipsnames]{xcolor}

\usepackage{tcolorbox}
\usepackage{subcaption}
\setlist[itemize]{leftmargin=0.5cm}

\newtheorem{theorem}{Theorem}[section]

\newtheorem{corollary}[theorem]{Corollary}
\newtheorem{definition}[theorem]{Definition}

\theoremstyle{remark}
\newtheorem{assumption}[theorem]{Assumption}
\newtheorem{remark}[theorem]{Remark}

\usepackage{adjustbox}
\usepackage{multirow}
\usepackage{siunitx}
\newcommand{\sd}[1]{{\color{black!70}\scriptstyle#1}}
\hypersetup{
  colorlinks=true,
  citecolor=MidnightBlue,
  linkcolor=MidnightBlue,
  urlcolor=MidnightBlue
}
\usepackage{wrapfig}

\title{Adaptive Ensemble Aggregation for Actor-Critics}

\author{%
  Nicklas~Werge \qquad Yi-Shan~Wu \qquad Manuel~Hau\ss mann\\
  \textbf{Bahareh~Tasdighi \qquad Melih~Kandemir} \\
  Department of Mathematics and Computer Science \\
  University of Southern Denmark \\
}

\begin{document}

\maketitle

\begin{abstract}
Ensembles are ubiquitous in off-policy actor-critic learning, yet their efficacy depends critically on how they are aggregated. Current methods typically rely on static rules or task-specific hyperparameters to balance overestimation bias and variance, leaving the challenge of a truly adaptive approach open. We introduce Adaptive Ensemble Aggregation (AEA), an algorithm that dynamically constructs ensemble-based targets for both critic and actor updates directly from training dynamics. We prove that AEA converges to a unique equilibrium where the aggregation parameter  minimizes value estimation error within a defined stability region. Theoretically, we establish that AEA achieves a shrinkage property where the estimation bias vanishes as the total ensemble size grows. Unlike subset-based methods like REDQ, which hit an information bottleneck determined by a fixed variance floor regardless of the ensemble size, AEA exploits the full ensemble to achieve optimal variance reduction—scaling inversely with the total number of models—and maximal Fisher information. Furthermore, we provide a formal guarantee for monotonic policy improvement under this adaptive regime. Extensive evaluations on various continuous control tasks demonstrate that AEA outperforms, on the majority of tasks, state-of-the-art baselines, providing a robust and self-calibrating framework for ensemble-based reinforcement learning.
\end{abstract}

\section{Introduction}

Off-policy reinforcement learning (RL) has become a powerful framework for continuous control, with deep actor-critic methods forming a central class~\citep{lillicrap2016continuous,haarnoja2017reinforcement, fujimoto2018addressing}. These methods jointly learn a \emph{critic} (the $Q$-function) that estimates expected returns and an \emph{actor} that optimizes its policy using the critic’s estimates~\citep{haarnoja2018soft}. A central difficulty is that policy improvement is conditioned on the reliability of the $Q$-value estimates; otherwise, the actor exploits estimation errors, and training becomes unstable. Off-policy bootstrapping with function approximation often leads to overestimation bias~\citep{thrun1993issues,fujimoto2018addressing,van2018deep}. To mitigate this, modern methods commonly use ensembles of critics and aggregate them conservatively to obtain more stable $Q$-value estimates~\citep{fujimoto2018addressing,haarnoja2018soft}.
Although ensembles can yield provably underestimating $Q$-value estimators~\citep{van2010double}, the gained stability comes at the expense of slower learning. Determining the optimum level of conservatism for the fastest adaptation remains a challenge. Existing methods rely on extensive manual tuning to balance critic stability and action exploration. 
Early methods hard-coded conservatism via min-clipping: TD3 \citep{fujimoto2018addressing} applied it only to the critic target while updating the actor through a single critic, and SAC \citep{haarnoja2018soft} extended it symmetrically to both references, while subsequent works refined this balance. For instance, some sought to encourage exploration by introducing optimism~\citep{ciosek2019better} or by using bandit-based online alternation between conservatism and optimism~\citep{moskovitz2021tactical}. Others refined critic conservatism by truncating quantiles~\citep{kuznetsov2020controlling} or by applying uncertainty-aware regularizers to penalize variance~\citep{cetin2023learning}. Most recently, \citet{tasdighi2025improving} demonstrated the utility of decoupling actor optimism from critic pessimism. Nonetheless, these approaches depend on fixed hyperparameters, e.g., truncation levels, regularization weights, and utility scales, that lead to instability across diverse learning tasks.

Searching for suitable critic ensemble aggregation hyperparameters is particularly challenging in high update-to-data (UTD) ratio regimes, since repeated use of the same transition data amplifies biases across multiple updates before corrective information becomes available~\citep{nikishin2022primacy}. This fact makes the rules that ensure stability in standard single-UTD and double-critic regimes overly conservative~\citep{chen2021randomized}. Conversely, less conservative choices tuned for sample-efficient learning can become unstable when ensembles are small or the update intensity changes~\citep{liang2022meanQ,wu2022aggressive,cetin2023learning}. These limitations point to four core questions:  
\begin{itemize}
\item Is it possible to develop an ensemble aggregation operator that can simultaneously minimize bootstrapping bias and sample variance?
\item Does this operator possess favorable estimator properties such as guaranteed convergence and information-theoretic optimality?
\item Can this operator's hyperparameters be learned on the fly, eliminating manual tuning?
\item Does the evolution of these hyperparameters reveal specific \emph{learning phases}, such as a transient need for high initial conservatism to counteract early-training gradient noise versus a shift toward actor optimism as the ensemble estimates stabilize?
\end{itemize}

We introduce Adaptive Ensemble Aggregation (AEA), which affirmatively answers these questions. We first formalize actor-critic learning using a decoupled ensemble framework with bootstrapping target and the policy reference modulated by independent, learnable parameters. These parameters are optimized on the fly via a scale-invariant objective that utilizes a robust pairwise-disagreement score based on the mean absolute difference of the ensemble. By employing a directional, fixed-magnitude update rule, the algorithm ensures numerical stability while directly neutralizing estimation bias according to the prevailing training dynamics.  AEA outperforms REDQ \citep{chen2021randomized}---the most competitive existing approach by a large margin---on the majority of 14 continuous control tasks from two benchmarking suites. We also demonstrate formally that AEA's estimator properties are superior to those of REDQ. The interpretable trajectories of these parameters expose the non-stationary nature of reinforcement learning: the critic rapidly settles into a conservative regime to anchor bootstrapping, while the actor maintains a neutral-to-optimistic stance that dynamically scales with the decreasing ensemble disagreement, automating the balance between stability and discovery.

\section{Continuous Control with Critic Ensembles}

We consider an infinite-horizon Markov Decision Process (MDP) defined by the tuple ${\mathcal{M} = \langle \mathcal{S}, \mathcal{A}, p, r, \gamma \rangle}$ \citep{puterman2014markov}, where $\mathcal{S}$ and $\mathcal{A}$ are continuous state and action spaces, and $\gamma \in (0,1)$ is the discount factor. The transition dynamics are governed by an unknown probability distribution $p(s' \mid s,a)$ over next states $s' \in \mathcal{S}$ given $(s,a) \in \mathcal{S} \times \mathcal{A}$, and the reward function $r:\mathcal{S} \times \mathcal{A} \rightarrow \mathbb{R}$ is assumed to be deterministic.\footnote{The deterministic reward assumption holds without loss of generality, as for every MDP with a probabilistic reward function that depends on the current state, current action, and next state, there exists a corresponding MDP with a deterministic reward function depending on the current state and current action \citep[Section 2.8]{albrecht2024marl}.} 
The agent's behavior is defined by a stochastic policy $\pi(\cdot \mid s)$, which maps a state $s \in \mathcal{S}$ to a probability density over actions $a \in \mathcal{A}$. At each time step $t \in \mathbb{N}$, the agent observes a state $s_t \in \mathcal{S}$, samples an action $a_t \sim \pi(\cdot \mid s_t)$, receives a reward $r(s_t, a_t)$, and transitions to a next state $s_{t+1} \sim p(\cdot \mid s_t, a_t)$. The agent seeks the policy distribution $\pi$ that maximizes the discounted sum of returns $J(\pi) = \mathbb{E}_{\pi} \left[\sum_{t=0}^{\infty} \gamma^t r(s_t,a_{t}) \right]$. The action-value function of a policy $\pi$ for a given $(s,a)$ is defined as $Q^\pi(s,a) = r(s,a) + \mathbb{E}_\pi\!\left[\sum_{t=1}^{\infty} \gamma^t r(s_t,a_{t}) \mid s,a\right]$. We denote the Bellman expectation operator as
$T^\pi Q (s,a) = r(s,a) + \gamma \mathbb{E}_{s' \sim p(\cdot \mid s,a), a' \sim \pi(\cdot \mid s')}[Q(s',a')]$, where $Q: \mathcal{S} \times \mathcal{A} \rightarrow \mathbb{R}$. Since this operator is a $\gamma$-contraction under the supremum norm, it has a unique fixed point at $Q^\pi$ by Banach's fixed point theorem, i.e., $Q^\pi(s,a) = T^\pi Q^\pi(s,a)$, $\forall (s,a)\in \mathcal{S} \times \mathcal{A}$. For any other $Q$, this equality does not hold, and a \emph{Bellman residual} occurs.

We consider model-free policy optimization for continuous control via function approximation in an actor-critic architecture. The critic $Q$ estimates the action value $Q^\pi$ (under $\pi$) by minimizing the Bellman residual~\citep{silver2014deterministic,bertsekas1996neuro,haarnoja2018soft}:
\begin{align} \label{eq:critic-training}
    \min_{Q} \mathbb{E}_{(s, a, r, s') \sim \mathcal{D}} \left[ ( Q(s,a) - y(s, a, r, s') )^2 \right],
\end{align}
using samples $(s,a,r,s')$ drawn from a replay buffer $\mathcal{D}$. The bootstrapped regression target $y(s,a,r,s')= r(s,a) + \gamma \overline{Q}(s', a')$ is a sample estimate of the Bellman target $T^\pi \overline{Q}(s,a)$, which is not accessible as it depends on the unknown $p(s'|s,a)$. The action for the next state is sampled from the current policy, $a'\sim \pi(\cdot|s')$. In this formulation, $\overline{Q}$ is a delayed copy of $Q$~\citep{mnih2015human}, referred to as the target critic. The actor is trained to maximize the expected action value:
\begin{align} \label{eq:actor-training}
\max_\pi \mathbb{E}_{s \sim \mathcal{D}, a \sim \pi(\cdot \vert s)} [ \widetilde{Q}(s,a) ]
\end{align}
where $\widetilde{Q}$ is an action-value estimate used for training the actor and may differ from both the critic $Q$ and the target critic $\overline{Q}$. In the idealized case of exact Bellman policy evaluation, $Q$, $\overline Q$, and $\widetilde Q$ all converge to the true value function $Q^\pi$~\citep{sutton2018reinforcement}.

Decoupling is already implicit in distributional reinforcement learning \citep{bellemare2023distributional}, where the full return distribution is learned by the critic, but only its mean (or a risk-sensitive functional) is used for policy optimization. Our framework makes this decoupling explicit and controllable through scalar parameters rather than through a distributional projection.
The coupling of the bootstrapped target $y$ with $Q$ introduces an overestimation bias \citep{thrun1993issues,van2010double,van2018deep,fujimoto2018addressing}, as the maximization (or policy optimization) over noisy value estimates leads to systematic positive bias, where $\mathbb{E}[\max_a \hat{Q}(s,a)] \geq \max_a \mathbb{E}[\hat{Q}(s,a)]$. In actor-critic settings, the policy $\pi$ overexploits regions where the function approximator erroneously predicts high values, leading to suboptimal convergence or divergence \citep{fujimoto2018addressing}. To mitigate these instabilities, modern architectures use a critic ensemble $\mathcal{Q} = \{Q_1, \dots, Q_N\}$, parameterized by independent weights $\{\theta^{(i)}\}_{i=1}^N$ \citep{osband2016deep, chen2021randomized}. Since overestimation can be controlled by taking the minimum over multiple estimates \citep{lan2020maxmin}, these methods maintain multiple estimates to characterize the distribution of the value function. For any state-action pair $(s,a)$, we define the ensemble mean $\mu_{\mathcal{Q}}(s,a) = \frac{1}{N} \sum_{i=1}^N Q_i(s,a)$ as a baseline estimator for the next-state value. The error of this estimator relative to $Q^\pi$ can be decomposed as:
\begin{align*}
\mu_{\mathcal{Q}}(s,a) - Q^\pi(s,a) = B(s,a) + \epsilon(s,a)
\end{align*}
where $B(s,a) = \mathbb{E}[\mu_{\mathcal{Q}}(s,a) - Q^\pi(s,a)]$ represents the systematic overestimation bias, while $\epsilon(s,a)$ denotes the zero-mean stochastic estimation noise—taken over the ensemble's random initialization and training noise. We follow standard analytical practice \citep{ciosek2019better} and assume $\epsilon$ to be $\sigma$-sub-Gaussian, satisfying $\mathbb{E}[\exp(\lambda \epsilon)] \leq \exp\left( \lambda^2 \sigma^2 / 2 \right)$ for all $\lambda \in \mathbb{R}$, consistent with the behavior of aggregated noise in wide neural ensembles \citep{jacot2018neural}.

\section{The Adaptive Ensemble Aggregation (AEA) Algorithm} \label{sec:aea}

Learning aggregation parameters in an online, off-policy, bootstrapped setup is inherently difficult due to numerical instabilities from the deadly triad \citep{van2018deep}. While USAC~\citep{tasdighi2025improving} circumvents these dynamics via exhaustive offline tuning, and OAC~\citep{ciosek2019better} relies on static heuristics, meta-learning approaches like TOP~\citep{moskovitz2021tactical} require expensive second-order gradients that introduce high variance. Furthermore, weighting schemes such as ADDQ~\citep{doering25a} exhibit inconsistent performance due to sensitivity to local bias in the ensemble members. Unlike existing ensemble methods that rely on a static min-pooling operator \citep{lan2020maxmin, chen2021randomized} or fixed quantile truncation \citep{kuznetsov2020controlling}, which can lead to pessimistic underexploration in continuous domains, AEA treats aggregation as a dynamic optimization problem. We propose a stable alternative that constructs the Bellman target through a learnable aggregation of the critic ensemble. We quantify ensemble disagreement using the normalized mean absolute difference across element pairs. By optimizing the aggregation objective, we derive a fixed-magnitude update rule whose direction is determined by a robust majority vote across the minibatch, ensuring stability even in high-noise learning regimes.
 
\textbf{The AEA estimator.} We construct the references $\overline Q_{\overline \kappa}$ and $\widetilde Q_{\kappa}$ by modulating the ensemble means with a scaled measure of ensemble disagreement:
\begin{align}
\overline{Q}_{\overline{\kappa}}(s,a) = \mu_{\overline{\mathcal{Q}}}(s,a) + \overline{\kappa} \cdot \overline{\delta}(s,a) \quad\text{and}\quad \widetilde{Q}_{\kappa}(s,a) = \mu_{\mathcal{Q}}(s,a) + \kappa \cdot \delta(s,a).
\end{align}
Here, $\overline{\kappa}$ and $\kappa$ are learnable parameters, while $\overline{\delta}$ and $\delta$ denote the ensemble disagreement for the target and active critics, respectively. The ensemble disagreement $\delta(s,a)$ as the average pairwise deviation among $Q$-value estimates:
$\delta(s,a) = \textstyle\binom{N}{2}^{-1} \textstyle\sum_{i>j} \lvert Q_i(s,a) - Q_j(s,a) \rvert.$
This score unbiasedly estimates the mean absolute difference (MAD)~\citep{yitzhaki2003gini}, which is more robust to outliers than the standard deviation via linear penalization. This makes $\overline\kappa$ and $\kappa$ interpretable: positive values indicate optimism, and negative values indicate conservatism relative to the ensemble mean.

\textbf{Stabilizing value propagation.} We introduce the normalized objective: 
\begin{align}
    \mathcal{L}_\text{crit}(\overline{\kappa}) =  \mathbb{E}_{\mathcal{D}} \left[ \lvert \widetilde{Q}_{\kappa}(s,a) - y_{\overline{\kappa}}(s,a,r,s') \rvert\big/{\overline{\delta}(s',a')} \right].
\end{align}
We update $\overline{\kappa}$ via gradient descent on this objective. Given that $\partial y_{\overline{\kappa}}/\partial \overline{\kappa} = \gamma \overline{\delta}(s',a')$, the gradient with respect to $\overline{\kappa}$ simplifies as follows:
\begin{align}
\nabla_{\overline{\kappa}} \mathcal{L}_\text{crit}(\overline{\kappa}) = \mathbb{E}_{\mathcal{D}} \left[ \frac{-\text{sign}(\widetilde{Q}_{\kappa}(s,a) - y_{\overline{\kappa}})}{\overline{\delta}(s',a')} \cdot \frac{\partial y_{\overline{\kappa}}}{\partial \overline{\kappa}} \right] \nonumber \
= -\gamma \mathbb{E}_{\mathcal{D}} \left[ \text{sign} \left( \widetilde{Q}_{\kappa}(s,a) - y_{\overline{\kappa}} \right) \right].
\end{align}
This result confirms that the update direction is a scale-invariant majority vote, depending solely on whether the current target $y_{\overline{\kappa}}$ overestimates or underestimates the active reference $\widetilde{Q}_{\kappa}$ across the minibatch. Such a mechanism ensures the numerical stability required for online tuning of scalar parameters in a bootstrapped off-policy setting.

\textbf{Decoupling exploration landscape.} Since $\widetilde{Q}_{\kappa}$ in Eq.~\ref{eq:actor-training} shapes the actor's objective, we independently optimize the policy reference to reconcile stable value propagation with effective discovery. We update the actor parameter $\kappa$ by minimizing:
\begin{align} \label{eq:kappa}
\min_{\kappa} \mathcal{L}_\text{act}(\kappa) = \min_{\kappa} \mathbb{E}_{\mathcal{D}} \left[ {\lvert \widetilde{Q}_{\kappa}(s,a) - y_{\overline{\kappa}}(s,a,r,s') \rvert}\big/{\delta(s,a)} \right],
\end{align}
yielding the gradient $\nabla_{\kappa} \mathcal{L}_\text{act}(\kappa) = \mathbb{E}_{\mathcal{D}} [ \text{sign} ( \widetilde{Q}_{\kappa} - y_{\overline{\kappa}} ) ]$. Unlike the $\overline{\kappa}$ update, the absence of the negative sign and the $\gamma$ factor confirms that $\kappa$ moves in an opposing functional direction: whereas $\overline{\kappa}$ anchors the target to the current critics, $\kappa$ pulls the actor reference toward the stable target. These scalars explicitly modulate the degree of optimism and conservatism on the fly.

\textbf{Implementation.} We train each critic $Q_i$ in the ensemble according to Eq.~\ref{eq:critic-training} using the modified regression target $y_{\overline{\kappa}}(s,a,r,s') = r(s,a) + \gamma \overline{Q}_{\overline{\kappa}}(s',a')$. Similarly, for policy optimization, we replace the generic $\widetilde Q$ in Eq.~\ref{eq:actor-training} with our modulated reference $\widetilde{Q}_{\kappa}$. While AEA is generic to any actor-critic architecture, we demonstrate its efficacy within the maximum-entropy framework \citep{ziebart2010modeling}, following soft actor-critic (SAC) best practices \citep{haarnoja2018soft} and automatically tuning the temperature $\alpha$ \citep{haarnoja2018softa}. 
We name our algorithm \emph{Adaptive Ensemble Aggregation (AEA)}. The comprehensive update cycle of our algorithm is detailed in Alg.~\ref{alg:AEA} within the Appendix.

\textbf{Computational complexity.} AEA introduces negligible computational overhead compared to baseline ensemble architectures such as REDQ. The computational cost is dominated by the forward and backward passes of the $N$ critic networks and the actor network. If $K$ denotes the FLOPs required for a single network pass, the ensemble cost scales linearly with $N$. In contrast, computing the ensemble disagreement $\delta(s,a)$ requires $O(N^2)$ scalar operations to evaluate the average pairwise deviation. For typical ensemble sizes ($N=10$), this involves only $\binom{N}{2} = 45$ absolute differences per sample—a negligible cost compared to the millions of FLOPs ($K$) required for neural function approximation. Furthermore, AEA utilizes a two-timescale update logic to maintain efficiency: while the critics are updated $G$ times per environment interaction (matching the UTD ratio), the policy, ensemble parameters $\overline{\kappa}$ and $\kappa$ are updated only once per interaction cycle. This decoupling ensures that the overhead for learning the aggregation scalars remains a fraction of the total training time.

\section{Theoretical Analysis}\label{sec:theory}

We establish the convergence and statistical optimality of AEA by mapping the ensemble dynamics to the \emph{Neural Tangent Kernel (NTK)} framework \citep{jacot2018neural}. Each critic $Q_i$ is a neural network with parameters $\theta^{(i)} \in \mathbb{R}^p$ and Jacobian $\mathbf{J}_\theta(s,a) = \nabla_\theta Q_\theta(s,a)$. In the infinite-width limit, the training dynamics enter a \emph{lazy training} regime \citep{chizat2019lazy}, where the Jacobian remains constant at its initial value, $\mathbf{J}_\theta \approx \mathbf{J}_{\theta_0}$, throughout optimization \citep{lee2019wide}. This constancy induces a positive semidefinite kernel $\mathbf{H}(x, x') = \langle \mathbf{J}_{\theta_0}(x), \mathbf{J}_{\theta_0}(x') \rangle$. Under this regime, the function space $\mathcal{F}$ behaves as a linear approximation around the initialization $\theta_0$:
${\mathcal{F} = \{ Q_{\theta_0} + \mathbf{J}_{\theta_0}(\theta - \theta_0) : \theta \in \mathbb{R}^p \}}$. 
Let $d^\pi$ be the state-action occupancy measure for policy $\pi$. We define $\Pi_{\mathcal{D}}$ as the orthogonal projection of an arbitrary function $\Phi \in \mathcal{L}^2(d^\pi)$ onto the tangent space $\mathcal{F}$ at $\theta_0$:
\begin{align*}
\Pi_{\mathcal{D}} \Phi = \arg\min_{f \in \mathcal{F}} \mathbb{E}_{(s,a) \sim d^\pi} [ (f(s,a) - \Phi(s,a))^2 ].
\end{align*}
This projection characterizes the unique element in $\mathcal{F}$ that minimizes the approximation error under the empirical distribution $d^\pi$.
For the ensemble $\mathcal{Q} \in \mathcal{F}^N$, we define the AEA-enhanced Bellman expectation operator ${\mathcal T_{\overline{\kappa}}^\pi : \mathcal{F}^N \to \mathcal{F}^N}$ as follows:
\begin{align*}
(\mathcal T_{\overline{\kappa}}^\pi\mathcal{Q})_i(s,a) \triangleq r(s,a) + \gamma \mathbb{E}_{s', a' \sim \pi} \left[ \mu_{\overline{\mathcal{Q}}}(s', a') + \overline{\kappa} \cdot \overline{\delta}(s', a') \right].
\end{align*}
This operator defines the target for each ensemble element. The Lipschitz constant $\text{Lip}(\delta)$ of the disagreement operator relative to the ensemble $\mathcal{Q}$ is the smallest $L \geq 0$ such that, for any two ensembles $\mathcal{Q}, \mathcal{Q}' \in \mathcal{F}^N$:
$|\delta(s,a) - \delta'(s,a)| \leq L \cdot \max_{i} |Q_i(s,a) - Q'_i(s,a)|$, 
where $\delta'$ denotes the disagreement of $\mathcal{Q}'$. This ensures that the disagreement function is stable to perturbations in individual $Q$-functions. The AEA learning process corresponds to a stochastic approximation of the projected Bellman update, wherein the ensemble value functions $\mathcal{Q}$ track the fixed point of $\Pi_{\mathcal{D}}T_{\overline{\kappa}}^\pi$.

\begin{theorem}\label{thm:aea_convergence}
Let $\{\theta_t^{(i)}\}_{i=1}^N$ be the parameters of an $N$-ensemble of neural networks in the NTK regime, and $\overline{\kappa}_t$ a learnable aggregation parameter. Define a two-timescale stochastic approximation:
\begin{align*}
\theta_{t+1}^{(i)} &= \theta_t^{(i)} + \eta_t \mathbf{J}_{\theta_t^{(i)}}^\top [ y_{\overline{\kappa}_t} - Q_{\theta_t^{(i)}} ], \hspace{1em}
\overline{\kappa}_{t+1} = \overline{\kappa}_t + \zeta_t \cdot \gamma \cdot \mathrm{sign} ( \widetilde{Q}_{\kappa}(s_t, a_t) - y_{\overline{\kappa}_t}(s_t, a_t, r_t, s'_t) ). \end{align*} Assume $d^\pi$ has full support and the step sizes satisfy the Robbins-Monro conditions\footnote{While the Robbins-Monro conditions require decaying step sizes, in practice, a small fixed learning rate suffices. The process then converges to a stationary distribution around the unique equilibrium $(\mathcal{Q}^*, \overline{\kappa}^*)$ \citep{kushner2003stochastic}.} with $\lim_{t \to \infty} \frac{\zeta_t}{\eta_t} = 0$. If $\overline{\kappa}_t$ is projected onto the stability region $\mathcal{K}_{\text{stable}} = \{ \overline{\kappa} : |\overline{\kappa}| < \frac{1 - \gamma}{\gamma \cdot \text{Lip}(\overline\delta)} \}$, then the joint process $(\mathcal{Q}_t, \overline{\kappa}_t)$ converges almost surely to a unique equilibrium $(\mathcal{Q}^*, \overline{\kappa}^*)$. At this point, $\mathcal{Q}^*$ is the fixed point of $\Pi_{\mathcal{D}} \mathcal{T}_{\overline{\kappa}^*}$, and $\overline{\kappa}^*$ minimizes the expected normalized absolute Bellman error.
\end{theorem}

\begin{assumption}[Conditional symmetry and ancillarity of the disagreement]\label{assump:symmetry}
Let
\begin{equation}\label{eq:Ndef}
N(s,a) \;\triangleq\; \widetilde{Q}_{\kappa}(s,a) - r(s,a) - \gamma\,\mu_{\overline{\mathcal{Q}}}(s',a'),
\qquad
u(s,a) \;\triangleq\; \frac{N(s,a)}{\overline{\delta}(s',a')},
\end{equation}\label{eq:condsym}
with randomness taken over $(s,a,r,s')\sim\mathcal{D}$ and ensemble noise. We assume:
    (i) \emph{Conditional symmetry.} Conditional on $\overline{\delta}(s',a')$, the random variable $N(s,a)$ is symmetric about its conditional mean;
    (ii) \emph{Ancillarity of the disagreement.} The pairwise-disagreement statistic $\overline{\delta}(s',a')$ is ancillary for the location of $N(s,a)$, i.e.,
    \begin{equation}\label{eq:basu_decouple}
    \mathbb{E}\bigl[N(s,a)\,\big|\,\overline{\delta}(s',a')\bigr] \;=\; \mathbb{E}[N(s,a)] \quad\text{a.s.,}
    \end{equation}
    which yields the Basu-type identity $\mathbb{E}[u(s,a)] = \mathbb{E}[N(s,a)]/\mathbb{E}[\overline{\delta}(s',a')]$.
\end{assumption}

Consequently, the AEA dynamics are globally stable. The two-timescale coupling ensures that the aggregation parameter $\overline{\kappa}$ remains within the stability region $\mathcal{K}_{\text{stable}}$, preserving the contraction properties of the operator, while the critic ensemble converges to a unique projected fixed point. Building on this convergence, the following result demonstrates that the AEA estimator provides an asymptotically unbiased and consistent approximation of the Bellman target.

\begin{theorem} \label{aea:consistency}
Let $B(s,a) = \mu_{\mathcal{Q}}(s,a) - Q^\pi(s,a)$ denote the ensemble approximation bias. Under the two-timescale process $(\mathcal{Q}_t, \overline{\kappa}_t) \to (\mathcal{Q}^*, \overline{\kappa}^*)$, the AEA target $y_{\overline{\kappa}^*}(s,a) = \mu_{\mathcal{Q}}(s,a) + \overline{\kappa}^* \delta(s,a)$ is asymptotically unbiased ($\mathbb{E}[y_{\overline{\kappa}^*}] = Q^\pi$) under Assumption~\ref{assump:symmetry}, provided the equilibrium parameter $\overline{\kappa}^* = -\mathbb{E}[B] / \mathbb{E}[\overline\delta]$ lies within 
$\mathcal{K}_{\text{stable}} \triangleq \left \{ \overline{\kappa} \in \mathbb{R} : |\overline{\kappa}| < (1 - \gamma)/(\gamma \cdot \text{Lip}(\overline\delta)) \right \}$. 
In this regime, the estimator $y_{\overline{\kappa}^*}$ is consistent such that $y_{\overline{\kappa}^*} \xrightarrow{p} Q^\pi$ as $N \to \infty$, with error concentration governed by:
\begin{align*}
\mathbb{P}\left( |y_{\overline{\kappa}^*}(s,a) - Q^\pi(s,a)| \geq \epsilon \right) \leq 2 \exp\left( -{N \epsilon^2}/{2 C \sigma^2} \right), \end{align*}
where $C$ depends on the sub-Gaussian norm of the ensemble. If the unbiased minimizer lies outside $\mathcal{K}_{\text{stable}}$, the equilibrium parameter $\overline{\kappa}^*$ converges to the projection $\Pi_{\mathcal{K}_{\text{stable}}}(\overline{\kappa}^*)$, maintaining the spectral radius of the ensemble Bellman operator strictly below unity while minimizing residual bias.
\end{theorem}

This result establishes that AEA asymptotically eliminates overestimation bias, ensuring that the target values concentrate around the true $Q$-function as the ensemble size increases. However, in practice, function approximation is finite, and residual errors persist. Our next result characterizes the safety margin required for monotonic policy improvement, ensuring that the learned advantage is robust to both the structural limitations of the NTK space and estimation bias at equilibrium.
\begin{theorem}\label{aea:improvement}
Let $\mathcal{Q}^*$ be the ensemble equilibrium for policy $\pi_k$ as defined in Theorem~\ref{thm:aea_convergence}, with mean $\mu_{\mathcal{Q}^*}$ and disagreement $\overline{\delta}^*$. Define the structural approximation error and residual estimation bias in expectation under the current state-action occupancy:
\begin{align*}
\epsilon_{\text{approx}} &\;=\; \mathbb{E}_{(s,a)\sim d^{\pi_k}}\!\bigl[\bigl|\Pi_{\mathcal{D}} Q^{\pi_k}(s,a) - Q^{\pi_k}(s,a)\bigr|\bigr], \\
\epsilon_{\text{est}} &\;=\; \mathbb{E}_{(s,a)\sim d^{\pi_k}}\!\bigl[\bigl|\mu_{\mathcal{Q}^*}(s,a) + \overline{\kappa}^* \overline{\delta}^*(s,a) - Q^{\pi_k}(s,a)\bigr|\bigr].
\end{align*}
Assume further (i) the state-distribution shift is bounded, $\|d^{\pi_{k+1}} - d^{\pi_k}\|_{\mathrm{TV}} \le \varepsilon_{\mathrm{TV}}$, and (ii) the soft-Q residual is uniformly bounded, $|\widetilde{Q}_\kappa(s,a) - \alpha\log\pi(a|s) - V^{\pi_k}_{\mathrm{soft}}(s)| \le R_{\max}$. Let $\pi_{k+1}$ be the maximizer of
\begin{equation*}
\mathcal{J}(\pi) := \mathbb{E}_{s\sim d^{\pi_k},\,a\sim\pi(\cdot|s)}\!\left[\widetilde{Q}_{\kappa}(s,a) - \alpha\log\pi(a|s)\right].
\end{equation*}
Then $V^{\pi_{k+1}}_{\mathrm{soft}}(s)\ge V^{\pi_k}_{\mathrm{soft}}(s)$ for $d^{\pi_{k+1}}$-almost every $s\in\mathcal{S}$ provided
\begin{equation*}
\mathcal{J}(\pi_{k+1}) - \mathcal{J}(\pi_k) \;>\; \frac{2\gamma}{1-\gamma}\bigl(\epsilon_{\text{approx}} + \epsilon_{\text{est}} + R_{\max}\,\varepsilon_{\mathrm{TV}}\bigr).
\end{equation*}
\end{theorem}
To analyze how this optimization minimizes value estimation error, we formalize the shrinkage of $\epsilon_{\text{est}}$ as a function of the ensemble size $N$. This establishes that the estimation error is a dynamic quantity that vanishes under optimal aggregation, rather than a static penalty inherent to the architecture.
\begin{theorem}\label{aea:shrinkage}
Let $\epsilon_{\text{est}} = \mathbb{E}_{(s,a)\sim d^\pi}\!\bigl[|\mu_{\mathcal{Q}^*}(s,a) + \overline{\kappa}^* \overline\delta^*(s,a) - Q^\pi(s,a)|\bigr]$ be the residual estimation error at equilibrium. Under the two-timescale convergence $(\mathcal{Q}_t, \overline{\kappa}_t) \to (\mathcal{Q}^*, \overline{\kappa}^*)$, where the ensemble errors are $\sigma$-sub-Gaussian, the following bound holds with probability at least $1 - \alpha$:
\begin{align*}
\epsilon_{\text{est}} \leq \underbrace{| \mathbb{E}[B] + \overline{\kappa}^* \mathbb{E}[\overline\delta] |}_{\text{Residual Bias}} + \underbrace{C \sigma \sqrt{{\log(2/\alpha)}/{N}}}_{\text{Concentration Error}},
\end{align*}
where $C > 0$ is a constant. If the unbiased equilibrium $\overline{\kappa}^* = -\mathbb{E}[B]/\mathbb{E}[\overline\delta]$ satisfies the stability constraint $\overline{\kappa}^* \in \mathcal{K}_{\text{stable}}$, the residual bias vanishes, and $\epsilon_{\text{est}}$ shrinks at the rate $\mathcal{O}(1/\sqrt{N})$. If ${\overline{\kappa}^* \notin \mathcal{K}_{\text{stable}}}$, then $\epsilon_{\text{est}}$ converges to the constrained minimum $\min_{\overline{\kappa} \in \mathcal{K}_{\text{stable}}} | \mathbb{E}[B] + \overline{\kappa} \mathbb{E}[\overline\delta] |$.
\end{theorem}
The key theoretical advantage of AEA lies in its ability to minimize bias and variance simultaneously: whereas the REDQ min-pooling operator introduces a static variance penalty that persists regardless of ensemble size, AEA leverages the full ensemble to drive the estimator variance toward zero. The following result demonstrates that the AEA target is strictly more efficient and asymptotically exact.
\begin{theorem}\label{thm:mse_comparison}

Under the equilibrium conditions defined in Theorem~\ref{thm:aea_convergence}, the following hold:

\begin{itemize}

\item The mean squared error (MSE) of the AEA target $y_{\overline{\kappa}} = \mu_{\mathcal{Q}} + \overline{\kappa} \overline\delta$ relative to $Q^\pi$ is:
\begin{align*}
\text{MSE}(y_{\overline{\kappa}}) = \text{Var}(\mu_{\mathcal{Q}}) + \overline{\kappa}^2 \text{Var}(\overline\delta) + 2\overline{\kappa} \text{Cov}(\mu_{\mathcal{Q}}, \overline\delta) + (\mathbb{E}[B] + \overline{\kappa} \mathbb{E}[\overline\delta])^2.
\end{align*}

\item Let $y_{\text{redq}} = \min_{j \in \mathcal{M}} \{Q_j\}$ with $|\mathcal{M}|=2$. If $\overline{\kappa}^* = -\mathbb{E}[B]/\mathbb{E}[\overline\delta] \in \mathcal{K}_{\text{stable}}$, there exists a finite ensemble size $N_0$ such that for all $N > N_0$,
$\text{MSE}(y_{\overline{\kappa}^*}) < \text{MSE}(y_{\text{redq}})$.

\item In the infinite-width and large-ensemble limit ($N \to \infty$), the AEA target satisfies
$\lim_{N \to \infty} \text{MSE}(y_{\overline{\kappa}^*}) = 0$.

\end{itemize}
\end{theorem}

To characterize the fundamental limit of estimation performance, we evaluate the ensemble through the lens of information theory. We define the \emph{Fisher information} $\mathcal{I}(Q^\pi)$ as the expected value of the squared partial derivative of the log-likelihood function $f(\mathcal{Q}; Q^\pi)$ with respect to the parameter $Q^\pi$: $\mathcal{I}(Q^\pi) = \mathbb{E} [ ( \frac{\partial}{\partial Q^\pi} \log f(\mathcal{Q}; Q^\pi) )^2 ]$, representing the total information the ensemble $\mathcal{Q}$ carries about the true action value. Unlike min-pooling, which relies on resampling to eventually incorporate ensemble-wide information over many iterations, AEA achieves high informational throughput in every update by utilizing the full ensemble simultaneously. According to the \emph{Cramér–Rao lower bound (CRLB)}, the variance of any unbiased estimator $\widehat{Q}$ is bounded by the reciprocal of this information: $\text{Var}(\widehat{Q}) \geq 1/\mathcal{I}(Q^\pi)$. While min-pooling techniques like REDQ discard a significant portion of the ensemble’s collective information by subsampling a small subset $M \ll N$—effectively reducing the available Fisher information to $\mathcal{I}(\mathcal{M})$—AEA leverages the full ensemble to maximize informational throughput and achieve the optimal variance-reduction rate.

\begin{theorem}\label{thm:aea_sample_efficiency}

Assume the ensemble errors $\epsilon_i$ are i.i.d.\ $\mathcal{N}(0, \sigma^2)$, so that $\{Q_i\}$ forms a Gaussian location family with parameter $\mu = Q^\pi + \mathbb E[B]$. For any finite ensemble size $N > M = 2$, the following efficiency properties hold:

\begin{itemize}
\item The variance of the AEA target $y_{\overline{\kappa}^*}$ satisfies:
$\text{Var}(y_{\overline{\kappa}^*}) \;=\; \frac{\sigma^2}{N} \left( 1 + (\overline{\kappa}^*)^2 C_{\delta} \right)$,
where $C_{\delta}$ is the asymptotic variance constant of the U-statistic associated with the chosen disagreement measure. Consequently, there exists a finite ensemble size $N_0$ such that for all $N>N_0$, $\text{Var}(y_{\overline{\kappa}^*}) < \text{Var}(y_{\text{redq}}) = \sigma^2\cdot\mathcal{V}(M)$, where $\mathcal{V}(M)$ is the variance factor of the $M$-th order statistic.

\item The Fisher information utilized by the AEA target scales as $\mathcal{I}(y_{\overline{\kappa}^*}) = \Omega(N)$, whereas the \emph{per-update} information utilized by the REDQ target is restricted by the subset size, $\mathcal{I}(y_{\text{redq}}) = \mathcal{O}(1)$.

\end{itemize}
\end{theorem}

The second result implies that the AEA target approaches the Cramér–Rao lower bound of $\mathcal{O}(1/N)$, whereas the REDQ target remains limited by the subset size $M$, demonstrating the informational optimality of adaptive aggregation. 
REDQ's resampling eventually incorporates all critics over many iterations; the advantage of AEA lies in its per-update informational throughput.

\textbf{The theory-to-practice gap.} While the above results establish the stability and concentration of AEA under sub-Gaussian noise and NTK assumptions, we acknowledge a potential divergence in practical deep RL. In the infinite-width limit, the NTK remains static, and the ensemble statistics follow predictable sub-Gaussian concentration. In finite-width networks with active feature learning, the lazy training assumption may be violated, leading to nonstationary disagreement dynamics. Furthermore, although our Cramér–Rao analysis demonstrates that AEA utilizes the full informational throughput of the ensemble, performance in high-dimensional state spaces is subject to the quality of the state–action sampling distribution $d^\pi$. If the exploration policy fails to cover critical regions of the manifold, the unbiasedness at equilibrium remains local to the visited transitions. Finally, although our analysis assumes independent ensemble errors $\epsilon_i$, practical use of a shared replay buffer introduces correlations that may reduce the effective ensemble size $N_{\text{eff}}$. Nevertheless, our empirical results in Section 5 demonstrate that the adaptive mechanism is robust to these deviations, maintaining superior sample efficiency even when the strict assumptions of the NTK regime are relaxed.

\section{Related Work} \label{sec::related_work}

Our work investigates the adaptive learning of conservatism and optimism in off-policy actor-critic methods without auxiliary hyperparameters or external decision layers. We distinguish our approach by its ability to dynamically construct the critic target ($\overline{Q}$) and actor ($\widetilde{Q}$) references.

\textbf{Static and decoupled conservatism.} Standard approaches like TD3 \citep{fujimoto2018addressing} and SAC \citep{haarnoja2018soft} employ static min-clipping ($\min_{i} Q_i$). While stable, this uniform pessimism often leads to underexploration \citep{lan2020maxmin, ciosek2019better}. 
MaxMin Q-learning \citep{lan2020maxmin} generalizes this by taking the minimum over the entire ensemble, reducing but not eliminating the pessimistic bias.
To mitigate this, OAC \citep{ciosek2019better} introduced a decoupled actor reference using a weighted standard deviation ($\mu + \beta \sigma$), although it relies on manually tuned coefficients and separate exploration policies. USAC \citep{tasdighi2025improving} formalized this decoupling by treating ensemble values as random variables under a Laplace assumption, demonstrating that independent parameters $(\kappa, \overline \kappa)$ for the critic and actor improve performance. However, USAC requires these parameters to be tuned offline per task and learning regime. AEA inherits the decoupled parameterization of USAC but contributes the missing online adaptation mechanism: a scale-invariant update rule that adjusts both parameters based on training dynamics without auxiliary optimization or offline search.

\textbf{Distributional and bandit-based adaptation.} Distributional methods like TQC \citep{kuznetsov2020controlling} and TOP \citep{moskovitz2021tactical} attempt adaptation through quantile truncation or bandit layers. TQC requires per-task tuning of the truncation level, whereas TOP utilizes an external bandit to switch between discrete, pre-specified $\beta$ values. These methods introduce significant methodological overhead and remain sensitive to the hyperparameters of the auxiliary decision layers.
More recently, ADDQ \citep{doering25a} proposed adaptive weighting of ensemble members based on local distributional information. 
However, ADDQ was developed for discrete action spaces, and its weighting scheme is sensitive to local bias in individual ensemble members. The authors explicitly acknowledge the extension to continuous control as an open direction for future work, leaving the problem of adaptive ensemble aggregation for actor-critic methods unaddressed.

\textbf{Ensemble regimes.} High-UTD (Update-to-Data) regimes amplify bootstrapping errors, necessitating larger ensembles \citep{nikishin2022primacy}. REDQ \citep{chen2021randomized} addresses this by combining a randomized subset-min target for the critic with a mean aggregator for the actor. While robust in complex hardware applications \citep{smith2023walk}, REDQ’s design is fixed and specialized for high-UTD settings. Similarly, other large-ensemble approaches, such as GPL \citep{cetin2023learning} and Sunrise \citep{lee2021sunrise}, rely on task-specific entropy targets or fixed schedules that do not generalize across different learning regimes.

\section{Experiments} \label{sec::experiments}

\begin{wraptable}{r}{0.8\textwidth}
\vspace{-\baselineskip}
\caption{InterQuantile Mean (IQM) of the final return with ensemble size 10 (AEA, REDQ, and SAC-N) or 2 (DSAC, DRND), Update-to-Data Ratio 20, and interaction budget \num{300000}, averaged over evaluation repetitions and ten seeds. $\pm$ denotes the standard deviation over seeds. Highest mean marked bold.}
\vspace{-0.3em}
\label{tab::main_iqm}
\begin{center}
\adjustbox{max width=0.79\textwidth}{
\begin{tabular}{lccccc}
\toprule
& AEA & REDQ & SAC-$N$ & DRND & DSAC \\
\emph{MuJoCo-v5} & (ours) & \citep{chen2021randomized} & \citep{haarnoja2018soft} & \citep{yang2024exploration} & \citep{ma2025dsac}\\
\cmidrule(lr){1-6}
\texttt{Ant} & $4744\sd{\pm260}$ & $\mathbf{4828\sd{\pm398}}$ & $1\sd{\pm1}$ &$797\sd{\pm 560}$ & $776\sd{\pm177}$\\
\texttt{HalfCheetah} & $\mathbf{9661\sd{\pm395}}$ & $8678\sd{\pm813}$ & $5425\sd{\pm191}$ & $7892\sd{\pm1180}$ & $9059\sd{\pm1191}$\\
\texttt{Hopper} & $\mathbf{3555\sd{\pm32}}$ & $3415\sd{\pm153}$ & $2063\sd{\pm1034}$ & $3015\sd{\pm840}$ & $2640\sd{\pm783}$\\
\texttt{Humanoid} & $5376\sd{\pm87}$ & $\mathbf{5404\sd{\pm61}}$ & $5341\sd{\pm124}$ & $2853\sd{\pm1684}$ & $908\sd{\pm171}$\\
\texttt{Walker2d} & $4652\sd{\pm267}$ & $4749\sd{\pm170}$ & $1035\sd{\pm830}$ & $5054\sd{\pm414}$ & $\mathbf{5362\sd{\pm349}}$ \\[0.5em]
\emph{DMC} \\
\cmidrule(lr){1-6}
\texttt{Cheetah-run} & $827\sd{\pm9}$ & $866\sd{\pm6}$ & $635\sd{\pm26}$ & $\mathbf{873\sd{\pm21}}$ & $753\sd{\pm37}$\\
\texttt{Hopper-hop} & $\mathbf{86\sd{\pm75}}$ & $54\sd{\pm81}$ & $0\sd{\pm1}$ & $3\sd{\pm7}$ & $54\sd{\pm58}$\\
\texttt{Hopper-stand} & $\mathbf{935\sd{\pm16}}$ & $109\sd{\pm268}$ & $0\sd{\pm1}$ & $251\sd{\pm339}$ & $39\sd{\pm58}$\\
\texttt{Humanoid-run} & $\mathbf{144\sd{\pm5}}$ & $131\sd{\pm8}$ & $1\sd{\pm1}$ & $1\sd{\pm0}$ & $17\sd{\pm32}$\\
\texttt{Humanoid-stand} & $\mathbf{665\sd{\pm55}}$ & $559\sd{\pm31}$ & $9\sd{\pm1}$ & $213\sd{\pm144}$ & $284\sd{\pm190}$\\
\texttt{Humanoid-walk} & $\mathbf{484\sd{\pm14}}$ & $454\sd{\pm35}$ & $2\sd{\pm1}$ & $166\sd{\pm160}$ & $2\sd{\pm0}$\\
\texttt{Quadruped-run} & $\mathbf{856\sd{\pm19}}$ & $847\sd{\pm24}$ & $43\sd{\pm52}$ & $224\sd{\pm379}$ & $536\sd{\pm80}$\\
\texttt{Quadruped-walk} & $\mathbf{947\sd{\pm14}}$ & $932\sd{\pm16}$ & $112\sd{\pm90}$ & $184\sd{\pm285}$ & $555\sd{\pm371}$ \\
\texttt{Walker-run} & $\mathbf{741\sd{\pm27}}$ & $649\sd{\pm56}$ & $258\sd{\pm84}$ & $689\sd{\pm93}$ & $658\sd{\pm86}$\\
\bottomrule
\end{tabular}}
\end{center}
\vspace{-\baselineskip}
\end{wraptable}
\textbf{Setup.} We evaluate AEA across five MuJoCo environments \citep{todorov2012mujoco} and nine DeepMind Control Suite tasks \citep{tassa2018deepmind} to investigate whether an ensemble aggregator can successfully tune itself online to navigate the trade-off between bias suppression and signal preservation. Our primary benchmark is the large-ensemble ($N=10$) and multi-UTD regime ($G=20$), where the potential for variance reduction is highest but the risk of bootstrapping bias is most severe. We conduct a consistency check in the standard twin-critic ($N=2$) and single-UTD ($G=1$) regime to ensure AEA maintains parity with established baselines. By tracking the trajectories of $(\overline{\kappa}, \kappa)$, we empirically test whether decoupling actor and critic references is the mechanical necessity required to reconcile bias suppression with the discovery of optimal policies.

\textbf{Baselines.} We compare AEA against \emph{REDQ} \citep{chen2021randomized}, which represents the current state-of-the-art in fixed ensemble aggregation and has proven effective on complex physical hardware \citep{smith2023walk}. We further benchmark against distributional RL approaches \citep{bellemare2023distributional} as the strongest alternative to ensemble building for capturing value function uncertainty. We represent the state of the art of continuous control with implicit quantile network learning with \emph{Distributional Soft Actor Critic (DSAC)} \citep{ma2025dsac}. We implement an enhanced version of the algorithm by integrating the fully parameterized quantile proposal mechanism of \citet{yang2019fully}, which learns an optimal distribution of quantile fractions rather than relying on uniform sampling.
We represent assumed value function density estimation approaches with \emph{Distributional Random Network Distillation (DRND)} \citep{yang2024exploration}. In this framework, the distillation error between a trainable predictor and a fixed random prior serves as a proxy for the epistemic uncertainty manifold.
As USAC \citep{tasdighi2025improving} showed better performance than \emph{OAC} \citep{ciosek2019better} and \emph{TOP} \citep{moskovitz2021tactical}, we do not compare against them in our experiments.

\textbf{Results.} We measure performance using the InterQuartile Mean (IQM) of final undiscounted returns across seeds \citep{agarwal2021deep}. For evaluation, the agent completes 20 episodes using a deterministic policy derived from the mean of the learned action distribution. \Cref{tab::main_iqm} summarizes results in the large-ensemble ($N=10$) and multi-UTD ($G=20$) regime. AEA outperforms all baselines in the vast majority of environments across both suites. This performance advantage over REDQ validates our theoretical findings, particularly \Cref{thm:mse_comparison}, which predicts that adaptive aggregation more effectively mitigates bootstrapping bias without sacrificing signal in high-sample-reuse settings. For completeness, \cref{tab::main_results_final_return,tab::main_results_iqm} in the Appendix provide final returns and the Area Under the Learning Curve (AULC) to quantify learning speed and stability.
\begin{wrapfigure}{r}{0.8\textwidth}
\vspace{-\baselineskip}
\centering

\includegraphics[width=0.38\textwidth]{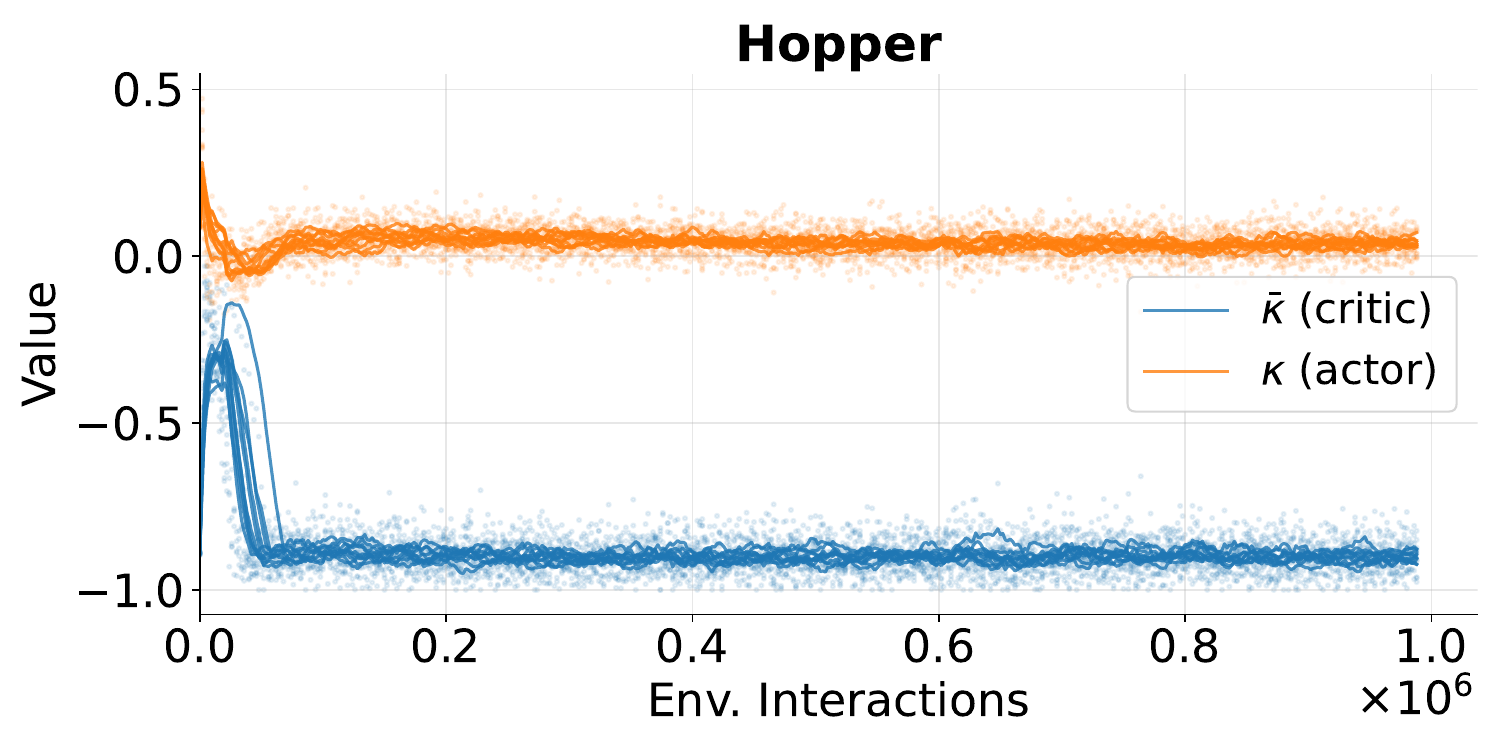}
\includegraphics[width=0.38\textwidth]{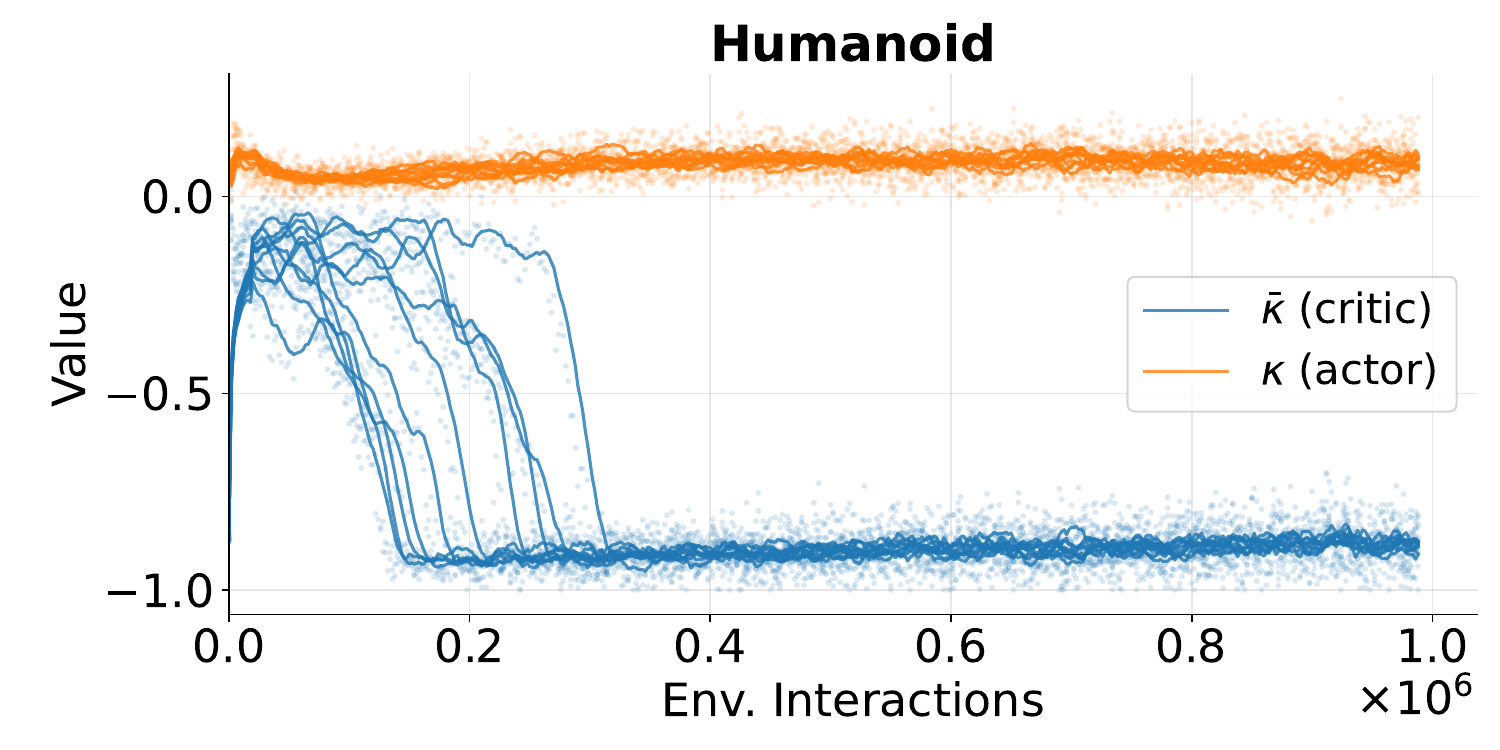}\\
\includegraphics[width=0.38\textwidth]{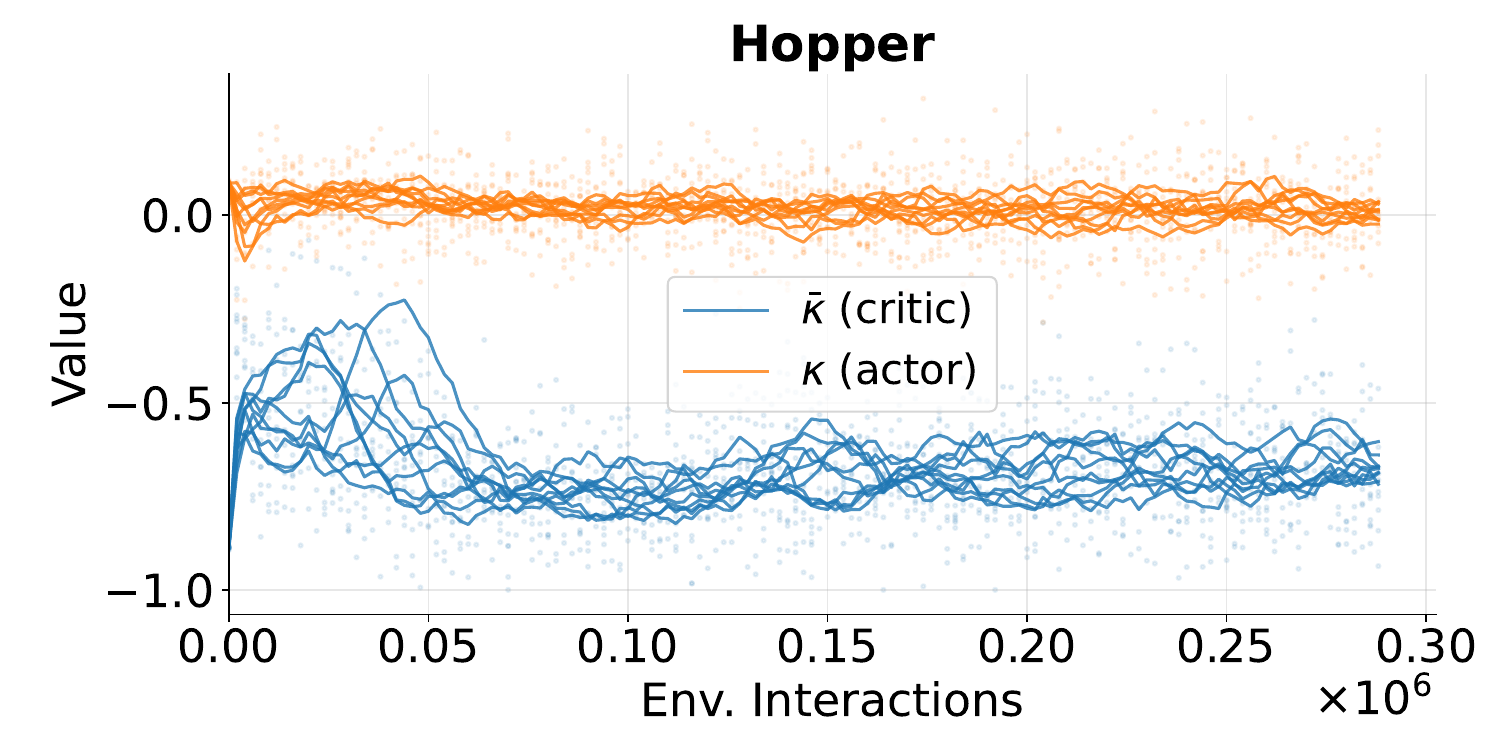}
\includegraphics[width=0.38\textwidth]{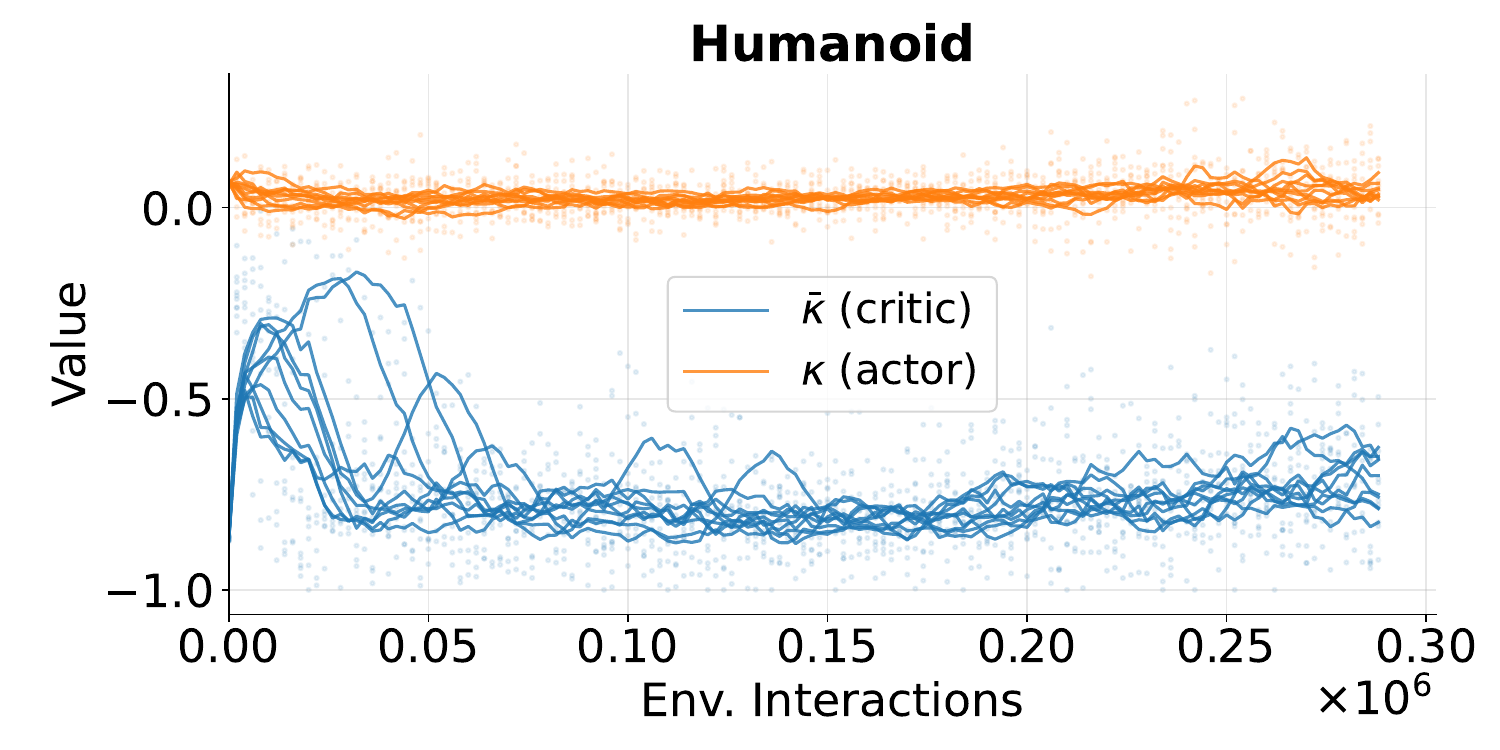}
\caption{ $\overline{\kappa}$ (critic) and $\kappa$ (actor) trajectories on MuJoCo. \textbf{Top row:} (${N=2, G=1}$), \textbf{Bottom row:} ($N=10, G=20$). Trajectories of all tasks, seeds, and learning regimes can be found in \cref{appendix:additional_results}.} \label{fig:kappa_trajectory}
\vspace{-\baselineskip}
\end{wrapfigure}
The relative rankings remain consistent across these metrics. Finally, \cref{tab::dmc_final_return,tab::dmc_iqm,tab::dmc_aulc} demonstrate that AEA performs comparably to the state of the art in the twin-critic ($N=2$, $G=1$) regime, confirming that our adaptive mechanism remains robust even when the theoretical conditions for its primary advantages are not present. Notably, SAC, DRND, and DSAC which struggle in the $N=10,G=20$ setting compared to AEA and REDQ perform very competitively in the twin-critic regime.  
\Cref{appendix:ablations} provides ablations on the initialization sensitivity of $\overline\kappa$ and the effect of replacing the learned calibration with fixed reference constructions.

\textbf{Interpretability.}  \Cref{fig:kappa_trajectory} (see also \Cref{fig:kappas_trajectory:interactive:mujoco,fig:kappas_trajectory:sample_efficient:mujoco,fig:kappas_trajectory:interactive:dmc,fig:kappas_trajectory:sample_efficient:dmc}) visualizes the trajectories of the critic parameter $\overline{\kappa}$ and the actor parameter $\kappa$. Three patterns are notable: (i) The critic target learns a negative bias ($\overline{\kappa} < 0$) to enforce stability via conservatism, whereas the actor reference typically remains neutral or optimistic ($\kappa \geq 0$) to facilitate exploration, validating the necessity of decoupling bootstrapping from policy improvement. (ii) The magnitudes and temporal evolutions of $(\overline{\kappa}, \kappa)$ vary significantly across environments, confirming that static, hard-coded aggregation rules are insufficient for diverse MDPs. (iii) AEA adapts to different update frequencies without manual tuning. In high-UTD regimes ($G=20$), $\overline{\kappa}$ exhibits higher volatility, indicating an automated stabilization response to the increased gradient noise and potential overestimation inherent in sample-efficient learning.

\section{Impact and Limitations}\label{sec:impact}
As predicted by Theorems \ref{thm:mse_comparison} and \ref{thm:aea_sample_efficiency}, the empirical results confirm that AEA successfully leverages the full ensemble's Fisher information to reduce variance without falling into the "pessimistic underexploration" trap. While the approach performs similarly to the state of the art in the two-critic, single-UTD regime, it significantly outperforms traditional architectures in the high-ensemble, high-UTD regime, where static min-pooling fails. Beyond the theoretical gains, this estimator offers practical utility for robotics and autonomous systems applications, where overestimation may lead to hardware failure and underestimation may cause latencies.
Despite these gains, the approach is subject to several theoretical and methodological limitations. The convergence analysis does not fully capture the non-stationary feature learning dynamics of finite-width deep networks. The assumption of independent ensemble errors is partially violated in practice by the use of a shared replay buffer, likely reducing the effective ensemble size $N_{\text{eff}}$. Furthermore, although the normalized objective ensures numerical stability, it introduces a two-timescale coupling that may require careful tuning of the relative learning rates $\zeta/\eta$ in environments with sparse rewards. These results open new questions regarding whether higher-order moments of the ensemble distribution could further refine the exploration-exploitation trade-off and how adaptive aggregation might scale to multi-agent settings, where bias dynamics are even more volatile.

\bibliographystyle{plainnat}
\bibliography{refs}

\newpage

\appendix

\begin{center}
    \Huge 
    \textsc{Appendix}
\end{center}

\section{Prior Results}\label{appsec:prior_results}

In our analysis, we rely on the foundational result of \citet{kushner1978stochastic} to prove the convergence of the ensemble parameters. We state the theorem here for completeness.

\begin{theorem}[\citep{kushner1978stochastic}]\label{thm:kushner_clark}
Consider a stochastic approximation algorithm of the form:
$\theta_{t+1} = \theta_t + \eta_t [h(\theta_t) + \xi_t + \psi_t]$
where $\theta_t \in \mathbb{R}^p$ are the parameters, $\eta_t$ is the step size, $h(\cdot)$ is a continuous mean vector field, $\xi_t$ is a noise term, and $\psi_t$ is a vanishing bias term. Assume the following conditions:
\begin{itemize}
    \item $\eta_t > 0$, $\sum_{t=0}^\infty \eta_t = \infty$, and $\lim_{t \to \infty} \eta_t = 0$.

    \item For any $T > 0$, $\lim_{n \to \infty} P \left( \sup_{j \geq n} \left\| \sum_{i=n}^{m(n,T)-1} \eta_i \xi_i \right\| \geq \epsilon \right) = 0$ for all $\epsilon > 0$, where $m(n,T) = \max \{ k : \sum_{i=n}^{k-1} \eta_i \leq T \}$.

    \item $\psi_t \to 0$ almost surely as $t \to \infty$.

    \item The sequence $\{\theta_t\}$ is bounded almost surely, and the ODE $\dot{\theta} = h(\theta)$ has a uniquely globally asymptotically stable equilibrium $\theta^*$.
\end{itemize}
Under these conditions, the sequence $\{\theta_t\}$ converges to $\theta^*$ almost surely:
$P\left(\lim_{t \to \infty} \theta_t = \theta^*\right) = 1.$
\end{theorem}

\begin{definition}[Completeness]
A statistic $T(X)$ is complete for a family of distributions $\mathcal{P} = \{P_\theta : \theta \in \Theta\}$ if, for every measurable function $g$, the condition $\mathbb{E}_\theta[g(T)] = 0$ for all $\theta \in \Theta$ implies that $P_\theta(g(T) = 0) = 1$ for all $\theta \in \Theta$.
\end{definition}

\begin{definition}[Ancillarity]
 A statistic $A(X)$ is ancillary if its distribution does not depend on the parameter $\theta$.
\end{definition}

\begin{theorem}[\citet{basu1955statistics}]\label{thm:basu}
 Let $X_1, \dots, X_N$ be a sample from a distribution $P_\theta$. If $T(X)$ is a boundedly complete sufficient statistic for $\theta$ and $A(X)$ is an ancillary statistic for $\theta$, then $T(X)$ and $A(X)$ are independent.   
\end{theorem}

\begin{definition}[$U$-statistic]
 Let $X_1, X_2, \dots, X_N$ be i.i.d. random variables. A $U$-statistic of degree $m$ with kernel $h: \mathbb{R}^m \to \mathbb{R}$ is defined as:
 \begin{align*}
U_N = \frac{1}{\binom{N}{m}} \sum_{1 \leq i_1 < \dots < i_m \leq N} h(X_{i_1}, \dots, X_{i_m}).
 \end{align*}
\end{definition}

\begin{theorem}[\citet{hoeffding1963probability}]\label{thm:general_hoeffding}
Assume the kernel $h$ is bounded such that $a \leq h(x_1, \dots, x_m) \leq b$ for all inputs. For any $\epsilon > 0$, the following concentration bound holds:
\begin{align*}
    P\left( |U_N - \mathbb{E}[U_N]| \geq \epsilon \right) \leq 2 \exp\left( - \frac{2 \lfloor N/m \rfloor \epsilon^2}{(b - a)^2} \right).
\end{align*}    
\end{theorem}

\begin{theorem}[Arcones-type concentration for $U$-statistics of order 2 \citep{arcones1995bernstein}]\label{thm:subgaussian_ustat}
Let $X_1,\dots,X_N$ be i.i.d., $h$ a symmetric kernel of order $2$, and set $\zeta_1^2 := \mathrm{Var}(h_1(X_1))$ where $h_1(x) = \mathbb{E}[h(x,X_2)] - \mathbb{E}[h]$. Suppose the centered kernel satisfies the sub-exponential bound $\|h(X_1,X_2) - \mathbb{E}[h]\|_{\psi_1} \leq \tau$. Then there exists a universal constant $c > 0$ such that for all $\epsilon > 0$,
\[
P\bigl(|U_N - \mathbb{E}[U_N]| \geq \epsilon\bigr) \;\leq\; 2\exp\!\left(-c\min\!\left(\frac{N\epsilon^2}{\zeta_1^2},\; \frac{N\epsilon}{\tau}\right)\right).
\]
\end{theorem}

\begin{corollary}[MAD on sub-Gaussian samples]\label{cor:mad_concentration}
If $X_1,\dots,X_N$ are i.i.d.\ $\sigma$-sub-Gaussian and $h(x,y) = |x-y|$, then $\tau \leq c_1\sigma$ and $\zeta_1 \leq c_2\sigma$ for universal constants $c_1,c_2 > 0$. Consequently, $\bar\delta_N = \binom{N}{2}^{-1}\sum_{i>j}|X_i - X_j|$ satisfies
\[
P\bigl(|\bar\delta_N - \mathbb{E}[\bar\delta_N]| \geq \epsilon\bigr) \leq 2\exp\!\left(-c\min\!\left(\frac{N\epsilon^2}{\sigma^2},\; \frac{N\epsilon}{\sigma}\right)\right).
\]
For $\epsilon \leq \sigma$, the first term dominates and concentration is at rate $\mathcal{O}(\sigma/\sqrt{N})$.
\end{corollary}

\section{New Results}\label{appsec:proofs}

\subsection{Proof of Theorem \ref{thm:aea_convergence}}

Under the condition $\lim_{t \to \infty} \zeta_t / \eta_t = 0$, the evolution of the network parameters $\theta_t^{(i)}$ (and thus the ensemble $\mathcal{Q}_t$) occurs at a faster timescale than the evolution of the aggregation scalar $\overline{\kappa}_t$. In the NTK regime, the functional dynamics of the ensemble for a quasi-static $\overline{\kappa}$ are governed by the linear ODE:
\begin{align} \label{eq:fast_ode}
\dot{Q}_i = \mathbf{H} \left( \Pi_{\mathcal{D}} (T_{\overline{\kappa}}^\pi \mathcal{Q})_i - Q_i \right), \quad \forall i \in [N]. 
\end{align} 
Since the NTK $\mathbf{H}$ is positive semi-definite and the operator $\Pi_{\mathcal{D}} T_{\overline{\kappa}}^\pi$ is a contraction in $\mathcal{L}^\infty(d^\pi)$ for any $\overline{\kappa} \in \mathcal{K}_{\text{stable}}$, the ODE \eqref{eq:fast_ode} possesses a globally asymptotically stable equilibrium. Specifically, for a fixed $\overline{\kappa}$, $\mathcal{Q}_t$ converges to the unique fixed point $\mathcal{Q}^*(\overline{\kappa}) = \Pi_{\mathcal{D}} T_{\overline{\kappa}}^\pi \mathcal{Q}^*(\overline{\kappa})$.

The scalar $\overline{\kappa}_t$ evolves according to the slower timescale $\zeta_t$. By the theory of stochastic approximation, its asymptotic behavior is determined by the mean ODE obtained by averaging over the equilibrium distribution of the fast process:
\begin{align} \label{eq:slow_ode}
\dot{\overline{\kappa}} = \gamma \mathbb{E}_{d^\pi} \left[ \text{sign} \left( \widetilde{Q}_{\kappa}(s,a) - y_{\overline{\kappa}}(s,a,r,s') \right) \right],
\end{align}
where $y_{\overline{\kappa}}$ is evaluated using the equilibrium ensemble $\mathcal{Q}^*(\overline{\kappa})$. We define the risk function ${J(\overline{\kappa}) = \mathbb{E}_{d^\pi} [ |\widetilde{Q}_{\kappa} - y_{\overline{\kappa}}| / \overline{\delta} ]}$. Recalling that $\partial y_{\overline{\kappa}} / \partial \overline{\kappa} = \gamma \overline{\delta}$, the gradient is:
\begin{align*}
\nabla_{\overline{\kappa}} J(\overline{\kappa}) = \mathbb{E}_{d^\pi} \left[ \frac{-\text{sign}(\widetilde{Q}_{\kappa} - y_{\overline{\kappa}})}{\overline{\delta}} \cdot \gamma \overline{\delta} \right] = -\gamma \mathbb{E}_{d^\pi} [ \text{sign}(\widetilde{Q}_{\kappa} - y_{\overline{\kappa}}) ].
\end{align*}
Thus, the ODE in \eqref{eq:slow_ode} is exactly $\dot{\overline{\kappa}} = -\nabla_{\overline{\kappa}} J(\overline{\kappa})$, representing a continuous-time gradient descent on the convex surface $J$.

To apply the Kushner-Clark theorem (\cref{thm:kushner_clark}), we verify the following conditions:
\begin{itemize}
\item {\bf Boundedness:} The projection onto $\mathcal{K}_{\text{stable}}$ ensures $\overline{\kappa}_t$ remains in a compact set where $T_{\overline{\kappa}}^\pi$ is a contraction.
\item {\bf Continuity:} The map $\overline{\kappa} \mapsto \mathcal{Q}^*(\overline{\kappa})$ is continuous because the fixed point of a contraction is continuous in its parameters.
\item {\bf Martingale Noise:} The noise terms $w_t = \text{sign}(\dots) - \mathbb{E}[\text{sign}(\dots)]$ are martingale differences. Since the sign function is bounded, the noise satisfies the required second-moment constraints.
\end{itemize}
As $t \to \infty$, the fast process $\mathcal{Q}_t$ tracks the manifold of fixed points $\mathcal{Q}^*(\overline{\kappa}_t)$ with an error $o(1)$. Simultaneously, the slow process $\overline{\kappa}_t$ converges to the unique minimizer $\overline{\kappa}^*$ of $J(\overline{\kappa})$. The joint process $(\mathcal{Q}_t, \overline{\kappa}_t)$ therefore converges almost surely to the unique equilibrium $(\mathcal{Q}^*(\overline{\kappa}^*), \overline{\kappa}^*)$. \hfill $\square$

\subsection{Proof of Theorem \ref{aea:consistency}}

\begin{remark}[Justification of clause (ii) in the NTK Gaussian regime]\label{rem:basu_justification}
In the regime of \cref{thm:aea_sample_efficiency} ($\epsilon_i \overset{\text{iid}}{\sim}\mathcal{N}(0,\sigma^2)$), the ensemble mean $\mu_{\overline{\mathcal{Q}}}$ is a complete sufficient statistic for the location parameter while $\overline{\delta}$ is location-invariant and hence ancillary. \Cref{thm:basu} (Basu) then implies $\mu_{\overline{\mathcal{Q}}}\perp\!\!\!\perp\overline{\delta}$, and the same reasoning extends to $N$ as a location functional of the next-state ensemble, yielding clause~(ii). For finite-width networks this is approximate, as flagged by the theory-to-practice gap discussion at the end of \cref{sec:theory}.
\end{remark}

We define the ensemble mean $\mu_{\mathcal{Q}}(s,a) = \frac{1}{N} \sum_{i=1}^N Q_i(s,a)$ and the approximation bias $B(s,a) = \mu_{\mathcal{Q}}(s,a) - Q^\pi(s,a)$. The AEA target construction is given by $y_{\overline{\kappa}} = \mu_{\mathcal{Q}} + \overline{\kappa} \delta$. Taking the expectation with respect to the ensemble distribution:
\begin{align*}
\mathbb{E}_{\mathcal{Q}}[y_{\overline{\kappa}}] &= \mathbb{E}_{\mathcal{Q}}[\mu_{\mathcal{Q}} + \overline{\kappa} \delta] \\
&= \mathbb{E}_{\mathcal{Q}}[Q^\pi + B + \overline{\kappa} \delta] \\ 
&= Q^\pi + \mathbb{E}_{\mathcal{Q}}[B] + \overline{\kappa} \mathbb{E}_{\mathcal{Q}}[\delta].
\end{align*}
\paragraph{Identification of the unbiased equilibrium.}
Setting $\mathbb{E}_{\mathcal{Q}}[y_{\overline{\kappa}^*}] = Q^\pi$ in the linear identity above and solving for $\overline{\kappa}$ yields the unique scalar that makes $y_{\overline{\kappa}}$ unbiased for $Q^\pi$:
\begin{align} \label{eq:kappa_star_proof}
\overline{\kappa}^* = -\frac{\mathbb{E}_{\mathcal{Q}}[B]}{\mathbb{E}_{\mathcal{Q}}[\overline{\delta}]}.
\end{align}
This identification is a population statement about the linear estimator $\mu_{\mathcal{Q}}+\overline{\kappa}\,\overline{\delta}$ alone: it does \emph{not} involve a ratio of random variables and does \emph{not} reference the actor parameter $\kappa$.

\paragraph{The stochastic approximation locates $\overline{\kappa}^*$.}
By \cref{thm:aea_convergence}, $\overline{\kappa}_t$ converges to a stationary point of $J(\overline{\kappa})$, characterized by $\mathbb{E}_{\mathcal{D}}[\text{sign}(\widetilde{Q}_{\kappa}-y_{\overline{\kappa}^*})]=0$. Because $\overline{\delta}(s',a')>0$ almost surely and $\text{sign}$ is positively homogeneous, this is equivalent to the median condition
\begin{align}\label{eq:median_cond}
\text{med}(u) \;=\; \gamma\,\overline{\kappa}^*,
\end{align}
with $u$ as in~\eqref{eq:condsym}. Under \cref{assump:symmetry}(i), $u$ is conditionally symmetric, hence $\text{med}(u)=\mathbb{E}[u]$; under \cref{assump:symmetry}(ii) (the Basu-type ancillarity, justified in \cref{rem:basu_justification} via \cref{thm:basu}), $\mathbb{E}[u] = \mathbb{E}[N]/\mathbb{E}[\overline{\delta}]$. Together with the projected Bellman fixed-point relation $\mathcal{Q}^* = \Pi_{\mathcal{D}}\mathcal{T}^\pi_{\overline{\kappa}^*}\mathcal{Q}^*$ from \cref{thm:aea_convergence} (which under stationarity of $d^\pi$ couples $\mathbb{E}[\mu_{\mathcal{Q}^*}]$ to $\overline{\kappa}^*\mathbb{E}[\overline{\delta}]$) and the actor's symmetric stationarity~\eqref{eq:kappa} (which drives $\kappa$ to a root of the same sign expectation, eliminating the $\kappa\,\mathbb{E}[\delta]$ contribution in $\mathbb{E}[N]$ at the joint equilibrium), the algorithmic fixed-point condition~\eqref{eq:median_cond} reduces to~\eqref{eq:kappa_star_proof}.

If $\overline{\kappa}^* \in \mathcal{K}_{\text{stable}}$, the process converges to this root almost surely, ensuring the bias-cancelling property.

Let $e_i = Q_i - Q^\pi$ be the error of the $i$th ensemble member. In the NTK regime, we assume $e_i$ are independent sub-Gaussian random variables with parameter $\sigma^2$. The target error is:
\begin{align*}
\mathcal{E} = y_{\overline{\kappa}^*} - Q^\pi = \left( \frac{1}{N} \sum_{i=1}^N e_i \right) + \overline{\kappa}^* \delta.
\end{align*}
Since the mean absolute difference (MAD) $\delta$ is a $U$-statistic of the ensemble errors, $y_{\overline{\kappa}^*}$ is a linear combination of sub-Gaussian variables. By \cref{thm:subgaussian_ustat} there exists a constant $C > 0$ such that the concentration of the sum follows:
\begin{align*}
P\left( |y_{\overline{\kappa}^*} - Q^\pi| \geq \epsilon \right) \leq 2 \exp\left( -\frac{N \epsilon^2}{2 C \sigma^2} \right). 
\end{align*}
As $N \to \infty$, the RHS vanishes for any $\epsilon > 0$, implying $y_{\overline{\kappa}^*} \xrightarrow{p} Q^\pi$.
The update rule for $\overline{\kappa}$ follows the projected ODE:
\begin{align*}
\dot{\overline{\kappa}} = \Pi_{\mathcal{K}_{\text{stable}}} [ -\nabla_{\overline{\kappa}} J(\overline{\kappa}) ].
\end{align*}
If the minimizer $\overline{\kappa}^*$ defined in \eqref{eq:kappa_star_proof} lies outside $\mathcal{K}_{\text{stable}}$, the projection onto the convex set $\mathcal{K}_{\text{stable}}$ ensures the parameter saturates at the boundary $\partial \mathcal{K}_{\text{stable}}$. Let $\overline{\kappa}_{\text{bound}} = \text{sgn}(\overline{\kappa}^*) \frac{1-\gamma}{\gamma L}$. The residual bias is then:
\begin{align*}
\text{Bias}_{\text{res}} = \mathbb{E}[B] + \overline{\kappa}_{\text{bound}} \mathbb{E}[\delta].
\end{align*}
Since $\overline{\kappa}_{\text{bound}} \in \mathcal{K}_{\text{stable}}$, the Lipschitz condition $|\overline{\kappa} L| < \frac{1-\gamma}{\gamma}$ holds, ensuring the spectral radius $\rho(T_{\overline{\kappa}}^\pi) \leq \gamma(1 + |\overline{\kappa}|L) < 1$. Thus, the system maintains convergent value propagation while minimizing the objective $J(\overline{\kappa})$ subject to the stability constraint. \hfill $\square$

\subsection{Proof of Theorem \ref{aea:improvement}}
We combine the soft policy improvement lemma of \citet{haarnoja2018soft} with an error-propagation argument for approximate policy iteration \citep{kakade2002approximately}. Throughout, $V^\pi_{\text{soft}}(s) = \mathbb{E}_{a \sim \pi}[Q^\pi(s,a) - \alpha\log\pi(a|s)]$ denotes the soft value function and ${A^\pi_{\text{soft}}(s,a) = Q^\pi(s,a) - \alpha\log\pi(a|s) - V^\pi_{\text{soft}}(s)}$ the soft advantage.

Let $\epsilon(s,a) := \widetilde{Q}_{\kappa}(s,a) - Q^{\pi_k}(s,a)$. By the triangle inequality,
\begin{align*}
\mathbb{E}_{(s,a)\sim d^{\pi_k}}[|\epsilon(s,a)|] \;\leq\; \mathbb{E}_{(s,a)\sim d^{\pi_k}}\!\bigl[|\widetilde{Q}_{\kappa} - \Pi_{\mathcal{D}} Q^{\pi_k}|\bigr] + \mathbb{E}_{(s,a)\sim d^{\pi_k}}\!\bigl[|\Pi_{\mathcal{D}} Q^{\pi_k} - Q^{\pi_k}|\bigr].
\end{align*}
By \cref{aea:consistency}, the first term is bounded by $\epsilon_{\text{est}}$ at the ensemble equilibrium, and the second term equals $\epsilon_{\text{approx}}$ by definition. Thus $\mathbb{E}_{d^{\pi_k}}[|\epsilon|] \leq \epsilon_{\text{approx}} + \epsilon_{\text{est}}$.

 The performance difference lemma for the soft objective states that for any two policies $\pi, \pi'$,
\begin{align} \label{eq:soft_perf_diff}
V^{\pi'}_{\text{soft}}(s) - V^{\pi}_{\text{soft}}(s) = \frac{1}{1-\gamma} \mathbb{E}_{s' \sim d^{\pi'}}\!\left[\mathbb{E}_{a \sim \pi'(\cdot|s')}[A^\pi_{\text{soft}}(s',a)]\right].
\end{align}
Equation~\eqref{eq:soft_perf_diff} implies $V^{\pi_{k+1}}_{\text{soft}} \geq V^{\pi_k}_{\text{soft}}$ for $d^{\pi_{k+1}}$-almost every $s$ whenever
\begin{align} \label{eq:soft_gap_true}
\mathcal{J}_{\text{true}}(\pi_{k+1}) - \mathcal{J}_{\text{true}}(\pi_k) \;\geq\; 0, \quad \mathcal{J}_{\text{true}}(\pi) := \mathbb{E}_{s \sim d^{\pi_k},\, a \sim \pi(\cdot|s)}\!\left[Q^{\pi_k}(s,a) - \alpha\log\pi(a|s)\right].
\end{align}

Since $\mathcal{J}(\pi)$ replaces $Q^{\pi_k}$ with $\widetilde{Q}_{\kappa}$, for every policy $\pi$ with bounded importance ratio relative to $\pi_k$,
\begin{align*}
|\mathcal{J}(\pi) - \mathcal{J}_{\text{true}}(\pi)| \;=\; \left|\mathbb{E}_{s \sim d^{\pi_k},\, a \sim \pi(\cdot|s)}[\epsilon(s,a)]\right| \;\leq\; \mathbb{E}_{(s,a)\sim d^{\pi_k}}[|\epsilon(s,a)|] \;\leq\; \epsilon_{\text{approx}} + \epsilon_{\text{est}}.
\end{align*}
The performance difference lemma~\eqref{eq:soft_perf_diff} evaluates expectations under $d^{\pi_{k+1}}$, while $\mathcal{J}_{\text{true}}$ is taken under $d^{\pi_k}$. Under the assumption $\|d^{\pi_{k+1}} - d^{\pi_k}\|_{\mathrm{TV}} \leq \varepsilon_{\mathrm{TV}}$ together with the uniform bound $R_{\max}$ on the soft residual, the standard Pinsker/total-variation comparison gives
\begin{align*}
\bigl|\mathbb{E}_{s\sim d^{\pi_{k+1}}}[f(s)] - \mathbb{E}_{s\sim d^{\pi_k}}[f(s)]\bigr| \;\leq\; R_{\max}\,\varepsilon_{\mathrm{TV}}\quad\text{for any $|f|\leq R_{\max}$.}
\end{align*}
Combining with the previous bound,
\begin{align*}
\mathcal{J}_{\text{true}}(\pi_{k+1}) - \mathcal{J}_{\text{true}}(\pi_k) \;\geq\; \mathcal{J}(\pi_{k+1}) - \mathcal{J}(\pi_k) - 2(\epsilon_{\text{approx}} + \epsilon_{\text{est}}) - 2R_{\max}\varepsilon_{\mathrm{TV}}.
\end{align*}
Using \eqref{eq:soft_perf_diff} with its horizon factor, $V^{\pi_{k+1}}_{\text{soft}} \geq V^{\pi_k}_{\text{soft}}$ holds for $d^{\pi_{k+1}}$-almost every $s$ whenever
\begin{align*}
\mathcal{J}(\pi_{k+1}) - \mathcal{J}(\pi_k) \;>\; \frac{2\gamma}{1-\gamma}\bigl(\epsilon_{\text{approx}} + \epsilon_{\text{est}} + R_{\max}\,\varepsilon_{\mathrm{TV}}\bigr),
\end{align*}
which completes the proof. \hfill $\square$

\subsection{Proof of Theorem \ref{aea:shrinkage}}

To prove the rate of shrinkage, we analyze the error of the aggregated estimator $y_{\overline{\kappa}} = \mu_{\mathcal{Q}} + \overline{\kappa} \delta$ at the two-timescale equilibrium.

The total estimation error $\epsilon_{\text{est}}$ can be expressed as the distance between the target at equilibrium and the true $Q$-function. Let $\mathcal{E} = y_{\overline{\kappa}^*} - Q^\pi$. We decompose $\mathcal{E}$ as:
\begin{align*}
\mathcal{E} = \underbrace{\mathbb{E}_{\mathcal{Q}}[y_{\overline{\kappa}^*} - Q^\pi]}_{\text{Bias}} + \underbrace{(y_{\overline{\kappa}^*} - \mathbb{E}_{\mathcal{Q}}[y_{\overline{\kappa}^*]})}_{\text{Concentration Deviation}}. 
\end{align*}
Applying the triangle inequality and taking expectation under $(s,a)\sim d^\pi$:
\begin{align} \label{eq:shrinkage_decomposition}
\epsilon_{\text{est}} \;=\; \mathbb{E}_{d^\pi}[|\mathcal{E}|] \;\leq\; \bigl|\mathbb{E}_{\mathcal{Q}}[B] + \overline{\kappa}^* \mathbb{E}_{\mathcal{Q}}[\overline\delta]\bigr| \;+\; \mathbb{E}_{d^\pi}\!\bigl[|\text{dev}(y_{\overline{\kappa}^*})|\bigr].
\end{align}
From the convergence results of Theorem~\ref{thm:aea_convergence}, the scalar $\overline{\kappa}$ tracks the minimizer of $J(\overline{\kappa})$.
\begin{itemize}
 \item {\bf Case 1:}  $\overline{\kappa}^*_{\text{unconstrained}} \in \mathcal{K}_{\text{stable}}$. The equilibrium point is exactly $\overline{\kappa}^* = -\mathbb{E}[B]/\mathbb{E}[\delta]$. Substituting this into the first term of \eqref{eq:shrinkage_decomposition}:    
    $\mathbb{E}[B] + \left( -\frac{\mathbb{E}[B]}{\mathbb{E}[\delta]} \right) \mathbb{E}[\delta] = 0.$
    The residual bias vanishes in expectation.
\item {\bf Case 2:}  $\overline{\kappa}^*_{\text{unconstrained}} \notin \mathcal{K}_{\text{stable}}$. The process converges to the boundary $\Pi_{\mathcal{K}_{\text{stable}}}(\overline{\kappa}^*)$, minimizing the residual $\ell_1$ norm subject to the spectral radius constraint.
\end{itemize}

The deviation term $\text{dev}(y_{\overline{\kappa}^*}) = (\mu_{\mathcal{Q}} - \mathbb{E}[\mu_{\mathcal{Q}}]) + \overline{\kappa}^* (\overline\delta - \mathbb{E}[\overline\delta])$ is a linear combination of a sample mean and a $U$-statistic of degree 2. Since the ensemble errors are $\sigma$-sub-Gaussian, the kernel $h(x,y)=|x-y|$ is unbounded but sub-exponential, so Hoeffding's bounded-kernel inequality does not apply. Applying the Arcones-type concentration for sub-Gaussian $U$-statistics (\cref{thm:subgaussian_ustat}, specialised in \cref{cor:mad_concentration}) and combining with the sub-Gaussian sample mean gives a constant $C>0$ such that:
\begin{align*}
P\left( |y_{\overline{\kappa}^*} - \mathbb{E}[y_{\overline{\kappa}^*}]| \geq \epsilon \right) \leq 2 \exp\left( -\frac{N \epsilon^2}{2 C \sigma^2} \right).
\end{align*}
Setting the right-hand side equal to $\alpha$ and solving for $\epsilon$:
\begin{align*}
\log(\alpha/2) = -\frac{N \epsilon^2}{2 C \sigma^2} \implies \epsilon = C \sigma \sqrt{\frac{2 \log(2/\alpha)}{N}}.
\end{align*}
This yields the concentration error term $\mathcal{O}(1/\sqrt{N})$.

Combining the results from Step 2 and Step 3 into \eqref{eq:shrinkage_decomposition}, we obtain:
\begin{align*}
\epsilon_{\text{est}} \leq | \mathbb{E}[B] + \overline{\kappa}^* \mathbb{E}[\delta] | + C \sigma \sqrt{\frac{\log(2/\alpha)}{N}}.
\end{align*}
This demonstrates that under optimal aggregation within the stability region, the error is dominated by the concentration term, which vanishes as $N \to \infty$ \hfill $\square$

\subsection{Proof of Theorem \ref{thm:mse_comparison}}
To establish the efficiency of the AEA target, we decompose its Mean Squared Error (MSE) and analyze the asymptotic behavior of its components relative to the REDQ baseline.

Let $Q^\pi$ be the true value, $\mu_{\mathcal{Q}}$ the ensemble mean, and $\delta$ the disagreement. We define the estimation error $e = y_{\overline{\kappa}} - Q^\pi$. Expanding $e$ around the expectations of the ensemble statistics:
\begin{align*}
e &= (\mu_{\mathcal{Q}} + \overline{\kappa} \delta) - Q^\pi \\
&= (\mu_{\mathcal{Q}} - \mathbb{E}[\mu_{\mathcal{Q}}]) + \overline{\kappa}(\delta - \mathbb{E}[\delta]) + (\mathbb{E}[\mu_{\mathcal{Q}}] + \overline{\kappa} \mathbb{E}[\delta] - Q^\pi).
\end{align*}
Let $\mathbb{E}[B] = \mathbb{E}[\mu_{\mathcal{Q}}] - Q^\pi$ be the systematic approximation bias, $\Delta\mu_{\mathcal{Q}} = \mu_{\mathcal{Q}} - \mathbb{E}[\mu_{\mathcal{Q}}]$, and ${\Delta\delta = \delta - \mathbb{E}[\delta]}$. The error becomes:
$$e = \Delta\mu_{\mathcal{Q}} + \overline{\kappa} \Delta\delta + (\mathbb{E}[B] + \overline{\kappa} \mathbb{E}[\delta]).$$
The MSE is defined as $\mathbb{E}[e^2]$. Since $\mathbb{E}[\Delta\mu_{\mathcal{Q}}] = 0$ and $\mathbb{E}[\Delta\delta] = 0$, the cross-terms between the stochastic fluctuations and the deterministic bias vanish in expectation:
\begin{align*}
\text{MSE}(y_{\overline{\kappa}}) &= \mathbb{E}[(\Delta\mu_{\mathcal{Q}} + \overline{\kappa} \Delta\delta)^2] + (\mathbb{E}[B] + \overline{\kappa} \mathbb{E}[\delta])^2 \\
&= \mathbb{E}[\Delta\mu_{\mathcal{Q}}^2] + \overline{\kappa}^2 \mathbb{E}[\Delta\delta^2] + 2\overline{\kappa} \mathbb{E}[\Delta\mu_{\mathcal{Q}} \Delta\delta] + (\mathbb{E}[B] + \overline{\kappa} \mathbb{E}[\delta])^2 \\
&= \underbrace{\text{Var}(\mu_{\mathcal{Q}}) + \overline{\kappa}^2 \text{Var}(\delta) + 2\overline{\kappa} \text{Cov}(\mu_{\mathcal{Q}}, \delta)}_{\text{Variance Component } \mathcal{V}_N(\overline{\kappa})} + \underbrace{(\mathbb{E}[B] + \overline{\kappa} \mathbb{E}[\delta])^2}_{\text{Bias Component } \mathcal{B}(\overline{\kappa})}.
\end{align*}

The two-timescale stochastic approximation of $\overline{\kappa}$ converges to the equilibrium point $\overline{\kappa}^*$ where the gradient of the risk function $J(\overline{\kappa})$ vanishes. As established in Theorem \ref{aea:shrinkage}, for $\overline{\kappa}^* \in \mathcal{K}_{\text{stable}}$, this corresponds to the root of the bias component $\mathcal{B}(\overline{\kappa})$. Setting $\mathcal{B}(\overline{\kappa}^*) = 0$ yields:
$$\overline{\kappa}^* = -\frac{\mathbb{E}[B]}{\mathbb{E}[\delta]}.$$
Substituting $\overline{\kappa}^*$ into the MSE expression:
$$\text{MSE}(y_{\overline{\kappa}^*}) = \mathcal{V}_N(\overline{\kappa}^*) + 0 = \text{Var}(\mu_{\mathcal{Q}}) + (\overline{\kappa}^*)^2 \text{Var}(\delta) + 2\overline{\kappa}^* \text{Cov}(\mu_{\mathcal{Q}}, \delta).$$
At this optimal point, the systematic bias is eliminated, and the MSE is purely a function of the ensemble variance.

We compare the MSE of AEA to the REDQ target $y_{\text{redq}} = \min_{j \in \mathcal{M}} \{Q_j\}$ with $|\mathcal{M}|=2$. By the properties of sub-Gaussian averages and $U$-statistics of degree 2:
$$\text{Var}(\mu_{\mathcal{Q}}) = \frac{\sigma^2}{N}, \quad \text{Var}(\delta) \leq \frac{C_1 \sigma^2}{N}, \quad \text{Cov}(\mu_{\mathcal{Q}}, \delta) \leq \frac{C_2 \sigma^2}{N}.$$
Thus, $\text{MSE}(y_{\overline{\kappa}^*}) = \mathcal{O}\left(\frac{\sigma^2}{N}\right)$. For the REDQ target, the variance is determined by the order statistics of a fixed-size subset:
$$\text{MSE}(y_{\text{redq}}) = \text{Var}(\min\{Q_1, Q_2\}) + (\mathbb{E}[B] - \text{Bias}_{\text{min}})^2.$$
For sub-Gaussian noise, $\text{Var}(\min\{Q_1, Q_2\}) = C \sigma^2$, where $C > 0$ is a constant independent of the total ensemble size $N$. Consequently:
$$\lim_{N \to \infty} \frac{\text{MSE}(y_{\overline{\kappa}^*})}{\text{MSE}(y_{\text{redq}})} = \lim_{N \to \infty} \frac{\mathcal{O}(1/N)}{\Omega(1)} = 0.$$
There exists an $N_0 \in \mathbb{N}$ such that for all $N > N_0$, $\text{MSE}(y_{\overline{\kappa}^*}) < \text{MSE}(y_{\text{redq}})$.

In the limit $N \to \infty$, the variance component $\mathcal{V}_N(\overline{\kappa}^*) \to 0$. Since the bias component is zero at equilibrium, we have $\lim_{N \to \infty} \text{MSE}(y_{\overline{\kappa}^*}) = 0$, proving asymptotic exactness. \hfill $\square$

\subsection{Proof of Theorem \ref{thm:aea_sample_efficiency}}
To establish the informational and variance optimality of AEA, we analyze the estimator in the Gaussian regime $\epsilon_i \overset{i.i.d.}{\sim} \mathcal{N}(0, \sigma^2)$, where $Q_i \sim \mathcal{N}(\mu, \sigma^2)$ with $\mu = Q^\pi + \mathbb{E}[B]$. The variance of the AEA target $y_{\overline{\kappa}^*} = \mu_{\mathcal{Q}} + \overline{\kappa}^* \delta$ is given by:
$\text{Var}(y_{\overline{\kappa}^*}) = \text{Var}(\mu_{\mathcal{Q}}) + (\overline{\kappa}^*)^2 \text{Var}(\delta) + 2\overline{\kappa}^* \text{Cov}(\mu_{\mathcal{Q}}, \delta).$
Since the ensemble is Gaussian, the sample mean $\mu_{\mathcal{Q}}$ is a complete sufficient statistic for the location parameter $\mu$. The average pairwise deviation $\delta = \frac{1}{\binom{N}{2}} \sum_{i>j} |Q_i - Q_j|$ is a location-invariant statistic. According to \cref{thm:basu}, any location-invariant statistic is independent of the complete sufficient statistic for a location family. Consequently, $\mu_{\mathcal{Q}}$ and $\delta$ are independent, and $\text{Cov}(\mu_{\mathcal{Q}}, \delta) = 0$.

For Gaussian $Q_i$, we have $\text{Var}(\mu_{\mathcal{Q}}) = \frac{\sigma^2}{N}$. The variance of the Gini mean difference (GMD) is known to scale as $\text{Var}(\delta) = \frac{\sigma^2}{N} C_{\delta} + \mathcal{O}(N^{-2})$, where $C_{\delta} = \frac{2\pi - 4)}{\pi} \approx 1.$ for large $N$. Substituting these into the variance expression:
\begin{align*}
\text{Var}(y_{\overline{\kappa}^*}) = \frac{\sigma^2}{N} + (\overline{\kappa}^*)^2 \frac{\sigma^2}{N} C_{\delta} = \frac{\sigma^2}{N} \left( 1 + (\overline{\kappa}^*)^2 C_{\delta} \right).
\end{align*}

Since $\text{Var}(y_{\overline{\kappa}^*}) = \mathcal{O}(1/N)$, while the REDQ variance for $M=2$ is $\text{Var}(y_{\text{redq}}) = \sigma^2(1 - \frac{1}{\pi}) = \Theta(1)$, there exists a finite $N_0$ such that for all $N > N_0$ the AEA target is strictly more efficient.

By the information processing inequality, the Fisher information $\mathcal{I}$ of a function of a data subset cannot exceed the information content of the subset itself. Since the REDQ target $y_{\text{redq}}$ is a function of a subset $\mathcal{M} \subset \mathcal{Q}$ with $|\mathcal{M}| = M$, we have $\mathcal{I}(y_{\text{redq}}) \leq \mathcal{I}(\mathcal{M}) = \mathcal{O}(M)$. As $M$ is a fixed hyperparameter, REDQ's per-update informational throughput is $\mathcal O(M) = \mathcal O(1)$ with respect to $N$.

The AEA target $y_{\overline{\kappa}^*}$ is a function of the ensemble-wide statistics $(\mu_{\mathcal{Q}}, \delta)$. Because $\mu_{\mathcal{Q}}$ is a sufficient statistic for the parameter $Q^\pi$ in the Gaussian limit, AEA preserves the full scaling of the ensemble’s Fisher information: $\mathcal{I}(y_{\overline{\kappa}^*}) = \Omega(N)$. According to the Cramér–Rao lower bound (CRLB), the variance of any estimator $\widehat{Q}$ is bounded by $\text{Var}(\widehat{Q}) \geq \mathcal{I}(Q^\pi)^{-1}$. It follows that $\text{CRLB}_{\text{AEA}} = \mathcal{O}(1/N)$ and $\text{CRLB}_{\text{REDQ}} = \mathcal{O}(1/M)$. While REDQ is limited by the subset size $M$, AEA approaches the theoretical limit of information utilization for the entire ensemble. \hfill $\square$

\section{Experimental Setup, Hyperparameters, and Implementation Details} \label{appendix:experimental_details}

Our experiments are implemented in \texttt{PyTorch} \citep[Version 2.1.0]. We conduct experiments on continuous control tasks using the \texttt{v5}-versions of the MuJoCo physics engine \citep{todorov2012mujoco,brockman2016openai,towers2024gymnasium}, and the DeepMind Control Suite (DMC) \citep{tassa2018deepmind}.

\begin{table}
  \caption{Model architectures and hyperparameters shared across all baselines.}
  \label{tab:hyperparameters}
  \centering\small
  \begin{tabular}{@{}ll@{}}
    \toprule
    Hyperparameter & Value \\
    \midrule
    \multicolumn{2}{@{}l}{\textit{Network architecture (actor and critics)}} \\
    \hspace{1em}Hidden layers & $2$ \\
    \hspace{1em}Layer size & $256$ \\
    \hspace{1em}Activation function & CReLU \\
    \midrule
    \multicolumn{2}{@{}l}{\textit{Optimization}} \\
    \hspace{1em}Batch size & $256$ \\
    \hspace{1em}Optimizer & Adam \citep{kingma2014adam} \\
    \hspace{1em}Learning rate & $3 \cdot 10^{-4}$ \\
    \midrule
    \multicolumn{2}{@{}l}{\textit{General training configuration}} \\
    \hspace{1em}Replay buffer size & $1 \cdot 10^6$ \\
    \hspace{1em}Random initial steps & $10{,}000$ \\
    \hspace{1em}Discount factor $(\gamma)$ & $0.99$ \\
    \hspace{1em}Target smoothing $(\tau)$ & $5 \cdot 10^{-3}$ \\
    \hspace{1em}Initial entropy $(\alpha)$ & $0.2$ \\
    \hspace{1em}Target entropy $(\mathcal{H}_{\text{target}})$ & $-\text{dim}(\mathcal{A})/2$ \\
    \hspace{1em}Seeds & $\{1, 2, \dots, 10\}$ \\
    \bottomrule
  \end{tabular}
\end{table}

The hyperparameters used in our experiments are summarized in \cref{tab:hyperparameters}. These configurations follow the original implementations of SAC \citep{haarnoja2018soft} and REDQ \citep{chen2021randomized}. 
We adopt CReLU activations following \citet{nauman2024overestimation}, who demonstrated that standard ReLU networks in deep RL suffer from progressive loss of plasticity \citep{abbas2023loss}: as training proceeds, an increasing fraction of neurons become permanently inactive, reducing the network's effective capacity. CReLU \citep{shang2016understanding} mitigates this by concatenating both the positive and negative parts of each pre-activation, ensuring that gradient information flows through all neurons regardless of the sign of the input. This property is particularly relevant for ensemble architectures, where N independent critics must each maintain sufficient representational capacity throughout training.
All methods use automatic entropy-temperature tuning of $\alpha$, which tunes $\alpha$ to match a target entropy based on the action dimensionality \citep{haarnoja2018softa}; see \cref{tab:hyperparameters} for specific values.

We initialize the aggregation parameters as $\overline{\kappa} = -0.8$ and $\kappa = 0.0$; an ablation of initialization sensitivity is provided in \cref{appendix:ablations}. The step sizes for the two scalar updates are set to $\eta_{\kappa}=\eta_{\overline \kappa}=0.1$ and are the same across all tasks and both learning regimes. To ensure stable optimization in an unconstrained parameter space, we map both parameters onto the open interval from minus one to one. In practice, the upper bound is rarely approached (e.g., see parameter trajectories in \cref{fig:kappas_trajectory:interactive:mujoco,fig:kappas_trajectory:interactive:dmc,fig:kappas_trajectory:sample_efficient:mujoco,fig:kappas_trajectory:sample_efficient:dmc}); the constraint primarily serves as a safeguard against extreme $Q$ values (e.g., the spikes visible in \cref{fig:dis_trajectory:interactive:mujoco,fig:dis_trajectory:interactive:dmc,fig:dis_trajectory:sample_efficient:mujoco,fig:dis_trajectory:sample_efficient:dmc}).

Across all learning regimes, we use the same network architecture, optimizer, and training configurations to ensure comparability. As SAC, REDQ, and AEA are matched in the number of critics and the UTD ratio, their computational costs are comparable. Most of the compute in these methods comes from forward and backward passes through the actor and critic networks, and matching the number of networks and gradient updates per data point keeps this workload similar. AEA adds only a negligible overhead over a standard REDQ implementation because it learns two extra scalars, $\kappa$ and $\overline{\kappa}$, and their gradients (see . 
We retain the twin-critic configuration ($N = 2$) for DSAC and DRND because these methods quantify value uncertainty through the learned return distribution rather than through ensemble disagreement; scaling them to $N = 10$ would effectively convert them into hybrid ensemble–distributional estimators and conflate the two uncertainty-quantification paradigms we aim to contrast.

All experiments were run on a workstation equipped with a single NVIDIA GeForce RTX 4090 GPU. Running a single seed took about 90-120 minutes of wallclock time. 

We provide a pseudo-code summary of the algorithm in \cref{alg:AEA}.

\begin{algorithm*}

\caption{Adaptive Ensemble Aggregation (AEA)}

\label{alg:AEA}

\begin{algorithmic}[1]

\State \textbf{Initialize:} Replay buffer $\mathcal{D}$; critic ensemble $\{Q_{\theta_i}\}_{i=1}^N$, actor $\pi_\phi$, and targets $\{\overline{Q}_{\overline{\theta}_i}\}_{i=1}^N$; aggregation scalars $\overline{\kappa}, \kappa$

\For{each environment interaction}

\State Take action $a_t \sim \pi_\phi$, observe $r_t$ and $s_{t+1}$, and add $(s_t, a_t, r_t, s_{t+1})$ to $\mathcal{D}$

\For{$G$ steps (update-to-data ratio)}

\State Sample mini-batch $B = \{(s, a, r, s')\}$ from $\mathcal{D}$ with $a' \sim \pi_{\phi}(\cdot|s')$

\State $y_{\overline{\kappa}} \gets r + \gamma \left( \mu_{\overline{\mathcal{Q}}}(s',a') + \overline{\kappa} \cdot \overline{\delta}(s',a') \right)$ \Comment{Reference for bootstrapping}

\State $\widetilde{Q}_{\kappa} \gets \mu_{\mathcal{Q}}(s,a) + \kappa \cdot \delta(s,a)$ \Comment{Reference for policy improvement}

\State $\theta_{i} \gets \theta_{i} - \eta_{\theta} \nabla_{\theta_i} \frac{1}{|B|} \sum_{B} (Q_{\theta_i}(s,a) - y_{\overline{\kappa}})^2, \quad \forall i \in [N]$ \Comment{Critic update}

\State $\overline{\theta}_i \gets \tau \theta_i + (1-\tau)\overline{\theta}_i, \quad \forall i \in [N]$ \Comment{Target EMA update}

\EndFor

\State \textcolor{gray}{// AEA two-timescale update (once per interaction):}

\State $\text{err}_{B} \gets \text{sign}(\widetilde{Q}_{\kappa} - y_{\overline{\kappa}})$

\State $\overline{\kappa} \gets \overline{\kappa} + \eta_{\overline{\kappa}} \cdot \gamma \cdot \text{mean}(\text{err}_{B})$ \Comment{Stable value propagation (majority vote)}

\State $\kappa \gets \kappa - \eta_{\kappa} \cdot \text{mean}(\text{err}_{B})$ \Comment{Optimizing exploration landscape}

\State $\widetilde{Q}_{\kappa} \gets \mu_{\mathcal{Q}}(s,a) + \kappa \cdot \delta(s,a)$ \Comment{Recompute with updated $\kappa$}

\State $\phi \gets \phi + \eta_{\phi} \nabla_{\phi} \frac{1}{|B|} \sum_{B} \left( \widetilde{Q}_{\kappa} - \alpha \log \pi_{\phi}(a|s) \right)$ \Comment{Actor update}

\State $\alpha \gets \exp \left( \log \alpha - \eta_{\alpha} \nabla_{\log \alpha} \frac{1}{|B|} \sum_{B} \left( -\alpha \log \pi_{\phi}(a|s) - \alpha \mathcal{H}_{\text{target}} \right) \right)$ \Comment{Entropy autotune}

\EndFor

\end{algorithmic}

\end{algorithm*}

\section{Additional Experimental Results} \label{appendix:additional_results}

\begin{table}
\caption{Final return on continuous control environments for ensemble size 10 (AEA, REDQ, SAC-N) or 2 (DSAC, DRND), Update-to-Data Ratio 20, and interaction budget \num{300000}. Values are averaged over evaluation repetitions and ten seeds. $\pm$ denotes standard deviation over seeds.}
\label{tab::main_results_final_return}
\begin{center}
\adjustbox{max width=\textwidth}{%
\begin{tabular}{lccccc}
\toprule
& AEA & REDQ & SAC-$N$ & DRND & DSAC \\
\emph{MuJoCo-v5} & (ours) & \citep{chen2021randomized} & \citep{haarnoja2018soft} & \citep{yang2024exploration} & \citep{ma2025dsac}\\
\cmidrule(lr){1-6}
\texttt{Ant}         & $\mathbf{4485}\sd{\pm1044}$ & $4473\sd{\pm1142}$ & $1\sd{\pm4}$       & $819\sd{\pm494}$   & $731\sd{\pm201}$ \\
\texttt{HalfCheetah} & $\mathbf{9721}\sd{\pm817}$  & $8692\sd{\pm1380}$ & $5586\sd{\pm662}$  & $7893\sd{\pm1181}$ & $9059\sd{\pm1194}$ \\
\texttt{Hopper}      & $\mathbf{3326}\sd{\pm583}$  & $3169\sd{\pm592}$  & $1992\sd{\pm1278}$ & $3097\sd{\pm725}$  & $2680\sd{\pm681}$ \\
\texttt{Humanoid}    & $\mathbf{5365}\sd{\pm446}$  & $4718\sd{\pm1544}$ & $4903\sd{\pm1376}$ & $2726\sd{\pm1313}$ & $1051\sd{\pm225}$ \\
\texttt{Walker2d}    & $4579\sd{\pm815}$  & $4697\sd{\pm599}$  & $1581\sd{\pm1384}$ & $5053\sd{\pm416}$  & $\mathbf{5363}\sd{\pm348}$ \\[0.5em]
\emph{DMC} \\
\cmidrule(lr){1-6}
\texttt{Cheetah-run}    & $823\sd{\pm58}$   & $851\sd{\pm48}$   & $596\sd{\pm134}$  & $\mathbf{871}\sd{\pm24}$   & $746\sd{\pm30}$ \\
\texttt{Hopper-hop}     & $\mathbf{130}\sd{\pm102}$  & $105\sd{\pm113}$  & $1\sd{\pm3}$      & $9\sd{\pm19}$     & $79\sd{\pm17}$ \\
\texttt{Hopper-stand}   & $\mathbf{779}\sd{\pm333}$  & $318\sd{\pm431}$  & $5\sd{\pm16}$     & $374\sd{\pm310}$  & $273\sd{\pm70}$ \\
\texttt{Humanoid-run}   & $\mathbf{145}\sd{\pm15}$   & $117\sd{\pm43}$   & $1\sd{\pm1}$      & $1\sd{\pm0}$      & $17\sd{\pm32}$ \\
\texttt{Humanoid-stand} & $\mathbf{678}\sd{\pm145}$  & $539\sd{\pm204}$  & $52\sd{\pm117}$   & $204\sd{\pm137}$  & $271\sd{\pm183}$ \\
\texttt{Humanoid-walk}  & $\mathbf{478}\sd{\pm57}$   & $416\sd{\pm147}$  & $2\sd{\pm1}$      & $158\sd{\pm149}$  & $2\sd{\pm0}$ \\
\texttt{Quadruped-run}  & $\mathbf{823}\sd{\pm124}$  & $728\sd{\pm260}$  & $209\sd{\pm338}$  & $234\sd{\pm372}$  & $520\sd{\pm55}$ \\
\texttt{Quadruped-walk} & $\mathbf{935}\sd{\pm58}$   & $878\sd{\pm188}$  & $284\sd{\pm385}$  & $200\sd{\pm276}$  & $570\sd{\pm325}$ \\
\texttt{Walker-run}     & $\mathbf{705}\sd{\pm100}$  & $631\sd{\pm129}$  & $270\sd{\pm130}$  & $688\sd{\pm93}$   & $650\sd{\pm86}$ \\
\bottomrule

\end{tabular}}
\end{center}
\end{table}

\begin{table}
\caption{InterQuantile Mean (IQM) of the final return on continuous control environments for ensemble size 10 (AEA, REDQ, SAC-N) or 2 (DSAC, DRND), Update-to-Data Ratio 20, and interaction budget \num{300000}. Values are averaged over evaluation repetitions and ten seeds. $\pm$ denotes standard deviation over seeds.}
\label{tab::main_results_iqm}
\begin{center}
\adjustbox{max width=\textwidth}{
\begin{tabular}{lccccc}
\toprule
& AEA & REDQ & SAC-$N$ & DRND & DSAC \\
\emph{MuJoCo-v5} & (ours) & \citep{chen2021randomized} & \citep{haarnoja2018soft} & \citep{yang2024exploration} & \citep{ma2025dsac}\\
\cmidrule(lr){1-6}
\texttt{Ant} & $4744\sd{\pm260}$ & $\mathbf{4828}\sd{\pm398}$ & $1\sd{\pm1}$ &$797\sd{\pm 560}$ & $776\sd{\pm177}$\\
\texttt{HalfCheetah} & $\mathbf{9661}\sd{\pm395}$ & $8678\sd{\pm813}$ & $5425\sd{\pm191}$ & $7892\sd{\pm1180}$ & $9059\sd{\pm1191}$\\
\texttt{Hopper} & $\mathbf{3555}\sd{\pm32}$ & $3415\sd{\pm153}$ & $2063\sd{\pm1034}$ & $3015\sd{\pm840}$ & $2640\sd{\pm783}$\\
\texttt{Humanoid} & ${5376}\sd{\pm87}$ & $\mathbf{5404}\sd{\pm61}$ & $5341\sd{\pm124}$ & $2853\sd{\pm1684}$ & $908\sd{\pm171}$\\
\texttt{Walker2d} & $4652\sd{\pm267}$ & $4749\sd{\pm170}$ & $1035\sd{\pm830}$ & $5054\sd{\pm414}$ & $\mathbf{5362}\sd{\pm349}$\\[0.5em]
\emph{DMC} \\
\cmidrule(lr){1-6}
\texttt{Cheetah-run} & $827\sd{\pm9}$ & $866\sd{\pm6}$ & $635\sd{\pm26}$ & $\mathbf{873}\sd{\pm21}$ & $753\sd{\pm37}$\\
\texttt{Hopper-hop} & $\mathbf{86}\sd{\pm75}$ & $54\sd{\pm81}$ & $0\sd{\pm1}$ & $3\sd{\pm7}$ & $54\sd{\pm58}$\\
\texttt{Hopper-stand} & $\mathbf{935}\sd{\pm16}$ & $109\sd{\pm268}$ & $0\sd{\pm1}$ & $251\sd{\pm339}$ & $39\sd{\pm58}$\\
\texttt{Humanoid-run} & $\mathbf{144}\sd{\pm5}$ & $131\sd{\pm8}$ & $1\sd{\pm1}$ & $1\sd{\pm0}$ & $17\sd{\pm32}$\\
\texttt{Humanoid-stand} & $\mathbf{665}\sd{\pm55}$ & $559\sd{\pm31}$ & $9\sd{\pm1}$ & $213\sd{\pm144}$ & $284\sd{\pm190}$\\
\texttt{Humanoid-walk} & $\mathbf{484}\sd{\pm14}$ & $454\sd{\pm35}$ & $2\sd{\pm1}$ & $166\sd{\pm160}$ & $2\sd{\pm0}$\\
\texttt{Quadruped-run} & $\mathbf{856}\sd{\pm19}$ & $847\sd{\pm24}$ & $43\sd{\pm52}$ & $224\sd{\pm379}$ & $536\sd{\pm80}$\\
\texttt{Quadruped-walk} & $\mathbf{947}\sd{\pm14}$ & $932\sd{\pm16}$ & $112\sd{\pm90}$ & $184\sd{\pm285}$ & $555\sd{\pm371}$ \\
\texttt{Walker-run} & $\mathbf{741}\sd{\pm27}$ & $649\sd{\pm56}$ & $258\sd{\pm84}$ & $689\sd{\pm93}$ & $658\sd{\pm86}$\\
\bottomrule
\end{tabular}}
\end{center}
\end{table}

\begin{table}
\caption{Area Under the Learning Curve (AULC) on continuous control environments for ensemble size 10 (AEA, REDQ, SAC-N) or 2 (DSAC, DRND), Update-to-Data Ratio 20, and interaction budget \num{300000}. Values are averaged over evaluation repetitions and ten seeds. }
\label{tab::main_results_aulc}
\begin{center}
\adjustbox{max width=\textwidth}{%
\begin{tabular}{lccccc}
\toprule
& AEA & REDQ & SAC-$N$ & DRND & DSAC \\
\emph{MuJoCo-v5} & (ours) & \citep{chen2021randomized} & \citep{haarnoja2018soft} & \citep{yang2024exploration} & \citep{ma2025dsac}\\
\cmidrule(lr){1-6}
\texttt{Ant}         & $\mathbf{2343} \sd{\pm 616}$ & $2303 \sd{\pm 431}$ & $33 \sd{\pm 11}$    & $324 \sd{\pm 163}$  & $407 \sd{\pm 62}$ \\
\texttt{HalfCheetah} & $\mathbf{7214} \sd{\pm 459}$ & $6722 \sd{\pm 721}$ & $4791 \sd{\pm 440}$ & $5705 \sd{\pm 594}$ & $6869 \sd{\pm 823}$ \\
\texttt{Hopper}      & $2723 \sd{\pm 322}$ & $\mathbf{2843} \sd{\pm 232}$ & $1318 \sd{\pm 912}$ & $2544 \sd{\pm 283}$ & $2334 \sd{\pm 459}$ \\
\texttt{Humanoid}    & $2954 \sd{\pm 523}$ & $\mathbf{3031} \sd{\pm 569}$ & $2357 \sd{\pm 914}$ & $872 \sd{\pm 223}$  & $750 \sd{\pm 72}$ \\
\texttt{Walker2d}    & $2992 \sd{\pm 620}$ & $\mathbf{3389} \sd{\pm 210}$ & $975 \sd{\pm 690}$  & $2073 \sd{\pm 220}$ & $3010 \sd{\pm 59}$ \\
\emph{DMC} \\
\cmidrule(lr){1-6}
\texttt{Cheetah-run}    & $627 \sd{\pm 42}$  & $\mathbf{670} \sd{\pm 56}$  & $466 \sd{\pm 102}$ & $623 \sd{\pm 50}$  & $544 \sd{\pm 26}$ \\
\texttt{Hopper-hop}     & $\mathbf{74} \sd{\pm 62}$   & $63 \sd{\pm 47}$   & $0 \sd{\pm 0}$     & $5 \sd{\pm 11}$    & $33 \sd{\pm 11}$ \\
\texttt{Hopper-stand}   & $\mathbf{567} \sd{\pm 166}$ & $257 \sd{\pm 226}$ & $4 \sd{\pm 1}$     & $215 \sd{\pm 138}$ & $181 \sd{\pm 64}$ \\
\texttt{Humanoid-run}   & $\mathbf{80} \sd{\pm 16}$   & $57 \sd{\pm 27}$   & $1 \sd{\pm 0}$     & $1 \sd{\pm 0}$     & $2 \sd{\pm 3}$ \\
\texttt{Humanoid-stand} & $\mathbf{315} \sd{\pm 65}$  & $270 \sd{\pm 98}$  & $27 \sd{\pm 41}$   & $80 \sd{\pm 36}$   & $89 \sd{\pm 62}$ \\
\texttt{Humanoid-walk}  & $\mathbf{239} \sd{\pm 51}$  & $196 \sd{\pm 97}$  & $2 \sd{\pm 0}$     & $53 \sd{\pm 64}$   & $3 \sd{\pm 2}$ \\
\texttt{Quadruped-run}  & $\mathbf{620} \sd{\pm 148}$ & $547 \sd{\pm 170}$ & $191 \sd{\pm 60}$  & $182 \sd{\pm 138}$ & $341 \sd{\pm 121}$ \\
\texttt{Quadruped-walk} & $\mathbf{744} \sd{\pm 59}$  & $653 \sd{\pm 154}$ & $231 \sd{\pm 42}$  & $147 \sd{\pm 119}$ & $374 \sd{\pm 96}$ \\
\texttt{Walker-run}     & $\mathbf{570} \sd{\pm 80}$  & $496 \sd{\pm 113}$ & $188 \sd{\pm 75}$  & $516 \sd{\pm 84}$  & $502 \sd{\pm 67}$ \\
\bottomrule
\end{tabular}}
\end{center}
\end{table}

\begin{table}
\caption{Final return on continuous control environments for ensemble size 2, Update-to-Data Ratio 1, and interaction budget \num{1000000}. Values are averaged over evaluation repetitions and ten seeds. $\pm$ denotes standard deviation over seeds.}
\label{tab::dmc_final_return}
\begin{center}
\adjustbox{max width=\textwidth}{%
\begin{tabular}{lccccc}
\toprule
& AEA & REDQ & SAC & DRND & DSAC \\[0.5em]
\emph{MuJoCo-v5} & (ours) & \citep{chen2021randomized} & \citep{haarnoja2018soft} & \citep{yang2024exploration} & \citep{ma2025dsac}\\
\cmidrule(lr){1-6}
\texttt{Ant} & $\mathbf{5355}\sd{\pm1126}$ & $4083\sd{\pm1477}$ & $4396\sd{\pm1911}$ & $4769\sd{\pm1139}$ & $4792\sd{\pm2016}$ \\
\texttt{HalfCheetah} & $10595\sd{\pm947}$ & $11437\sd{\pm264}$ & $10947\sd{\pm630}$ & $10574\sd{\pm887}$ & $\mathbf{11574}\sd{\pm584}$ \\
\texttt{Hopper} & $\mathbf{3425}\sd{\pm379}$ & $2344\sd{\pm842}$ & $3060\sd{\pm885}$ & $3241\sd{\pm692}$ & $2034\sd{\pm837}$ \\
\texttt{Humanoid} & $4780\sd{\pm1295}$ & $4646\sd{\pm1420}$ & $4761\sd{\pm1225}$ & $5034\sd{\pm451}$ & $\mathbf{6062}\sd{\pm593}$ \\
\texttt{Walker2d} & $4270\sd{\pm1350}$ & $4182\sd{\pm1352}$ & $4396\sd{\pm245}$ & $4207\sd{\pm647}$ & $\mathbf{5207}\sd{\pm416}$\\[0.5em]
\emph{DMC} \\
\cmidrule(lr){1-6}
\texttt{Cheetah-run} & $802\sd{\pm30}$ & $\mathbf{814}\sd{\pm105}$ & $784\sd{\pm28}$ & $\mathbf{814}\sd{\pm44}$ & $655\sd{\pm99}$ \\
\texttt{Hopper-hop} & $219\sd{\pm44}$ & $\mathbf{242}\sd{\pm86}$ & $114\sd{\pm89}$ & $30\sd{\pm60}$ & $165\sd{\pm96}$ \\
\texttt{Hopper-stand} & $789\sd{\pm305}$ & $\mathbf{921}\sd{\pm99}$ & $753\sd{\pm311}$ & $781\sd{\pm192}$ & $727\sd{\pm241}$ \\
\texttt{Humanoid-run} & $159\sd{\pm36}$ & $\mathbf{180}\sd{\pm28}$ & $159\sd{\pm18}$ & $1\sd{\pm0}$ & $158\sd{\pm21}$ \\
\texttt{Humanoid-stand} & $638\sd{\pm265}$ & $728\sd{\pm156}$ & $644\sd{\pm151}$ & $661\sd{\pm141}$ & $\mathbf{893}\sd{\pm30}$ \\
\texttt{Humanoid-walk} & $517\sd{\pm92}$ & $557\sd{\pm22}$ & $564\sd{\pm89}$ & $113\sd{\pm198}$ & $\mathbf{629}\sd{\pm148}$ \\
\texttt{Quadruped-run} & $823\sd{\pm67}$ & $\mathbf{878}\sd{\pm59}$ & $852\sd{\pm88}$ & $843\sd{\pm53}$ & $792\sd{\pm30}$ \\
\texttt{Quadruped-walk} & $914\sd{\pm164}$ & $\mathbf{955}\sd{\pm36}$ & $848\sd{\pm236}$ & $936\sd{\pm41}$ & $952\sd{\pm12}$\\
\texttt{Walker-run} & $756\sd{\pm33}$ & $\mathbf{779}\sd{\pm26}$ & $735\sd{\pm49}$ & $732\sd{\pm38}$ & $710\sd{\pm57}$ \\
\bottomrule
\end{tabular}}
\end{center}
\end{table}

\begin{table}
\caption{InterQuantile Mean (IQM) of the final return on continuous control environments for ensemble size 2, Update-to-Data Ratio 1, and interaction budget \num{1000000}. Values are averaged over evaluation repetitions and ten seeds. $\pm$ denotes standard deviation over seeds.}
\label{tab::dmc_iqm}
\begin{center}
\adjustbox{max width=\textwidth}{%
\begin{tabular}{lccccc}
\toprule
& AEA & REDQ & SAC & DRND & DSAC \\[0.5em]
\emph{MuJoCo-v5} & (ours) & \citep{chen2021randomized} & \citep{haarnoja2018soft} & \citep{yang2024exploration} & \citep{ma2025dsac}\\
\cmidrule(lr){1-6}
\texttt{Ant} & $\mathbf{5708}\sd{\pm200}$ & $4595\sd{\pm451}$ & $5164\sd{\pm502}$ & $5067\sd{\pm1008}$ & $4773\sd{\pm2097}$ \\
\texttt{HalfCheetah} & $11007\sd{\pm70}$ & $11400\sd{\pm80}$ & $11074\sd{\pm138}$ & $10572\sd{\pm894}$ & $\mathbf{11572}\sd{\pm587}$ \\
\texttt{Hopper} & $\mathbf{3531}\sd{\pm22}$ & $2254\sd{\pm538}$ & $3458\sd{\pm80}$ & $3233\sd{\pm717}$ & $2053\sd{\pm869}$ \\
\texttt{Humanoid} & $5286\sd{\pm35}$ & $5271\sd{\pm360}$ & $5281\sd{\pm102}$ & $5283\sd{\pm176}$ & $\mathbf{6468}\sd{\pm228}$ \\
\texttt{Walker2d} & $4674\sd{\pm155}$ & $4378\sd{\pm246}$ & $4396\sd{\pm108}$ & $4292\sd{\pm462}$ & $\mathbf{5217}\sd{\pm406}$ \\[0.5em]
\emph{DMC} \\
\cmidrule(lr){1-6}
\texttt{Cheetah-run} & $797\sd{\pm13}$ & $\mathbf{844}\sd{\pm15}$ & $779\sd{\pm14}$ & $814\sd{\pm43}$ & $655\sd{\pm99}$ \\
\texttt{Hopper-hop} & $228\sd{\pm8}$ & $\mathbf{242}\sd{\pm17}$ & $80\sd{\pm76}$ & $26\sd{\pm61}$ & $149\sd{\pm119}$ \\
\texttt{Hopper-stand} & $925\sd{\pm16}$ & $\mathbf{940}\sd{\pm9}$ & $896\sd{\pm50}$ & $787\sd{\pm265}$ & $704\sd{\pm327}$ \\
\texttt{Humanoid-run} & $164\sd{\pm9}$ & $\mathbf{178}\sd{\pm13}$ & $159\sd{\pm7}$ & $1\sd{\pm0}$ & $161\sd{\pm20}$ \\
\texttt{Humanoid-stand} & $686\sd{\pm129}$ & $749\sd{\pm89}$ & $616\sd{\pm49}$ & $664\sd{\pm142}$ & $\mathbf{904}\sd{\pm29}$ \\
\texttt{Humanoid-walk} & $529\sd{\pm14}$ & $558\sd{\pm10}$ & $563\sd{\pm19}$ & $113\sd{\pm199}$ & $\mathbf{632}\sd{\pm150}$ \\
\texttt{Quadruped-run} & $831\sd{\pm28}$ & $\mathbf{891}\sd{\pm21}$ & $865\sd{\pm33}$ & $847\sd{\pm46}$ & $794\sd{\pm29}$ \\
\texttt{Quadruped-walk} & $945\sd{\pm15}$ & $945\sd{\pm11}$ & $941\sd{\pm17}$ & $946\sd{\pm24}$ & $\mathbf{956}\sd{\pm15}$ \\
\texttt{Walker-run} & $764\sd{\pm8}$ & $\mathbf{781}\sd{\pm11}$ & $740\sd{\pm19}$ & $732\sd{\pm38}$ & $710\sd{\pm57}$ \\
\bottomrule

\end{tabular}}
\end{center}
\end{table}

\begin{table}
\caption{Area Under the Learning Curve (AULC) on continuous control environments for ensemble size 2, Update-to-Data Ratio 1, and interaction budget \num{1000000}. Values are averaged over evaluation repetitions and ten seeds.}
\label{tab::dmc_aulc}
\begin{center}
\adjustbox{max width=\textwidth}{%
\begin{tabular}{lccccc}
\toprule
& AEA & REDQ & SAC & DRND & DSAC \\[0.5em]
\emph{MuJoCo-v5} & (ours) & \citep{chen2021randomized} & \citep{haarnoja2018soft} & \citep{yang2024exploration} & \citep{ma2025dsac}\\
\cmidrule(lr){1-6}
\texttt{Ant} & $\mathbf{3294} \sd{\pm 542}$ & $2161 \sd{\pm 445}$ & $3059 \sd{\pm 524}$ & $3010 \sd{\pm 666}$ & $2903 \sd{\pm 1020}$ \\
\texttt{HalfCheetah} & $8486 \sd{\pm 845}$ & $8852 \sd{\pm 434}$ & $8594 \sd{\pm 824}$ & $8327 \sd{\pm 740}$ & $\mathbf{9029} \sd{\pm 436}$ \\
\texttt{Hopper} & $\mathbf{2924} \sd{\pm 145}$ & $2115 \sd{\pm 360}$ & $2500 \sd{\pm 422}$ & $2589 \sd{\pm 355}$ & $1794 \sd{\pm 389}$ \\
\texttt{Humanoid} & $\mathbf{3713} \sd{\pm 286}$ & $2740 \sd{\pm 185}$ & $3363 \sd{\pm 372}$ & $3502 \sd{\pm 307}$ & $3663 \sd{\pm 328}$ \\
\texttt{Walker2d} & $3295 \sd{\pm 521}$ & $2762 \sd{\pm 395}$ & $2977 \sd{\pm 385}$ & $3046 \sd{\pm 421}$ & $\mathbf{3801} \sd{\pm 311}$ \\[0.5em]
\emph{DMC} \\
\cmidrule(lr){1-6}
\texttt{Cheetah-run} & $628 \sd{\pm 21}$ & $651 \sd{\pm 23}$ & $617 \sd{\pm 24}$ & $\mathbf{663} \sd{\pm 34}$ & $508 \sd{\pm 73}$ \\
\texttt{Hopper-hop} & $\mathbf{150} \sd{\pm 41}$ & $149 \sd{\pm 46}$ & $64 \sd{\pm 27}$ & $14 \sd{\pm 30}$ & $98 \sd{\pm 54}$ \\
\texttt{Hopper-stand} & $606 \sd{\pm 189}$ & $\mathbf{735} \sd{\pm 54}$ & $429 \sd{\pm 215}$ & $524 \sd{\pm 236}$ & $550 \sd{\pm 218}$ \\
\texttt{Humanoid-run} & $105 \sd{\pm 12}$ & $\mathbf{109} \sd{\pm 16}$ & $77 \sd{\pm 16}$ & $1 \sd{\pm 0}$ & $80 \sd{\pm 17}$ \\
\texttt{Humanoid-stand} & $372 \sd{\pm 106}$ & $383 \sd{\pm 103}$ & $312 \sd{\pm 102}$ & $291 \sd{\pm 105}$ & $\mathbf{535} \sd{\pm 88}$ \\
\texttt{Humanoid-walk} & $325 \sd{\pm 26}$ & $\mathbf{339} \sd{\pm 15}$ & $296 \sd{\pm 87}$ & $37 \sd{\pm 71}$ & $322 \sd{\pm 93}$ \\
\texttt{Quadruped-run} & $634 \sd{\pm 77}$ & $\mathbf{656} \sd{\pm 78}$ & $640 \sd{\pm 74}$ & $627 \sd{\pm 67}$ & $647 \sd{\pm 38}$ \\
\texttt{Quadruped-walk} & $782 \sd{\pm 58}$ & $761 \sd{\pm 39}$ & $718 \sd{\pm 126}$ & $664 \sd{\pm 84}$ & $\mathbf{813} \sd{\pm 31}$ \\
\texttt{Walker-run} & $641 \sd{\pm 37}$ & $\mathbf{655} \sd{\pm 40}$ & $608 \sd{\pm 49}$ & $629 \sd{\pm 40}$ & $611 \sd{\pm 65}$ \\
\bottomrule
\end{tabular}}
\end{center}
\end{table}

\paragraph{Full results in two learning regimes.} 
\Cref{tab::main_results_final_return,tab::main_results_iqm,tab::main_results_aulc,tab::dmc_final_return,tab::dmc_iqm,tab::dmc_aulc}
present three performance metrics (final return, IQM, and AULC) for all methods across both learning regimes. In the large-ensemble regime (N=10, G=20) at 300,000 interaction steps, AEA achieves the highest scores in most environments across all three metrics, with particularly pronounced gains on DMC tasks where SAC-N collapses almost entirely. The consistency across metrics confirms that AEA's advantage is not an artifact of a single evaluation criterion. In the standard twin-critic regime ($N=2, G=1$) at \num{1000000} steps, AEA performs comparably to the strongest baselines on each task. Notably, no single baseline dominates. This pattern is consistent with our theoretical prediction that AEA's primary advantage emerges when larger ensembles provide sufficient disagreement signal for the adaptive mechanism to exploit.

\begin{figure}
\centering
\includegraphics[width=0.45\linewidth]{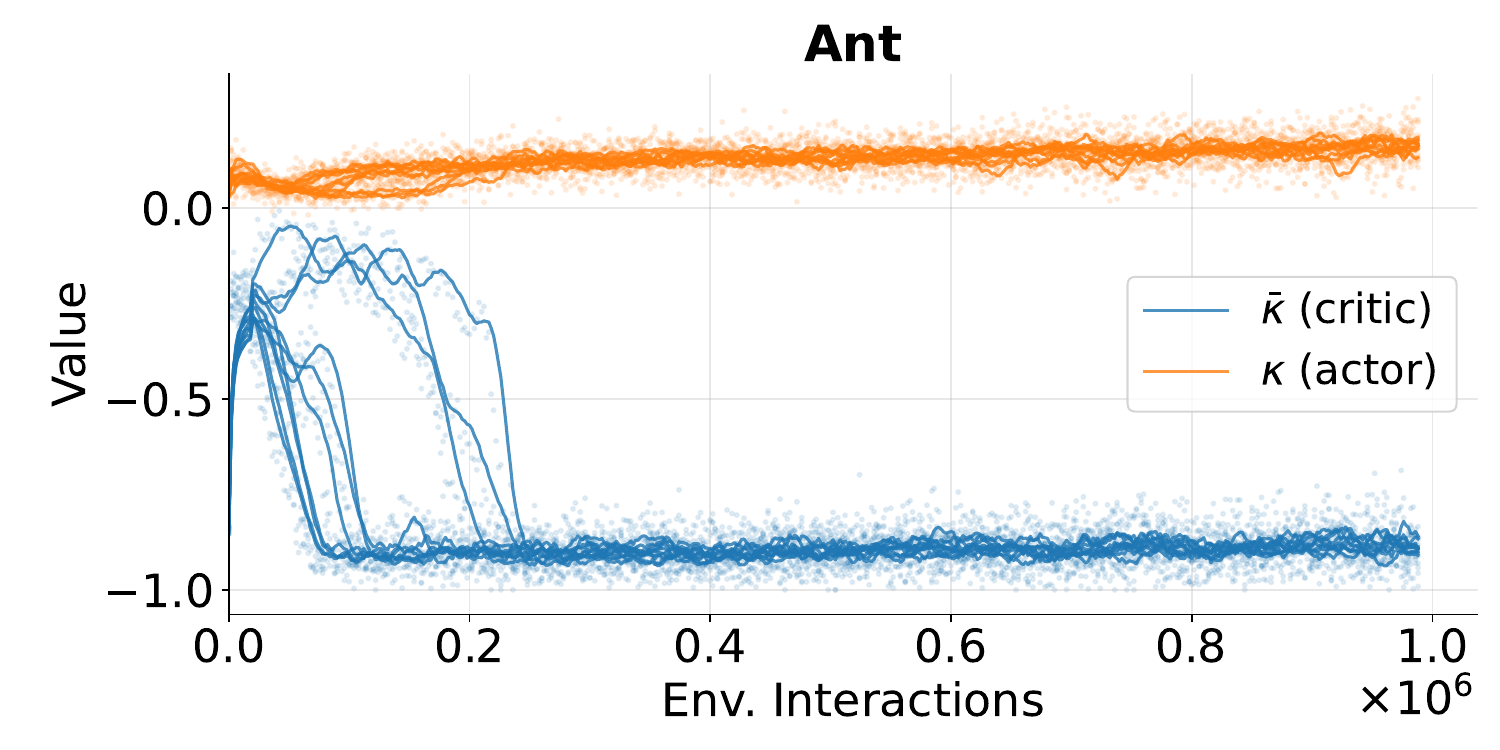}
\includegraphics[width=0.45\linewidth]{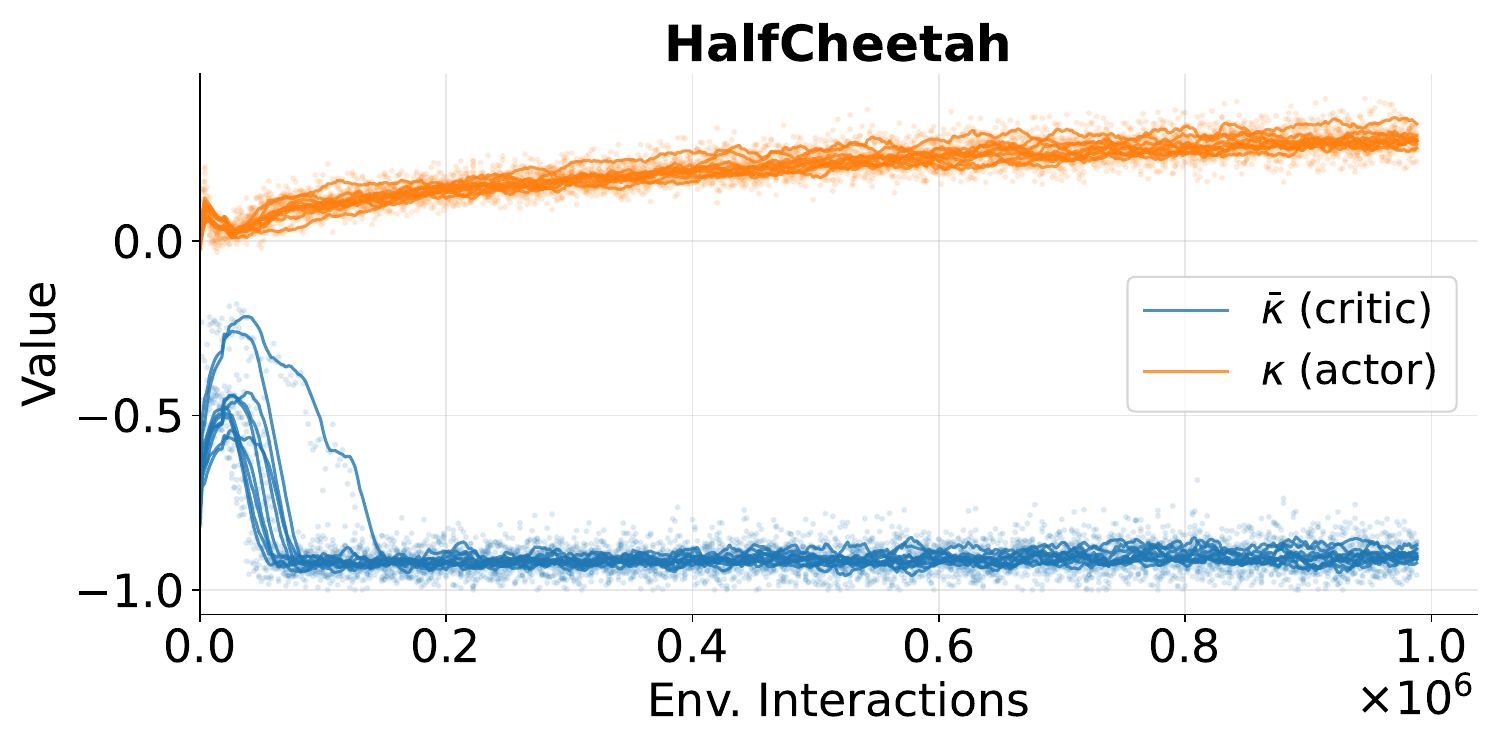}
\includegraphics[width=0.45\linewidth]{figures_final/1utd_hopper_scatter_ma.pdf}
\includegraphics[width=0.45\linewidth]{figures_final/1utd_humanoid_scatter_ma.pdf}
\includegraphics[width=0.45\linewidth]{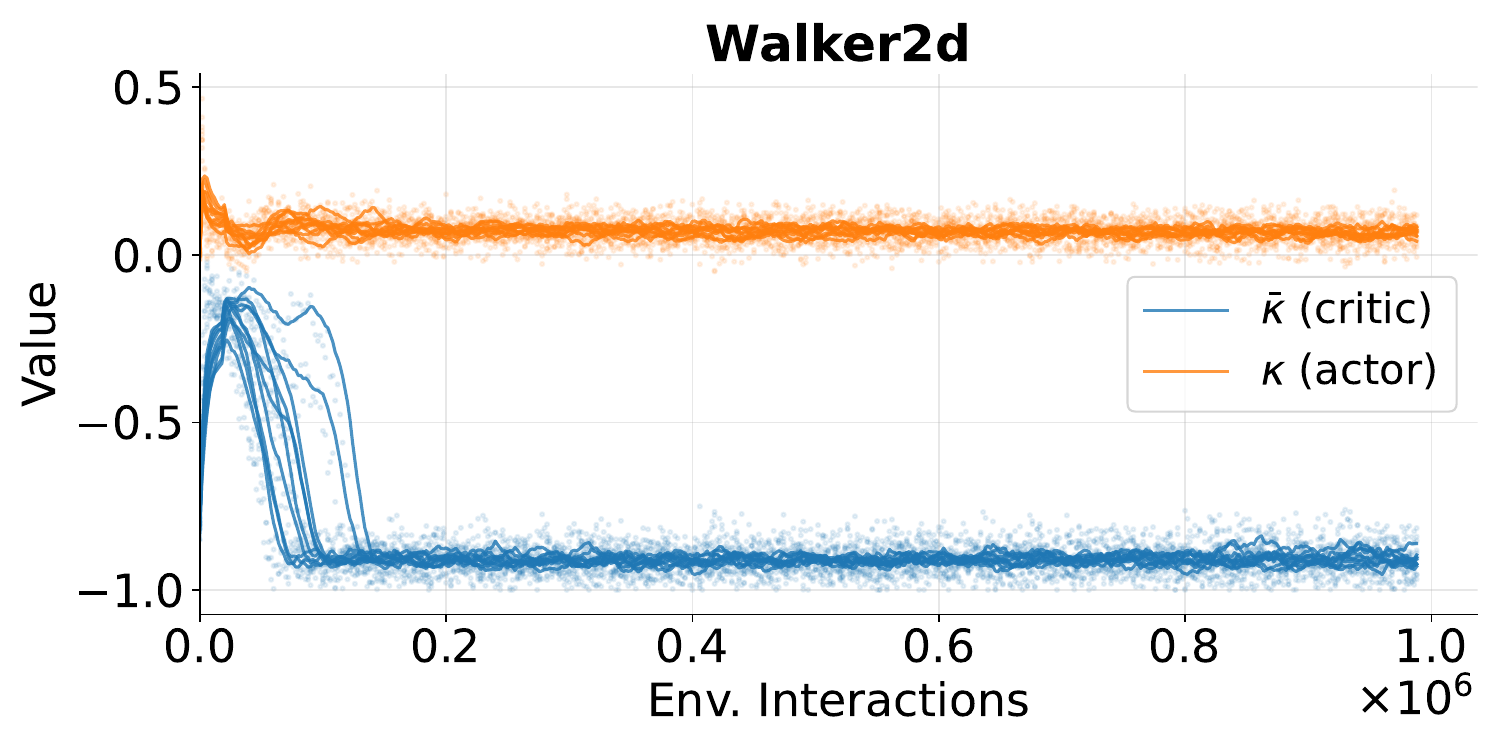}
\caption{Trajectories of the ensemble aggregation parameters $\overline{\kappa}$ (critic) and $\kappa$ (actor) across seeds in MuJoCo environments under $(N=2,G=1)$. Plotted are the raw values and moving averages per seed.}
\label{fig:kappas_trajectory:interactive:mujoco}
\end{figure}

\begin{figure}
\centering
\includegraphics[width=0.45\linewidth]{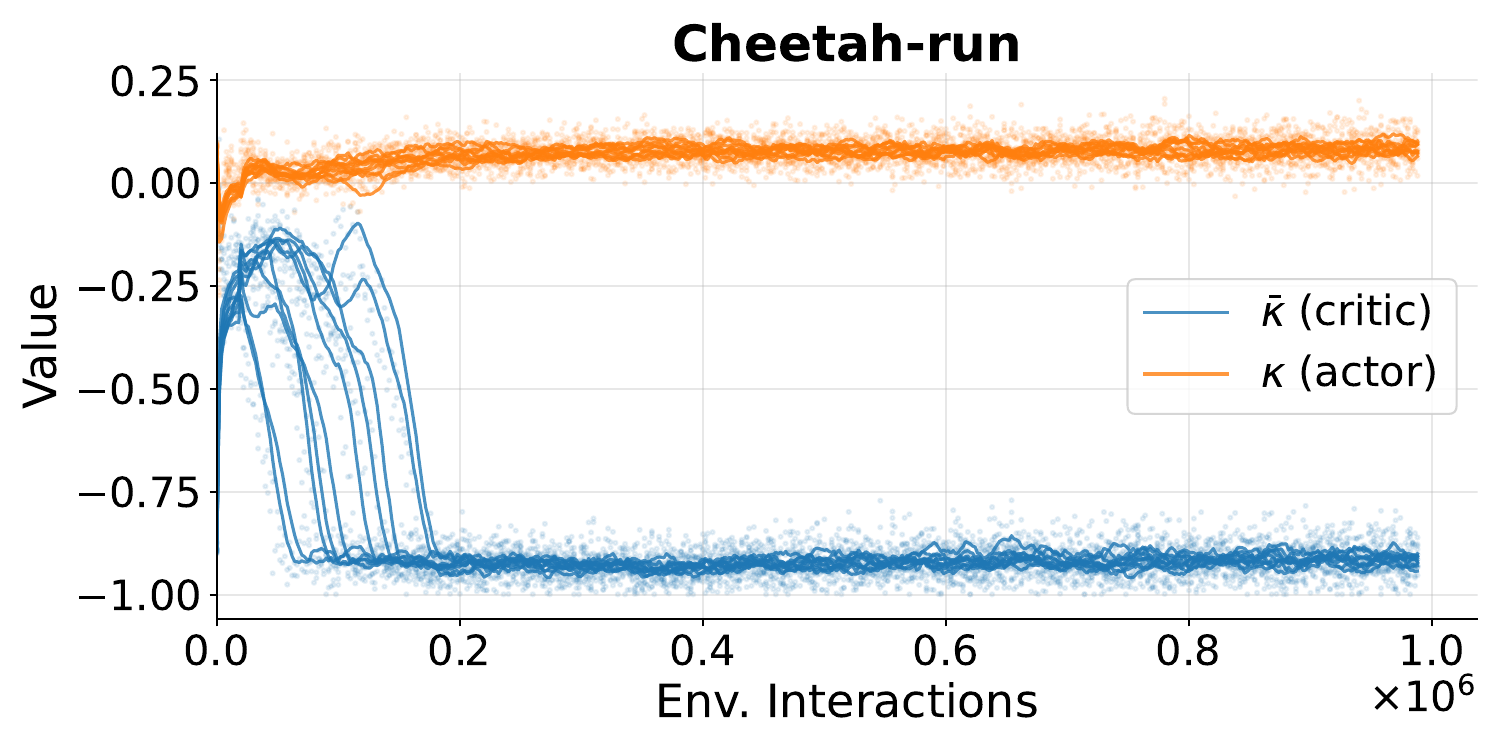}
\includegraphics[width=0.45\linewidth]{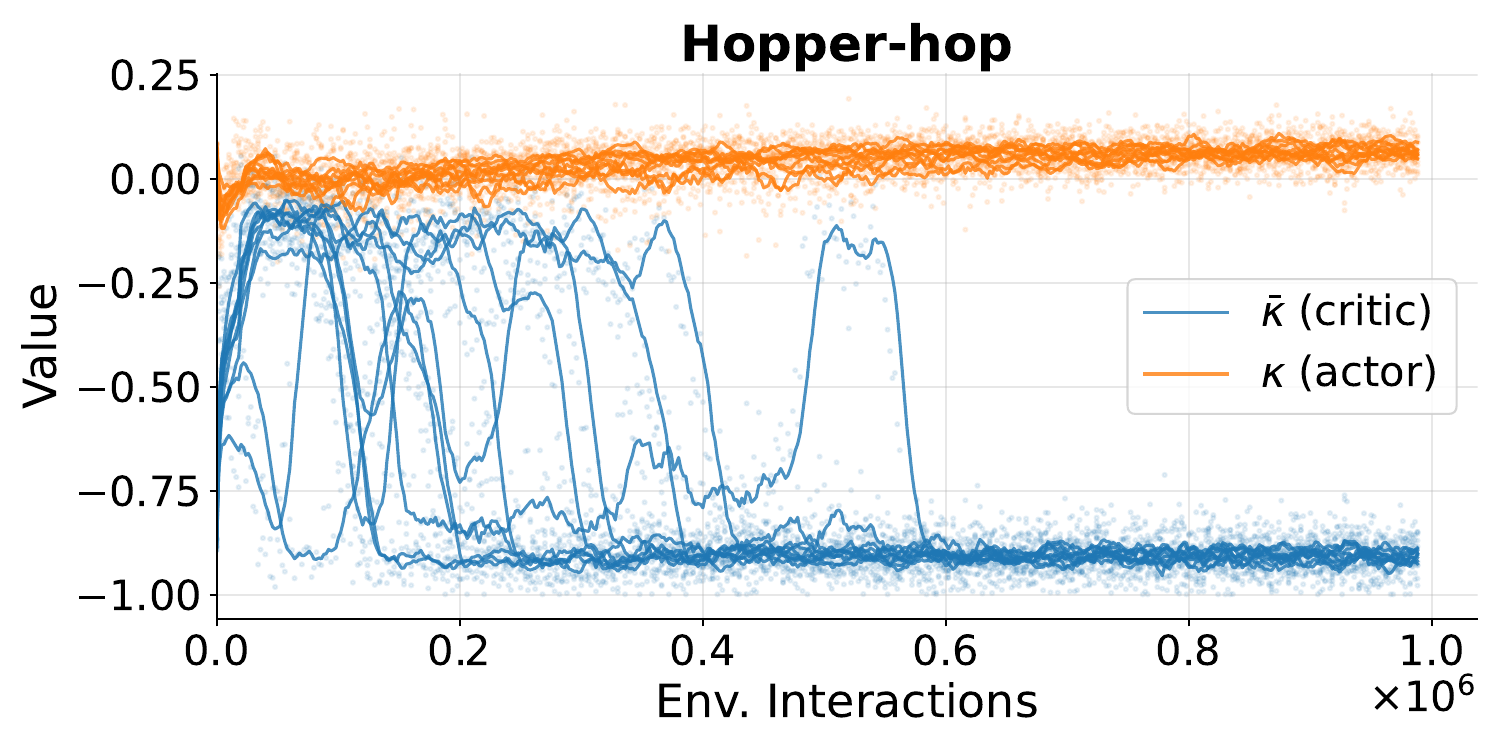}
\includegraphics[width=0.45\linewidth]{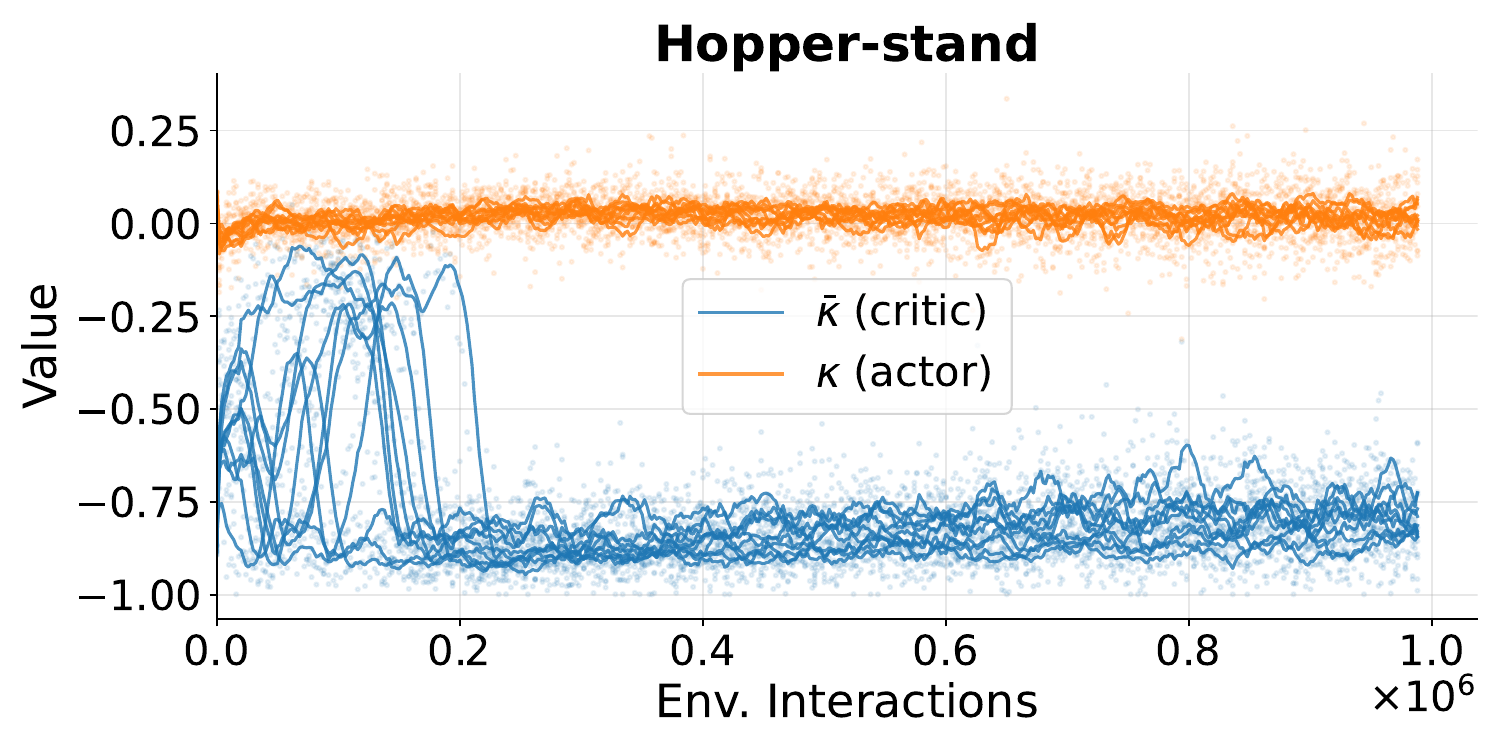}
\includegraphics[width=0.45\linewidth]{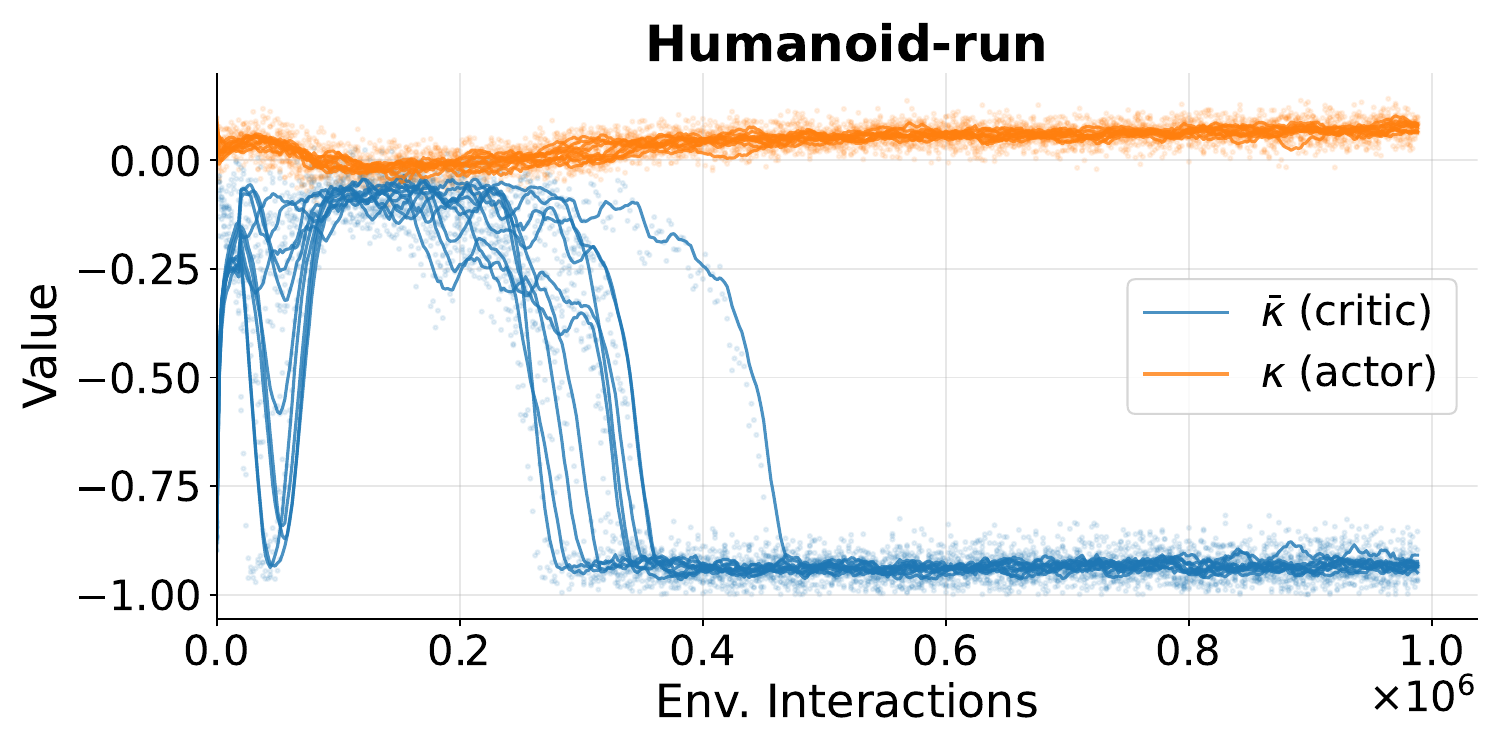}
\includegraphics[width=0.45\linewidth]{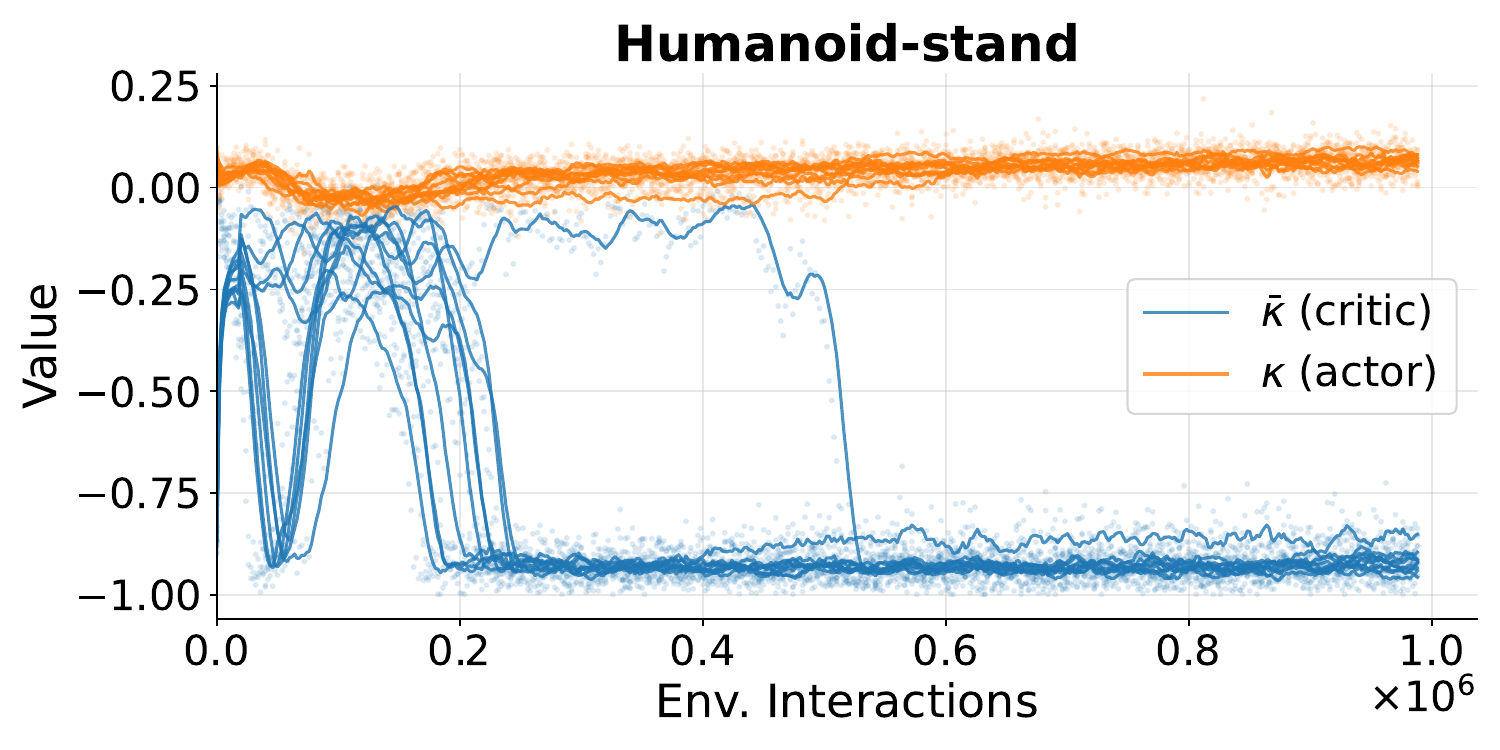}
\includegraphics[width=0.45\linewidth]{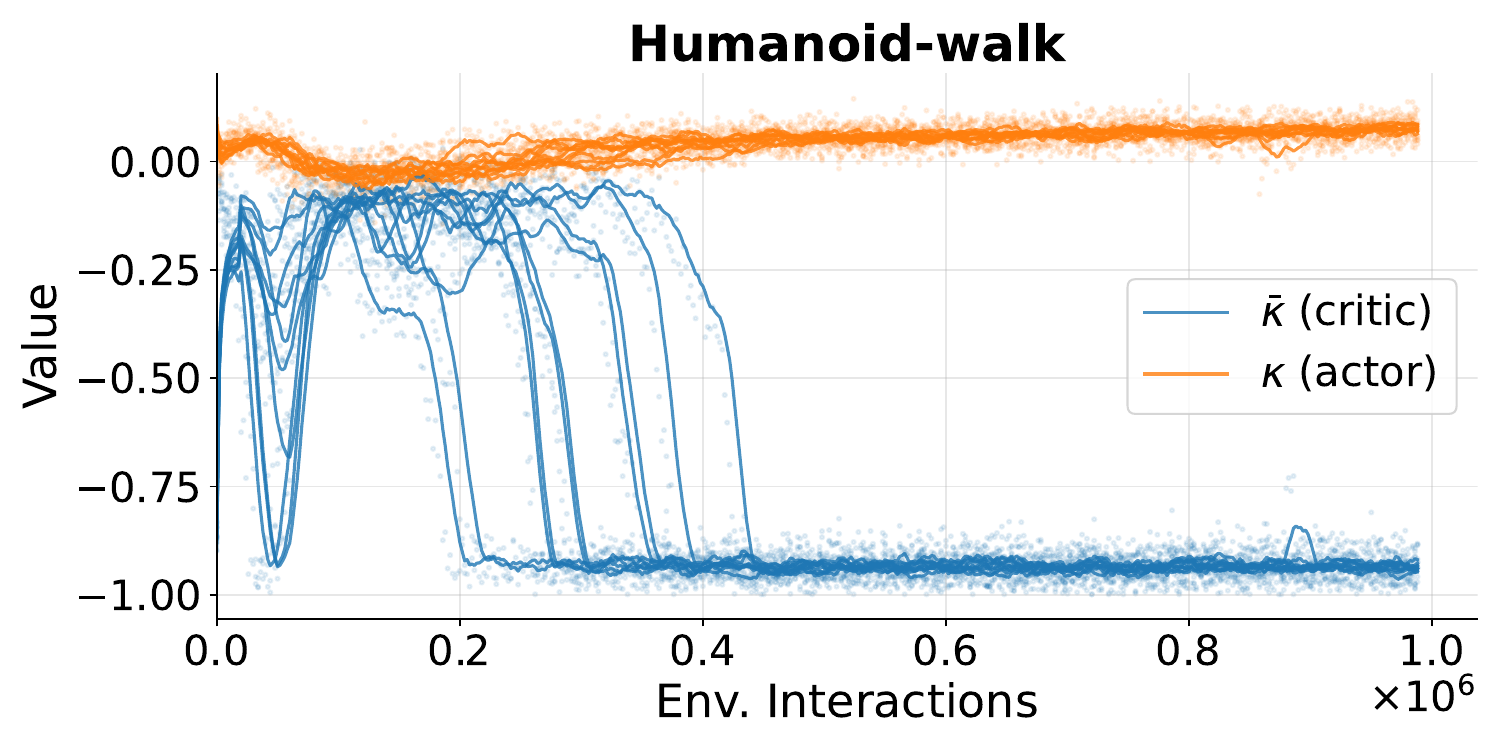}
\includegraphics[width=0.45\linewidth]{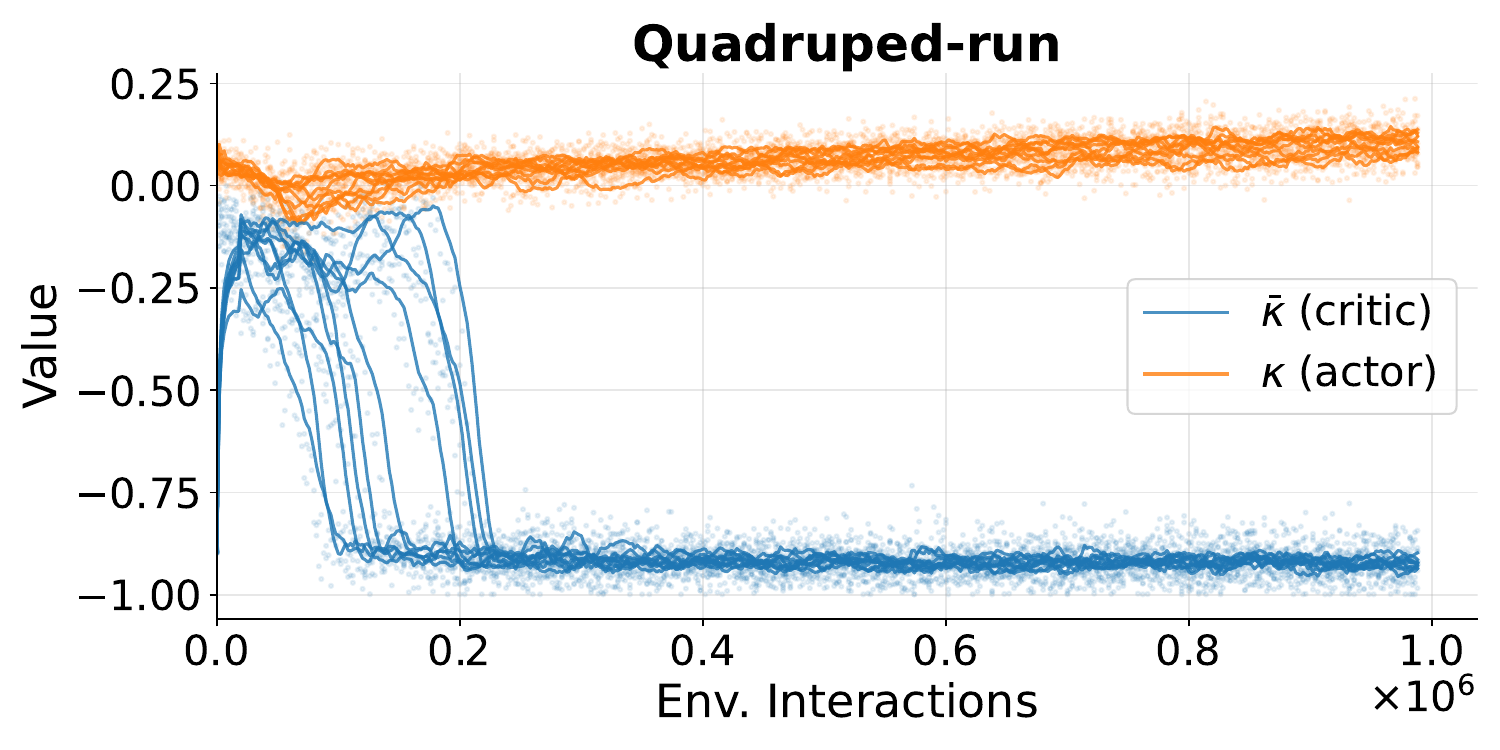}
\includegraphics[width=0.45\linewidth]{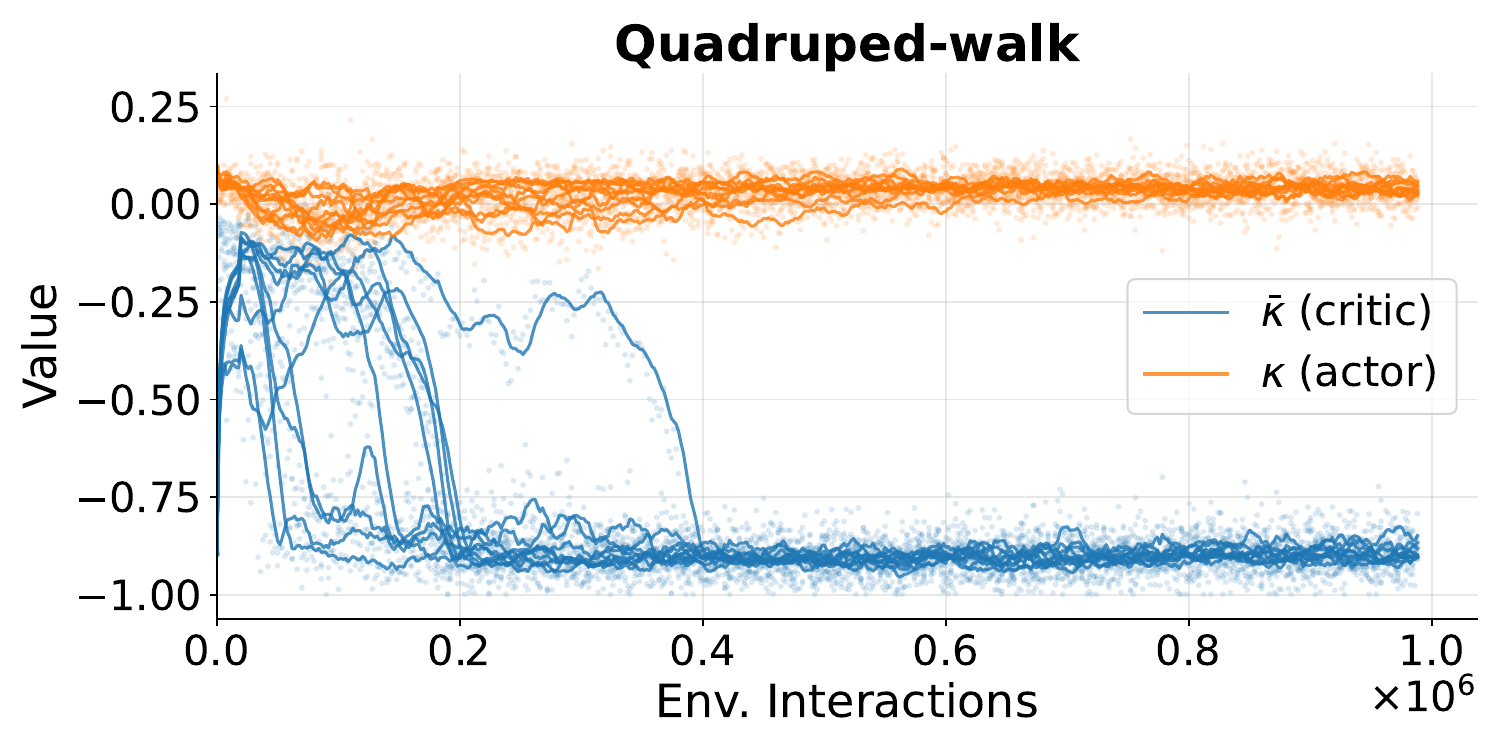}
\includegraphics[width=0.45\linewidth]{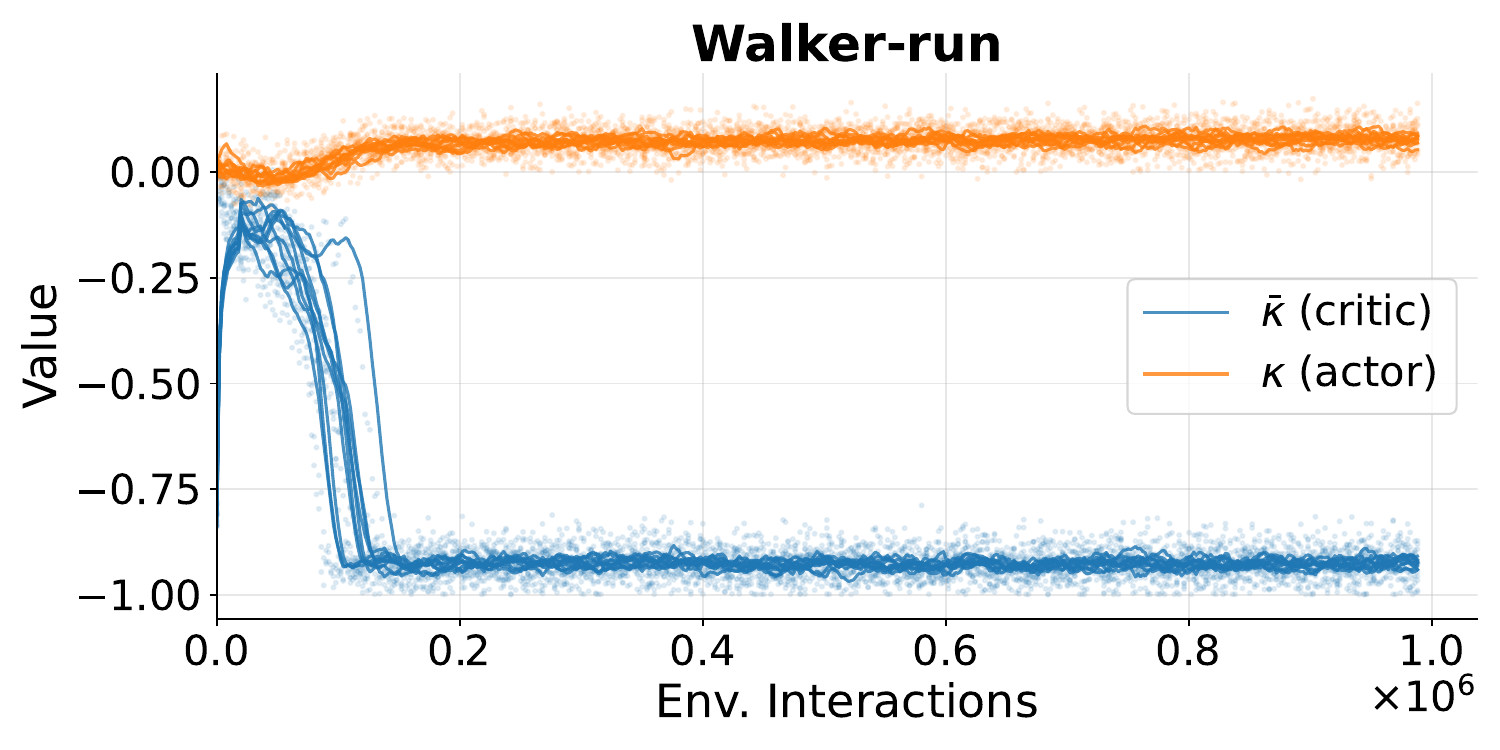}
\caption{Trajectories of the ensemble aggregation parameters $\overline{\kappa}$ (critic) and $\kappa$ (actor) across seeds in DMC environments under $(N=2, G=1)$. Plotted are the raw values and moving averages per seed.}
\label{fig:kappas_trajectory:interactive:dmc}
\end{figure}

\begin{figure}
\centering
\includegraphics[width=0.45\linewidth]{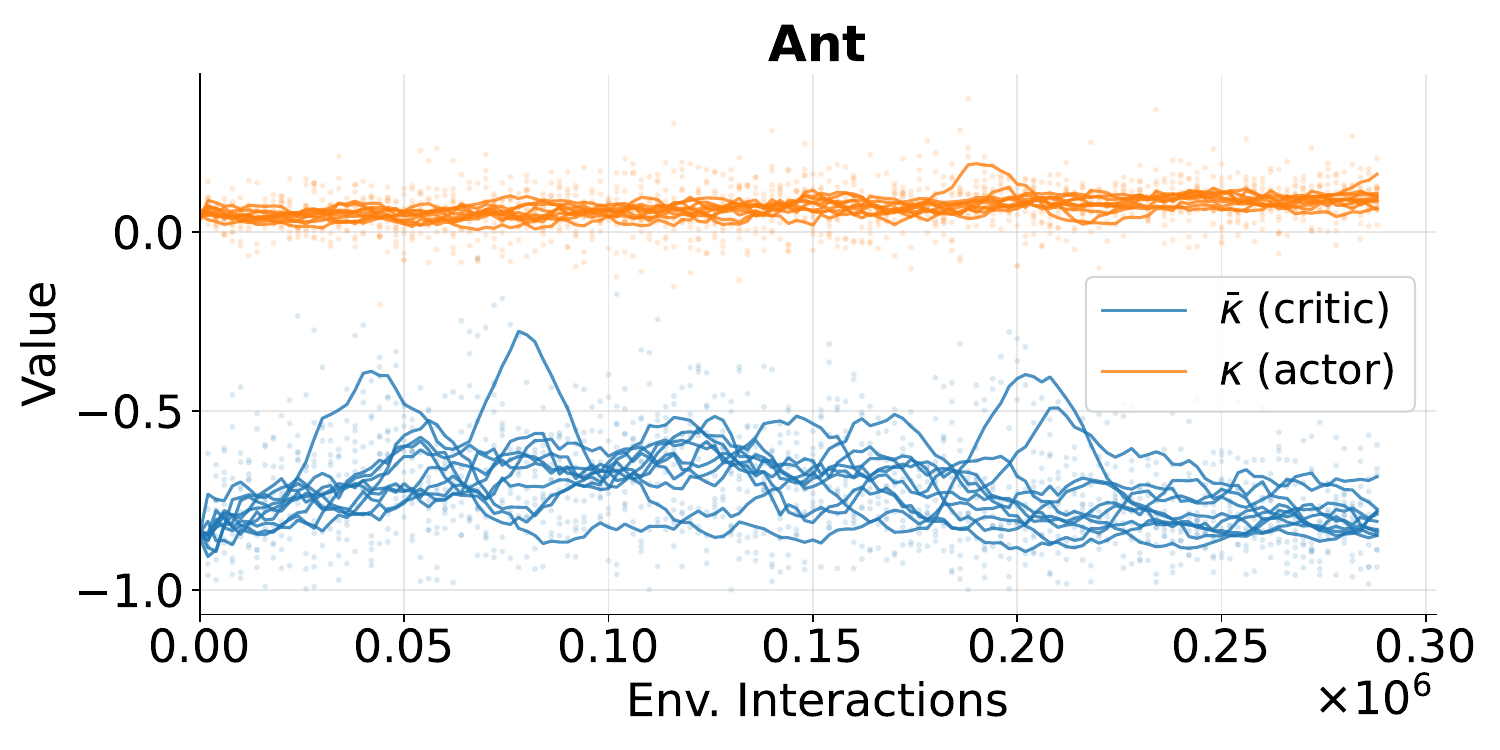}
\includegraphics[width=0.45\linewidth]{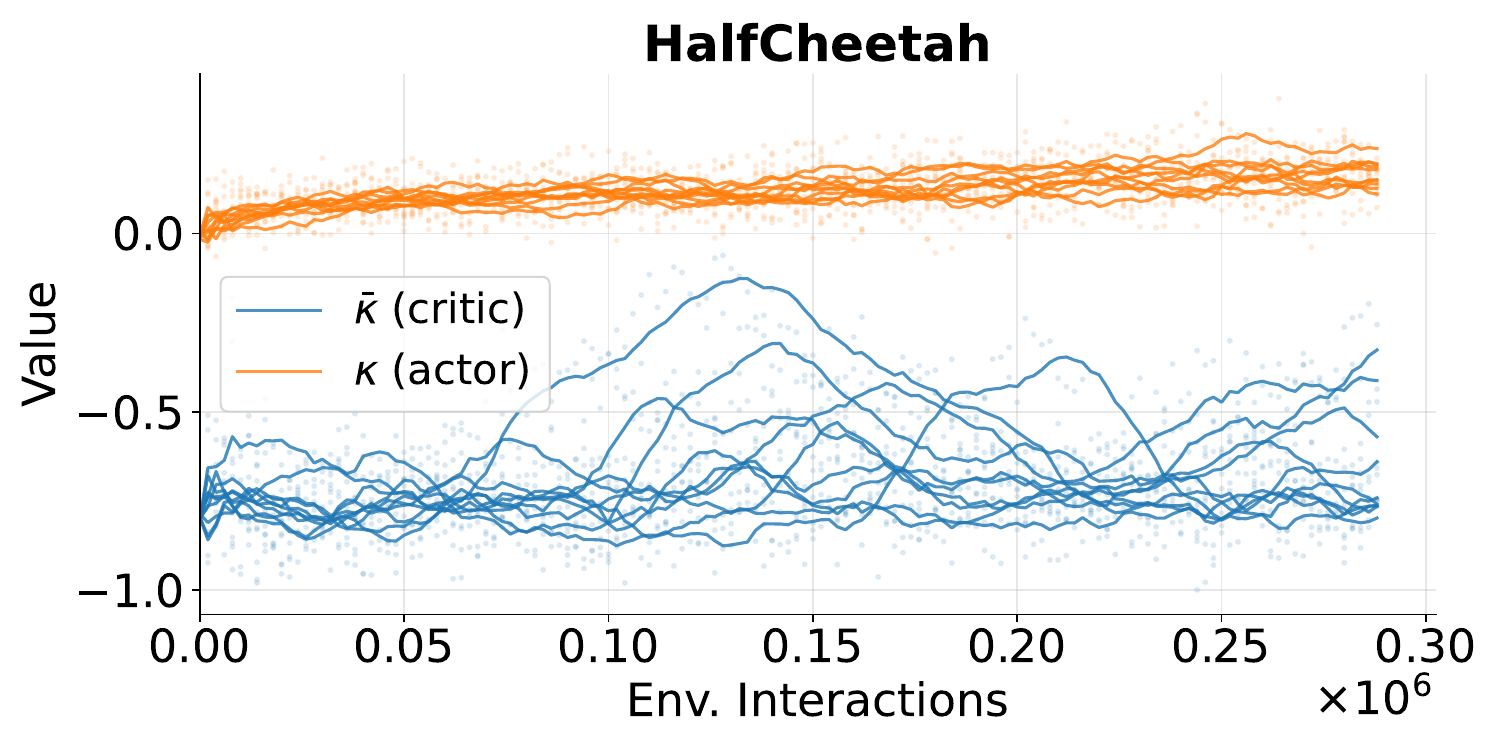}
\includegraphics[width=0.45\linewidth]{figures_final/20utd_hopper_scatter_ma.pdf}
\includegraphics[width=0.45\linewidth]{figures_final/20utd_humanoid_scatter_ma.pdf}
\includegraphics[width=0.45\linewidth]{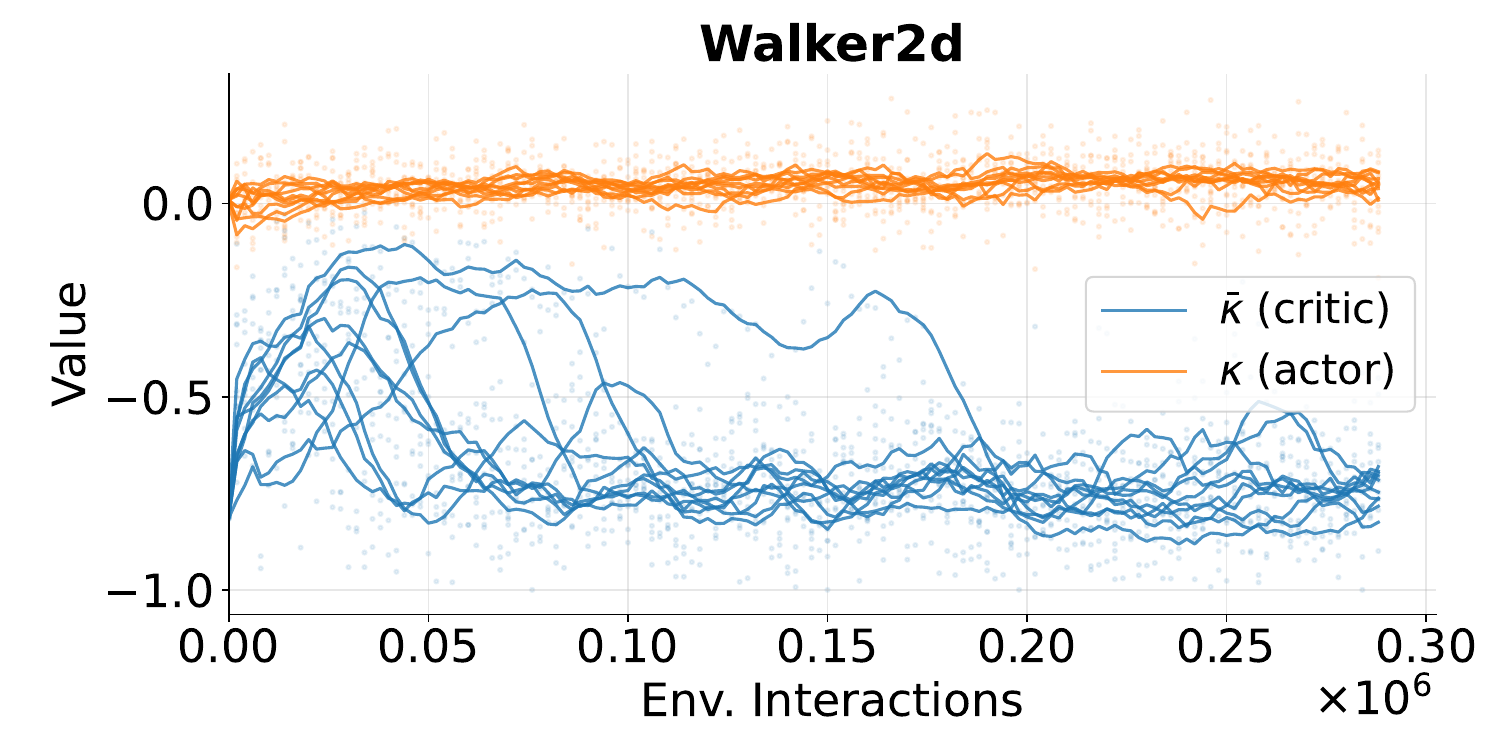}
\caption{Trajectories of the ensemble aggregation parameters $\overline{\kappa}$ (critic) and $\kappa$ (actor) across seeds in MuJoCo environments under $(N=10,G=20)$. Plotted are the raw values and moving averages per seed.}
\label{fig:kappas_trajectory:sample_efficient:mujoco}
\end{figure}

\begin{figure}
\centering
\includegraphics[width=0.45\linewidth]{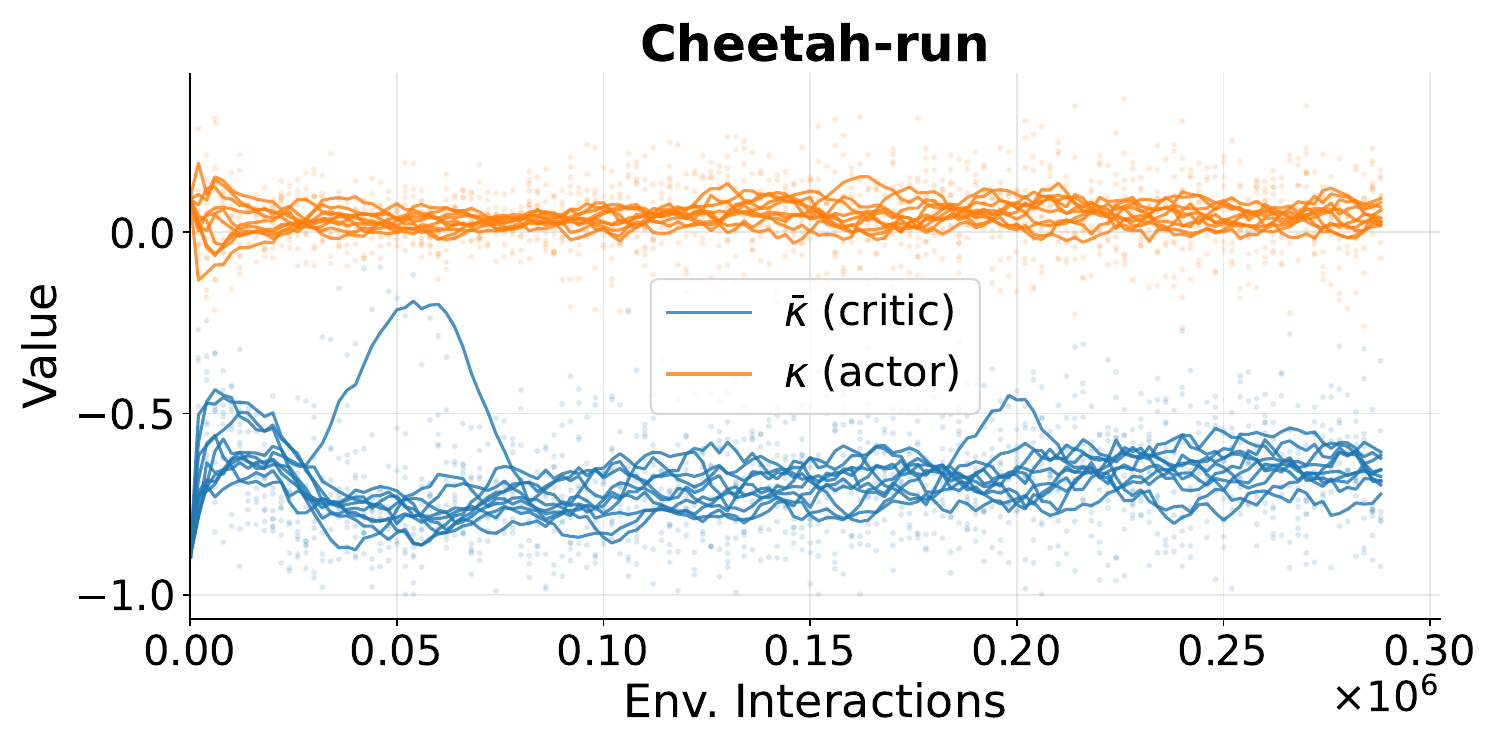}
\includegraphics[width=0.45\linewidth]{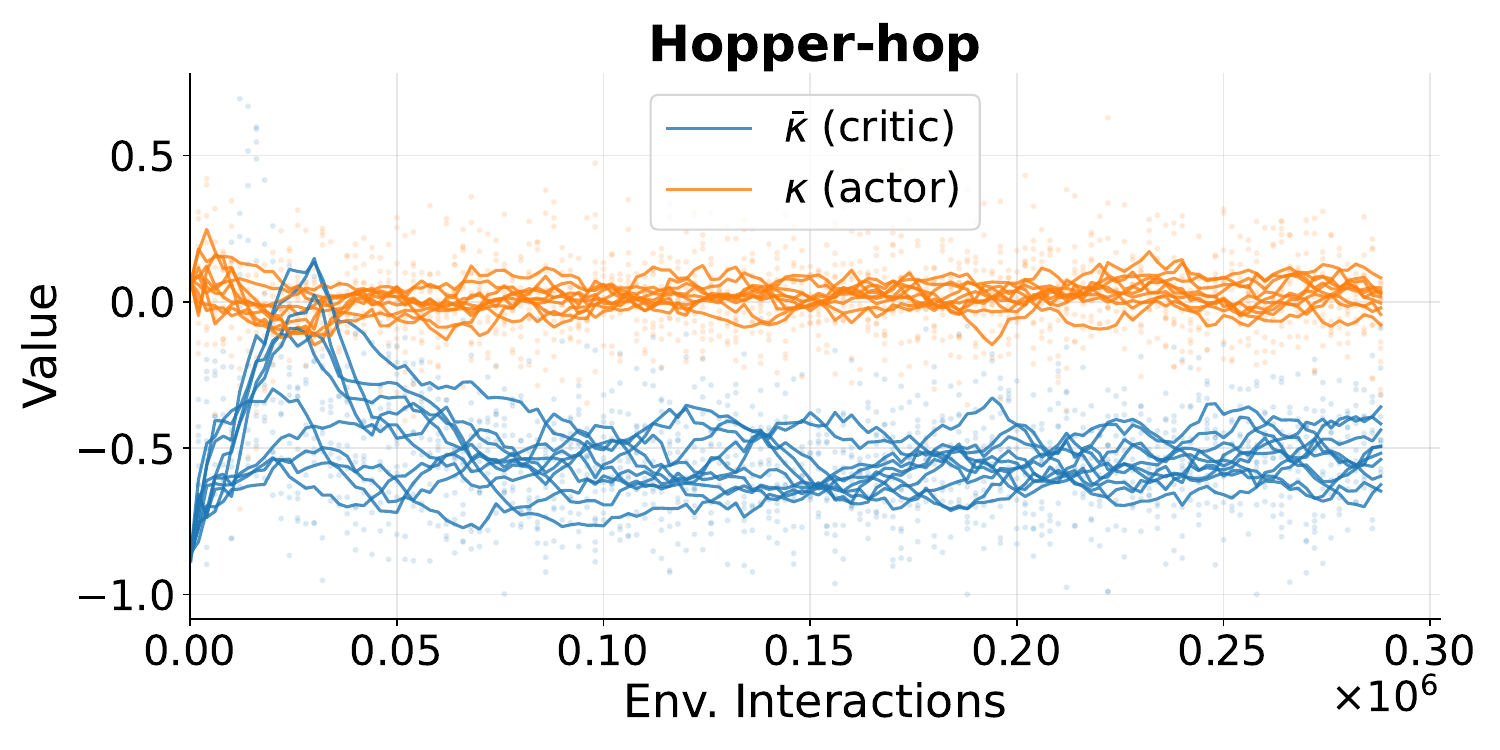}
\includegraphics[width=0.45\linewidth]{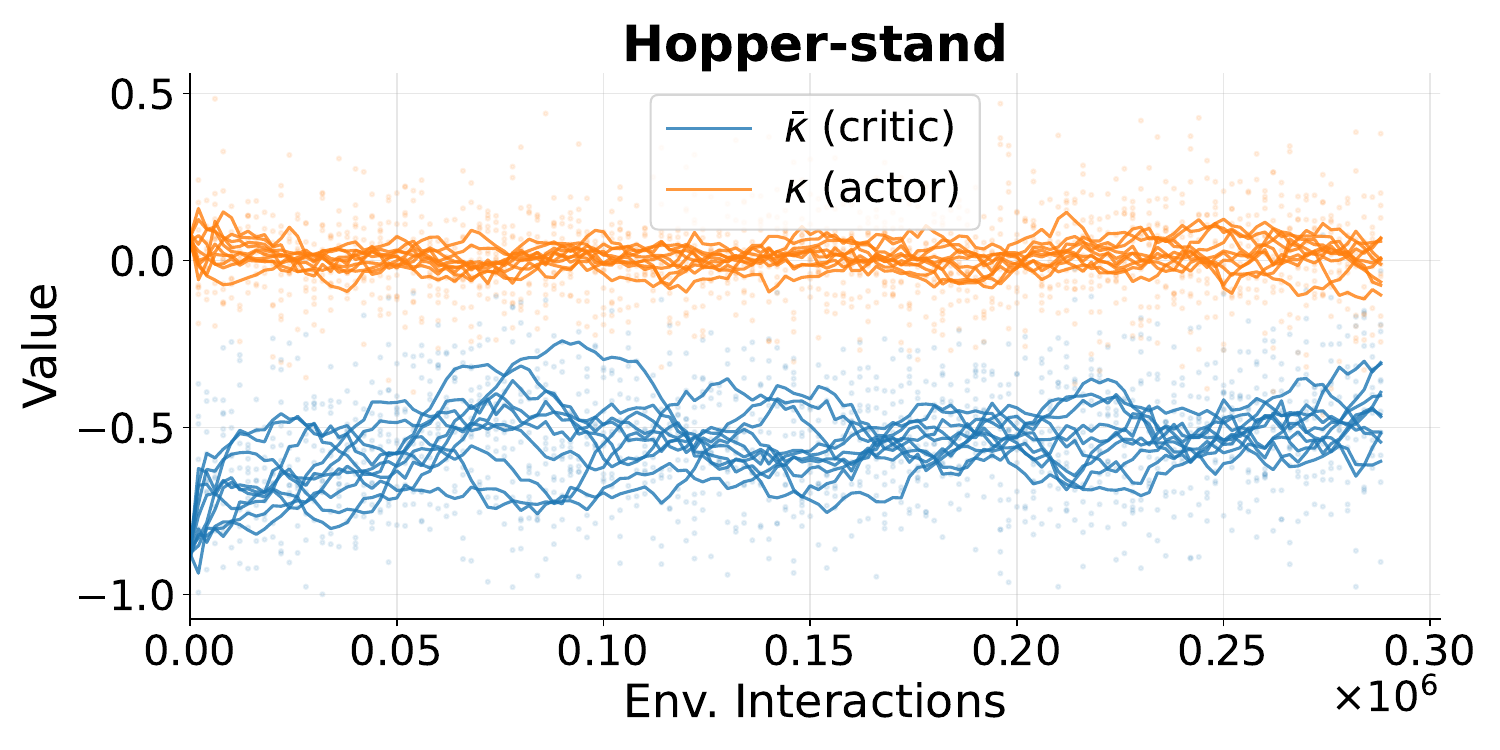}
\includegraphics[width=0.45\linewidth]{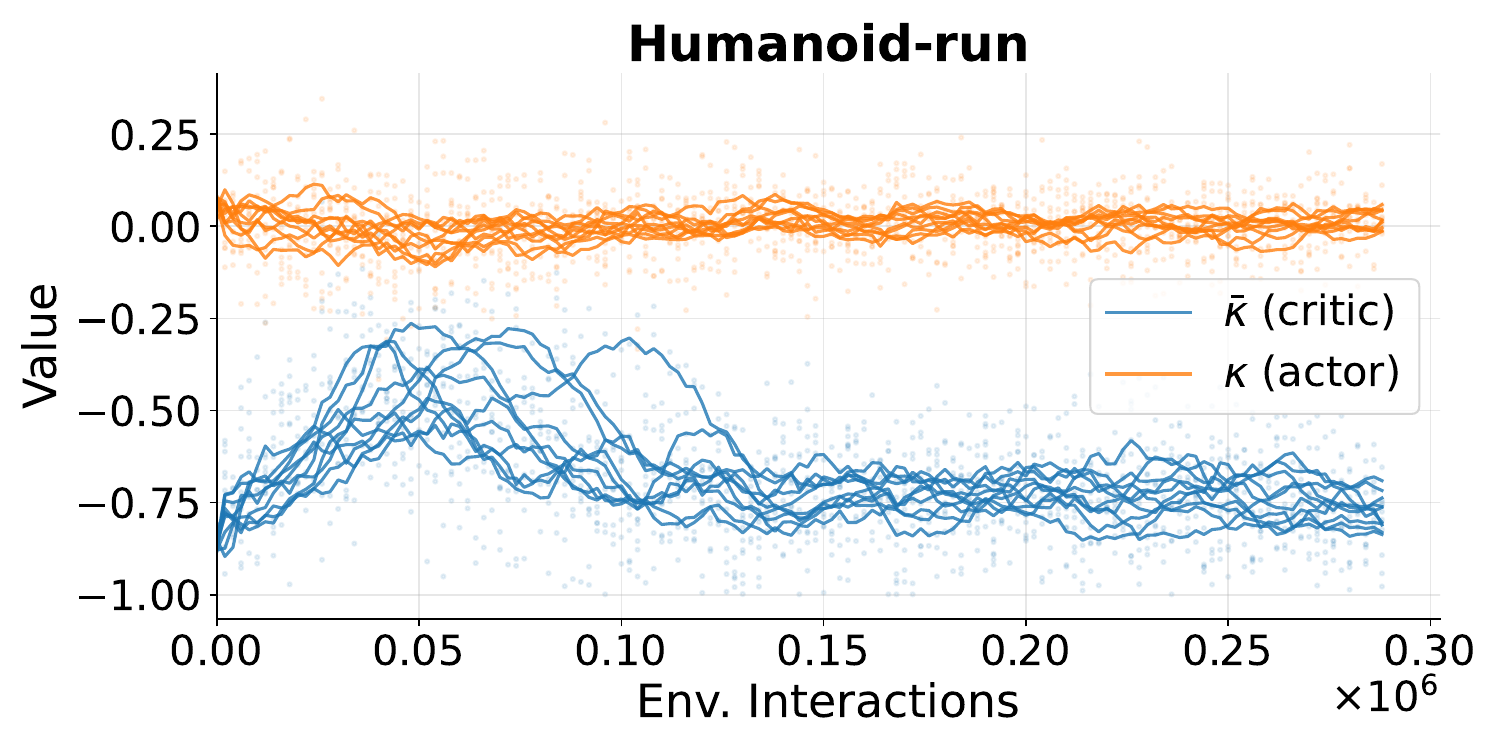}
\includegraphics[width=0.45\linewidth]{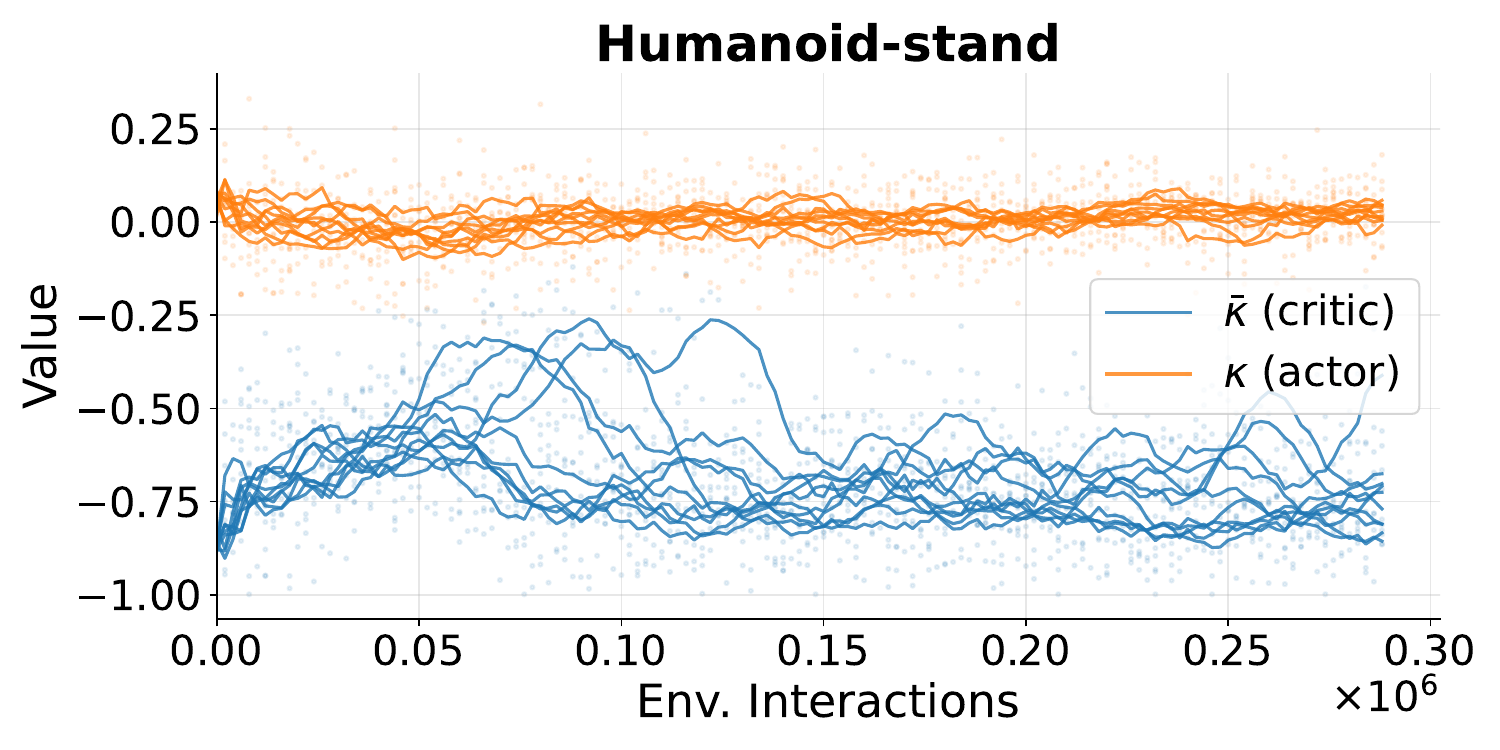}
\includegraphics[width=0.45\linewidth]{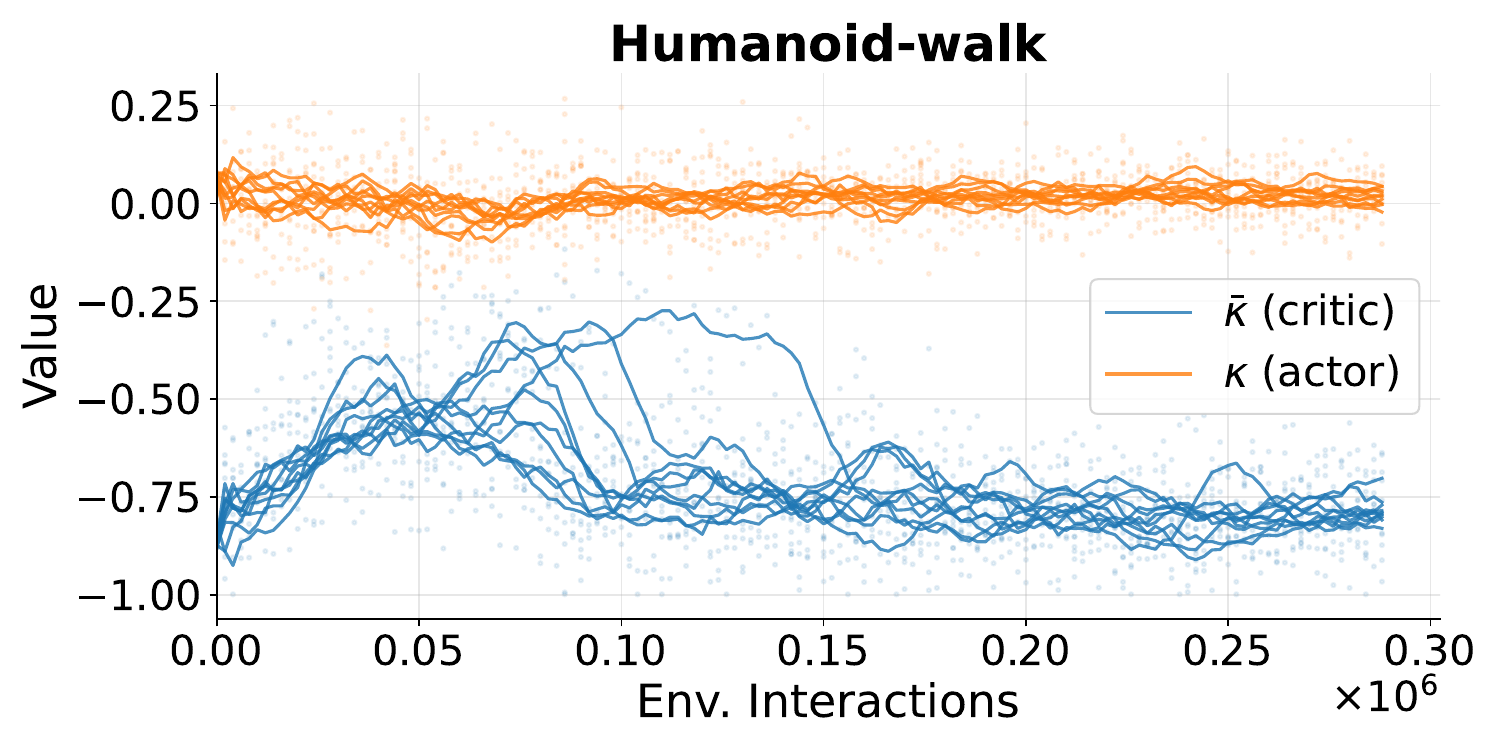}
\includegraphics[width=0.45\linewidth]{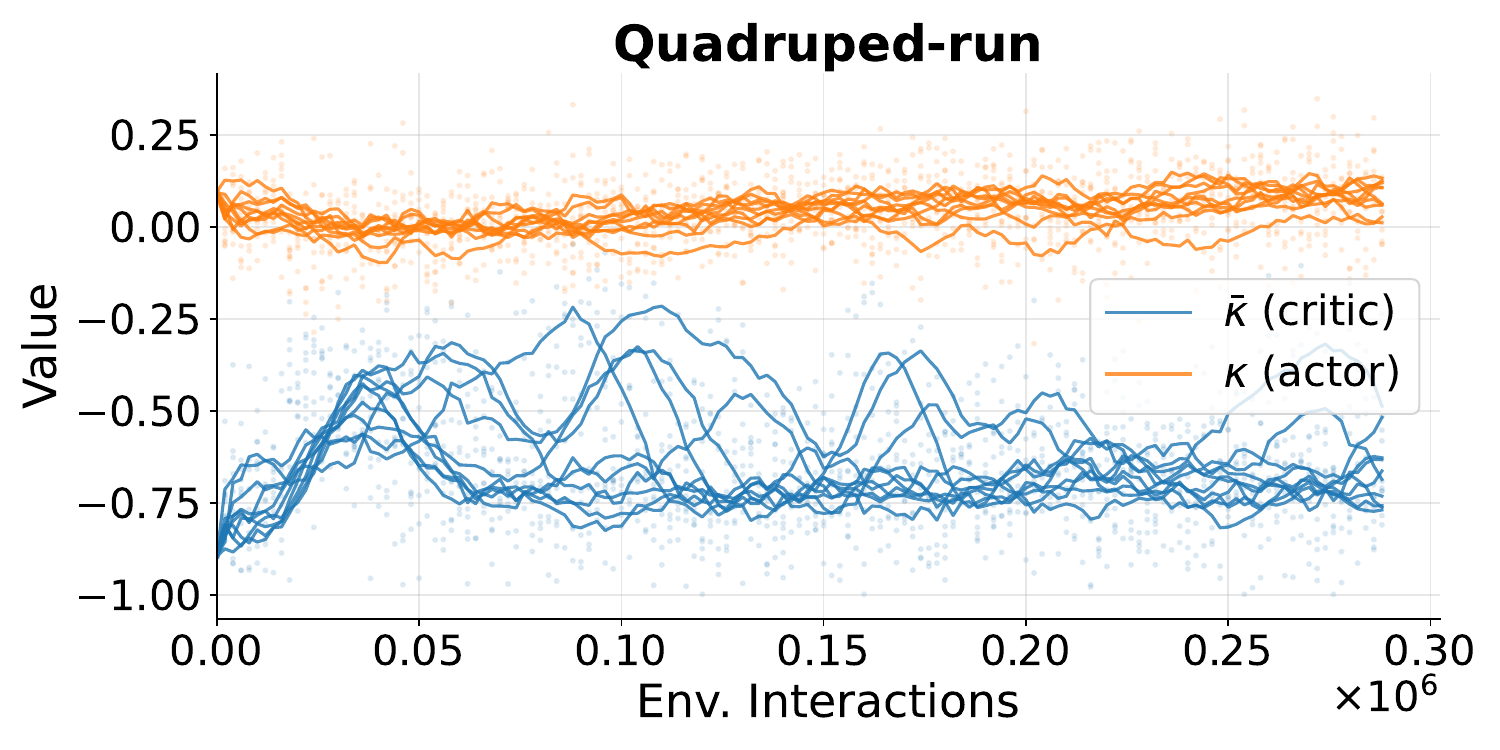}
\includegraphics[width=0.45\linewidth]{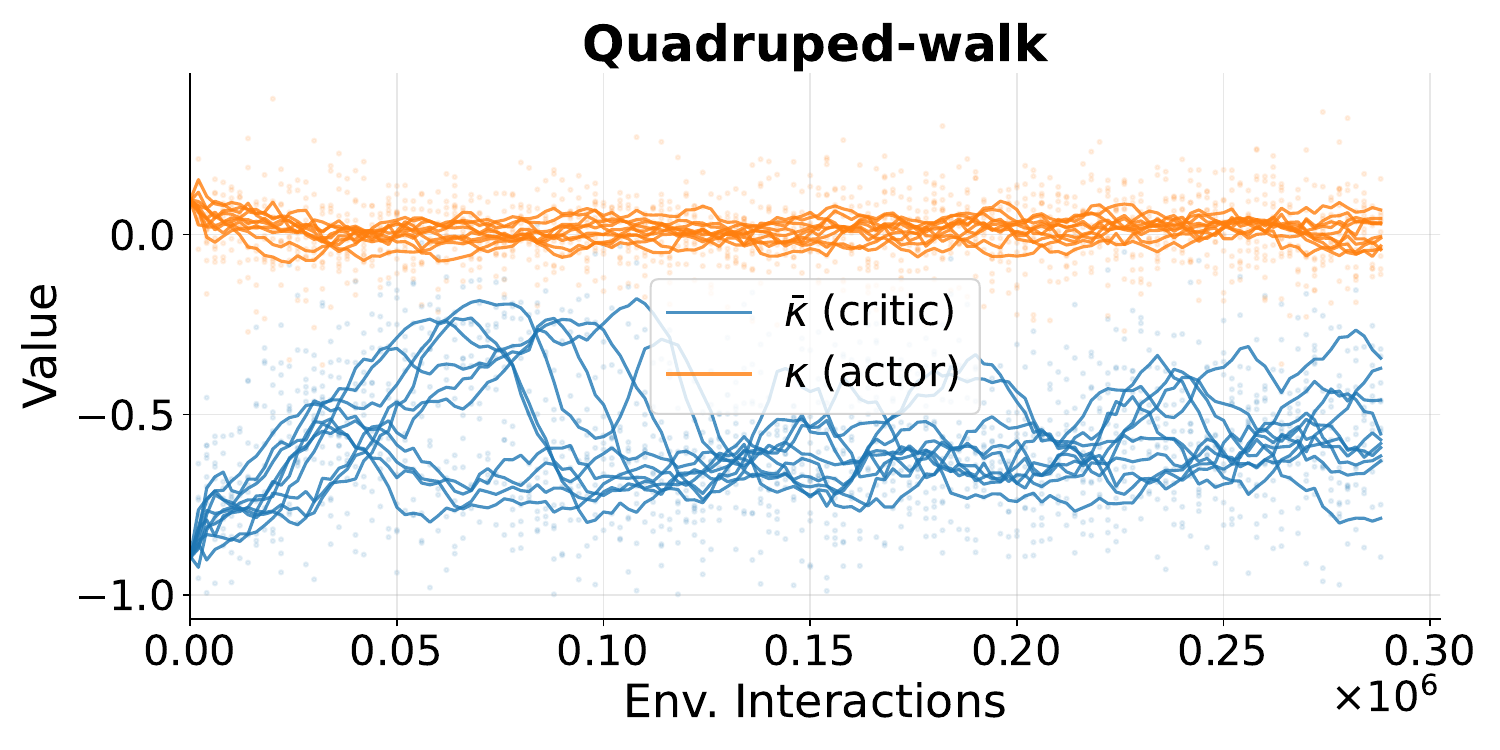}
\includegraphics[width=0.45\linewidth]{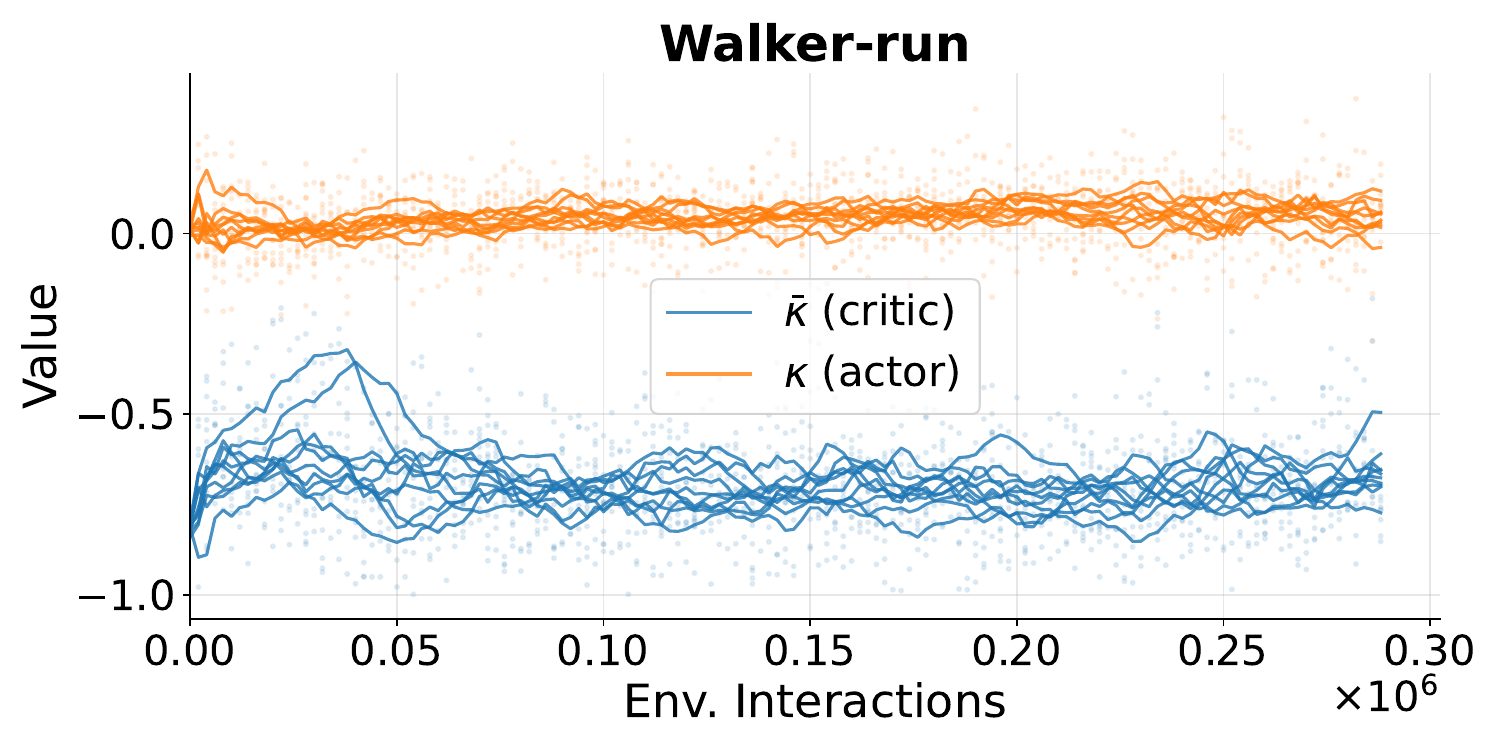}
\caption{Trajectories of the ensemble aggregation parameters $\overline{\kappa}$ (critic) and $\kappa$ (actor) across seeds in DMC environments under $(N=10, G=20)$. Plotted are the raw values and moving averages per seed.}
\label{fig:kappas_trajectory:sample_efficient:dmc}
\end{figure}

\paragraph{Aggregation parameter trajectories.} 
\Cref{fig:kappas_trajectory:interactive:mujoco,fig:kappas_trajectory:sample_efficient:mujoco,fig:kappas_trajectory:interactive:dmc,fig:kappas_trajectory:sample_efficient:dmc} shows the evolution of the learned ensemble aggregation parameters ($\overline{\kappa}$ for the critic and $\kappa$ for the actor) throughout training across the two learning regimes for each seed. These figures illustrate how AEA adjusts aggregation behavior in response to training dynamics. $\overline{\kappa}$ typically remains negative, anchoring conservative critic estimates, while $\kappa$ tends to increase, supporting more explorative actor behavior.

\begin{figure}
\centering
\includegraphics[width=0.45\linewidth]{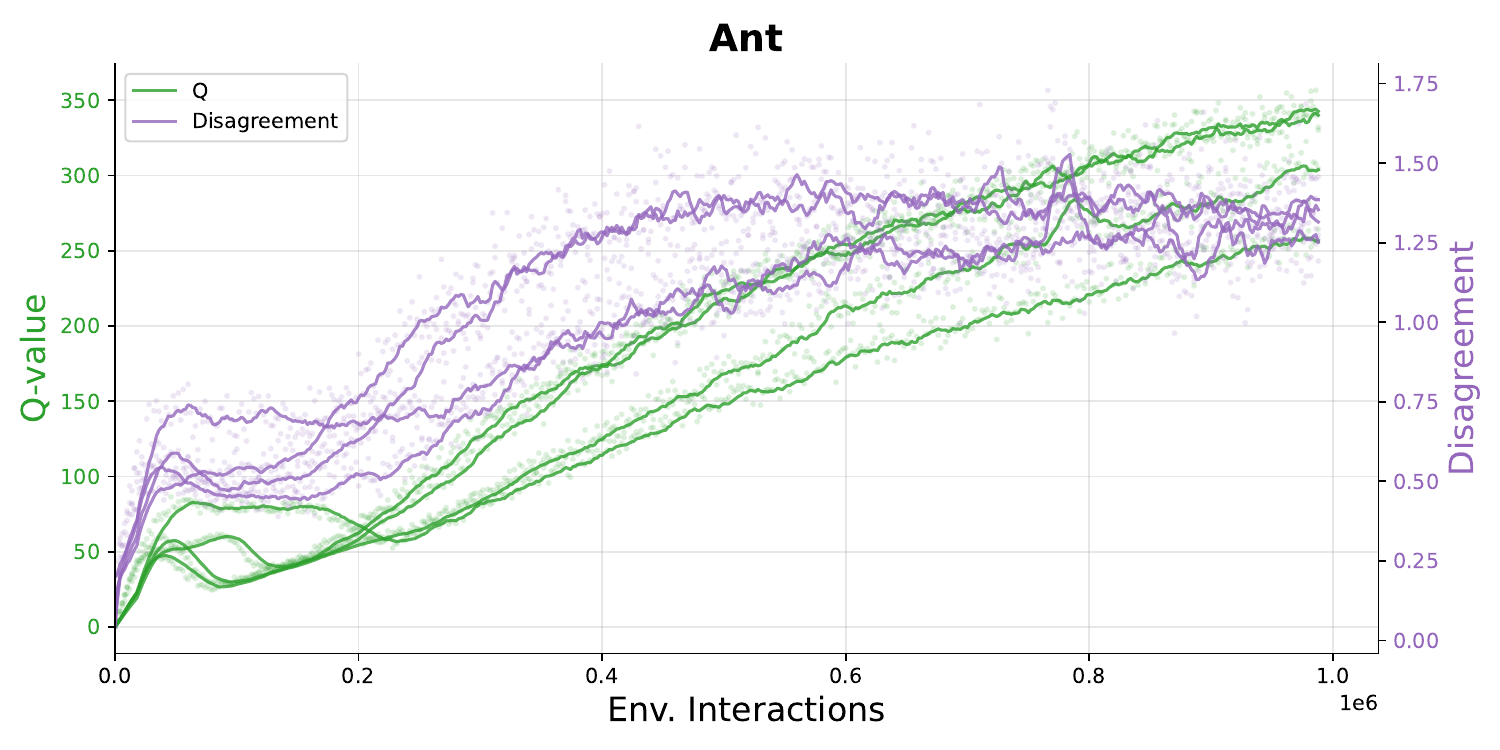}
\includegraphics[width=0.45\linewidth]{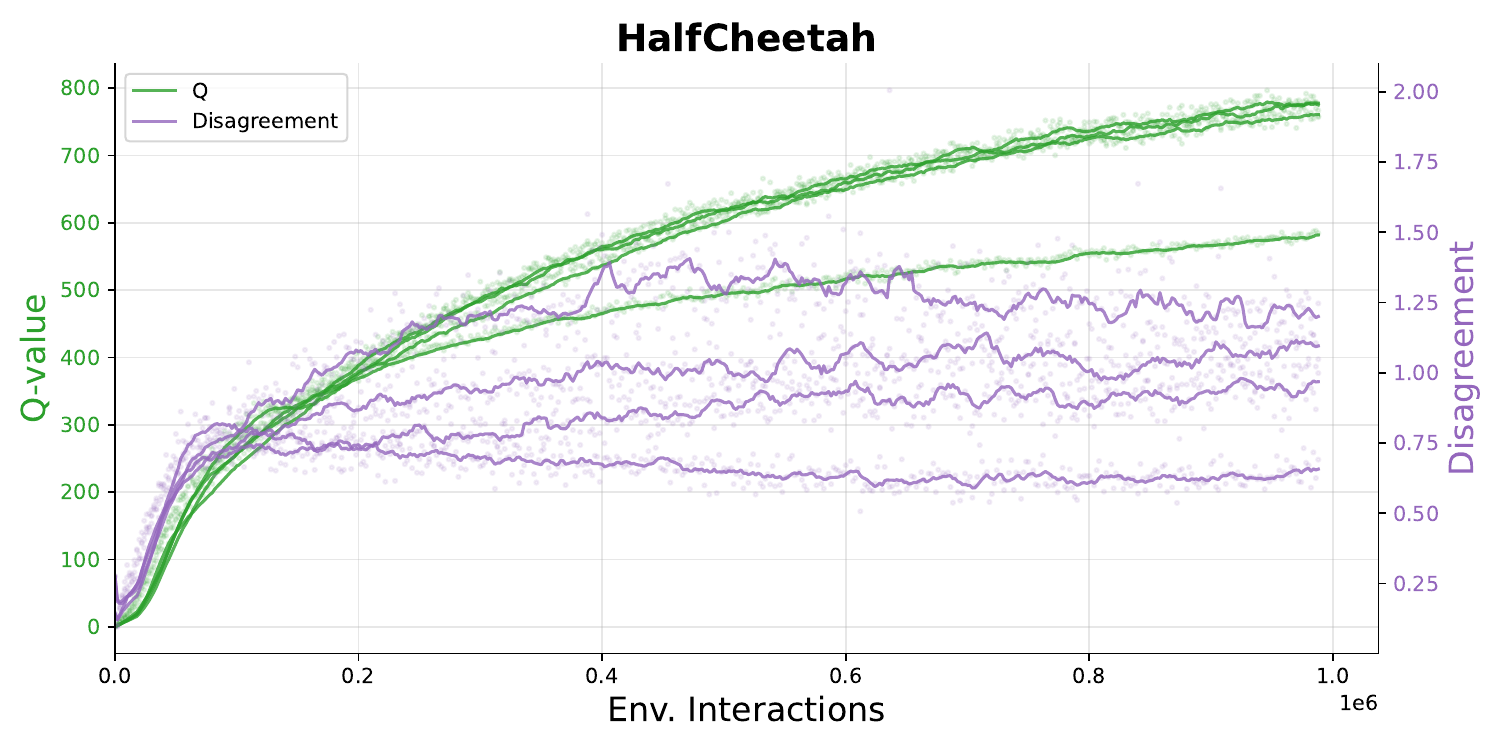}
\includegraphics[width=0.45\linewidth]{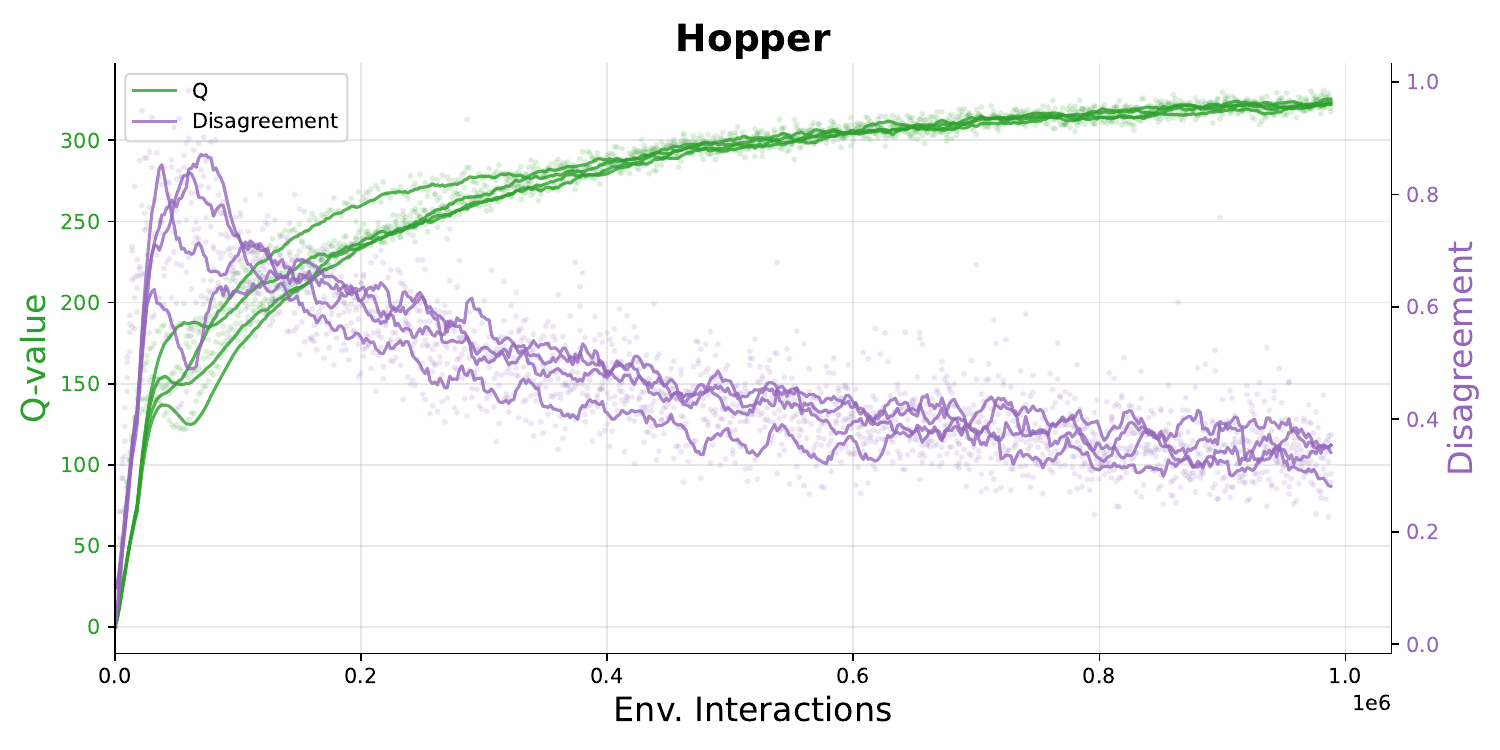}
\includegraphics[width=0.45\linewidth]{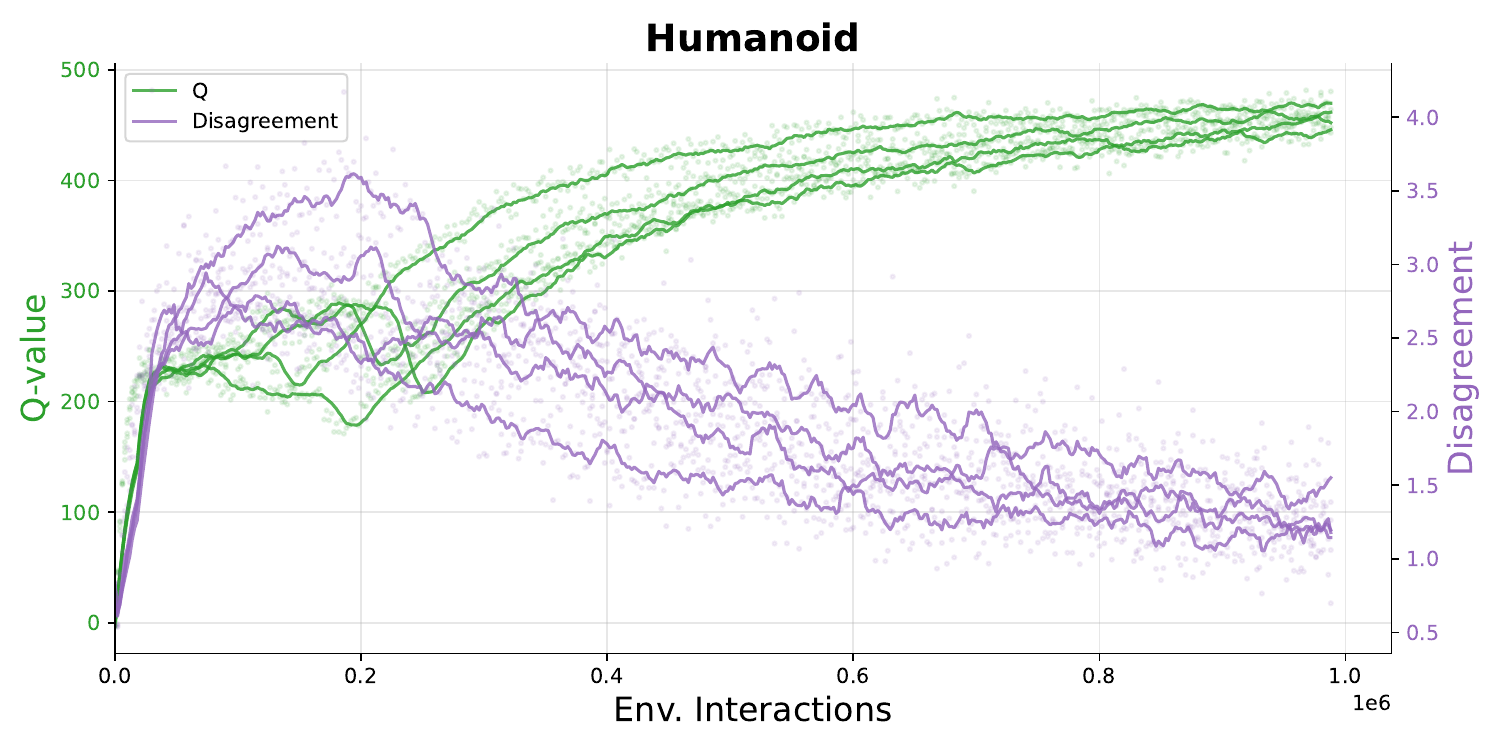}
\includegraphics[width=0.45\linewidth]{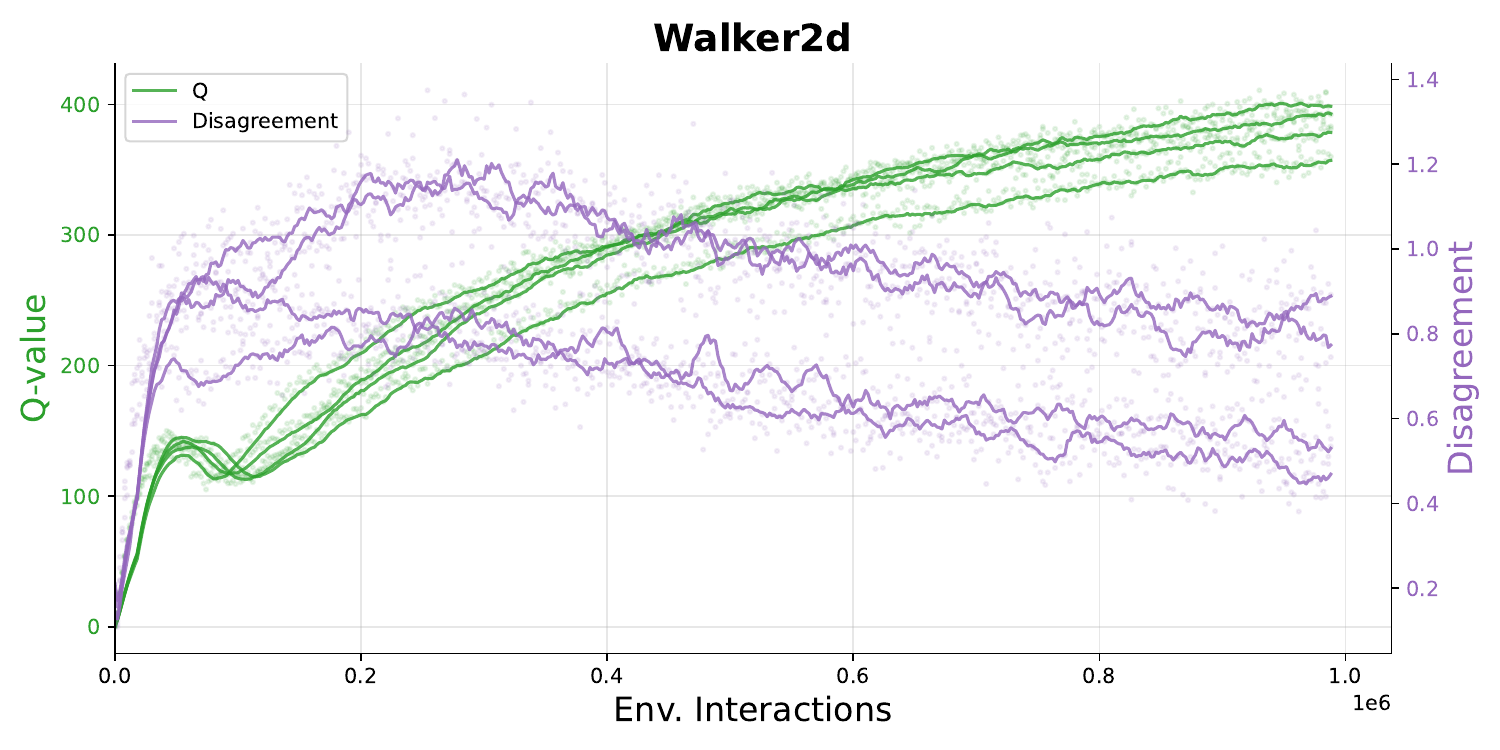}
\caption{Trajectories of $Q$-values and ensemble disagreement in MuJoCo environments under $(N=2, G=1)$. Plotted are the raw values and moving averages per seed.}
\label{fig:dis_trajectory:interactive:mujoco}
\end{figure}

\begin{figure}
\centering
\includegraphics[width=0.45\linewidth]{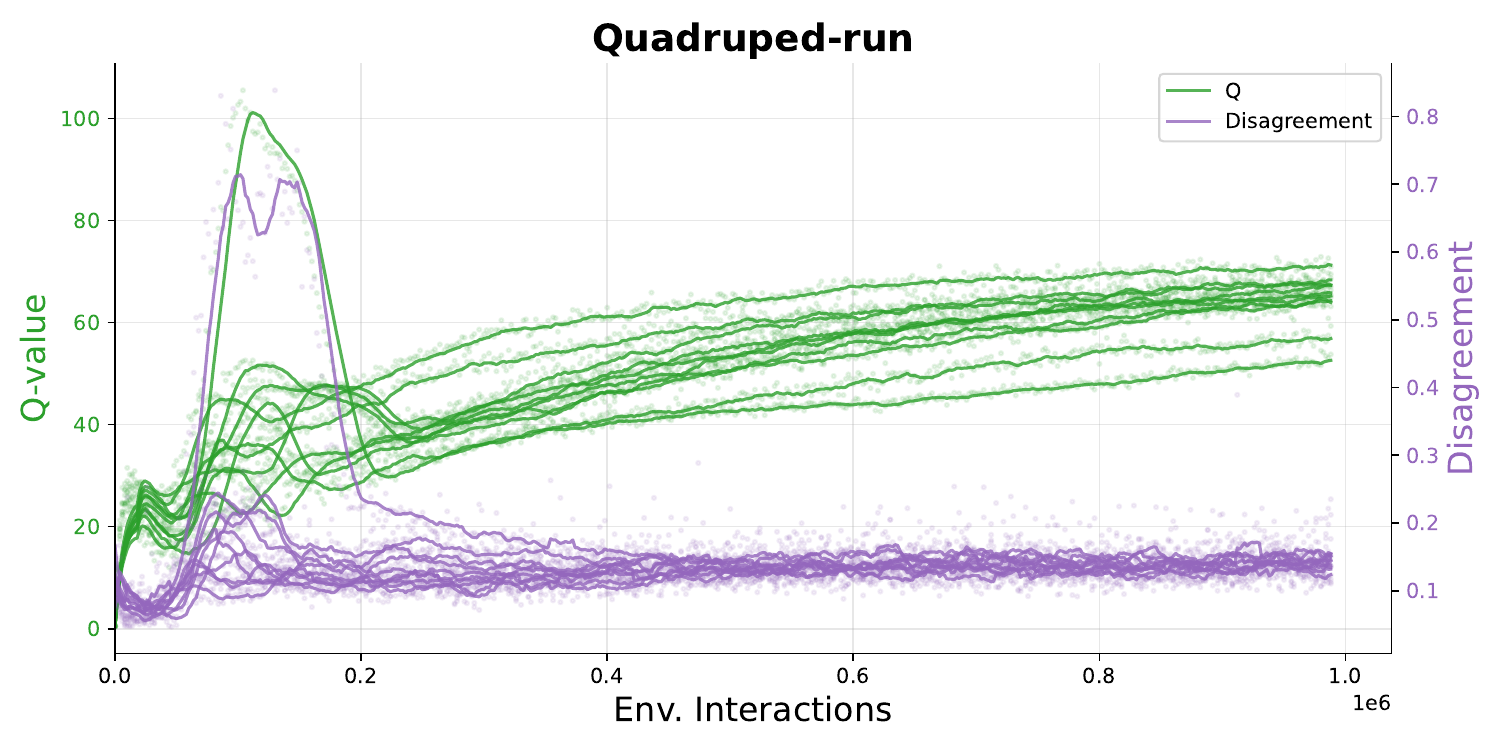}
\includegraphics[width=0.45\linewidth]{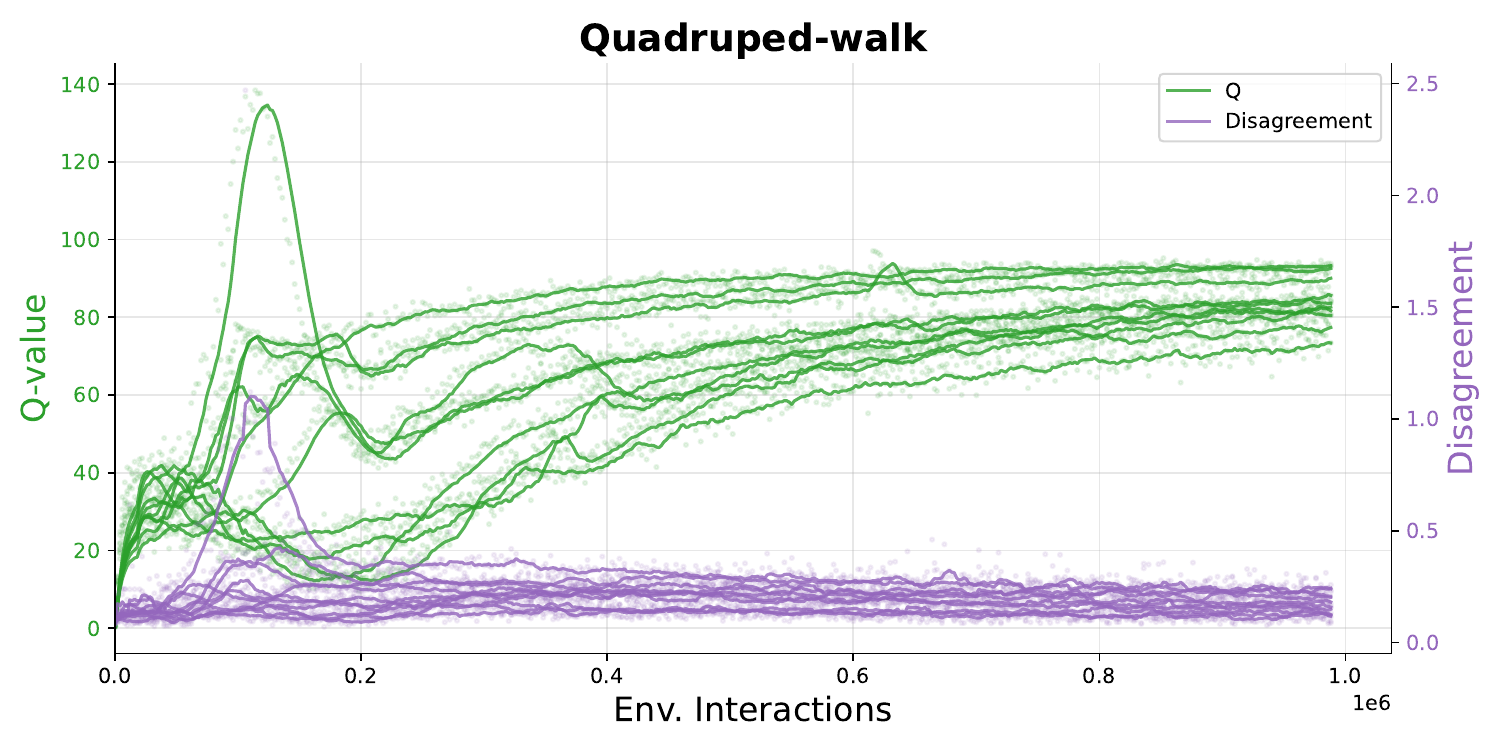}
\includegraphics[width=0.45\linewidth]{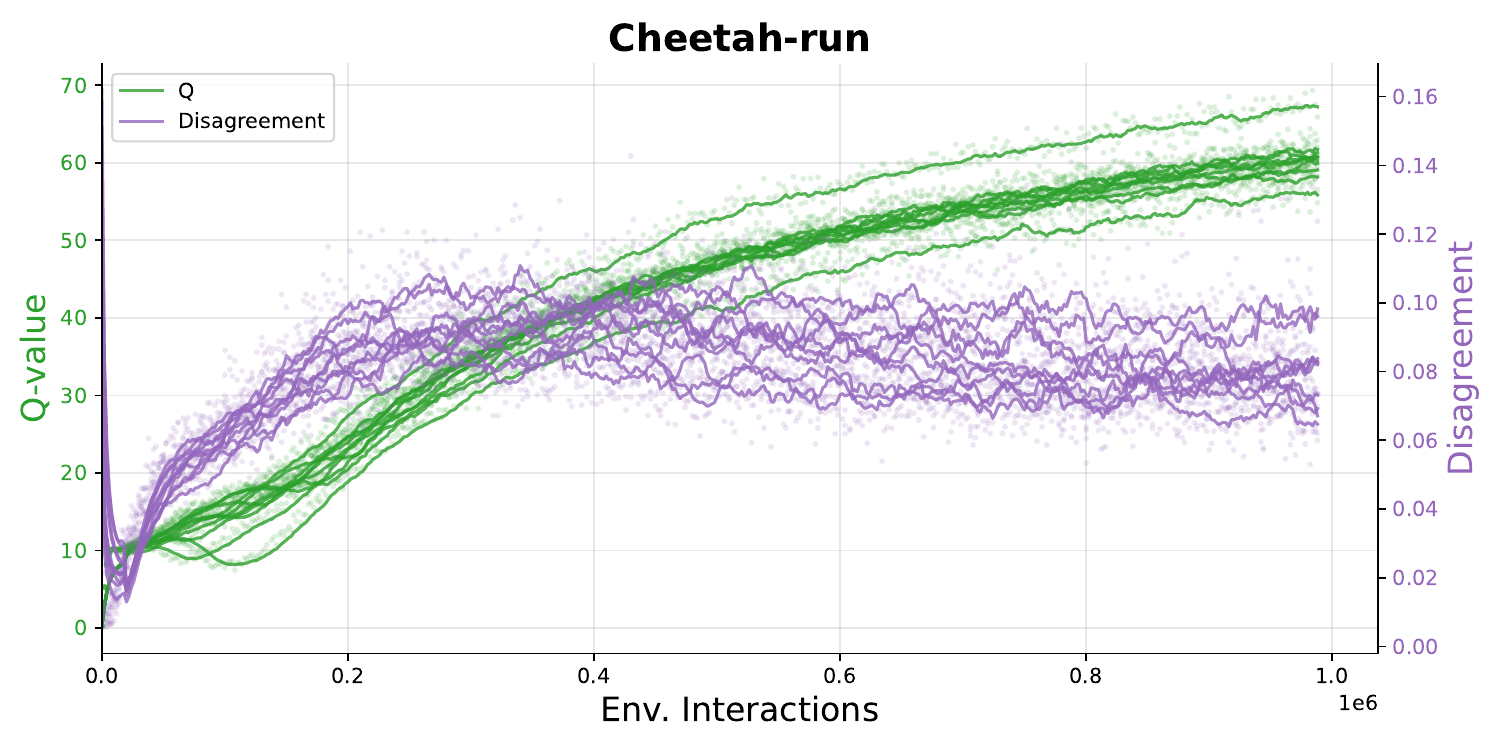}
\includegraphics[width=0.45\linewidth]{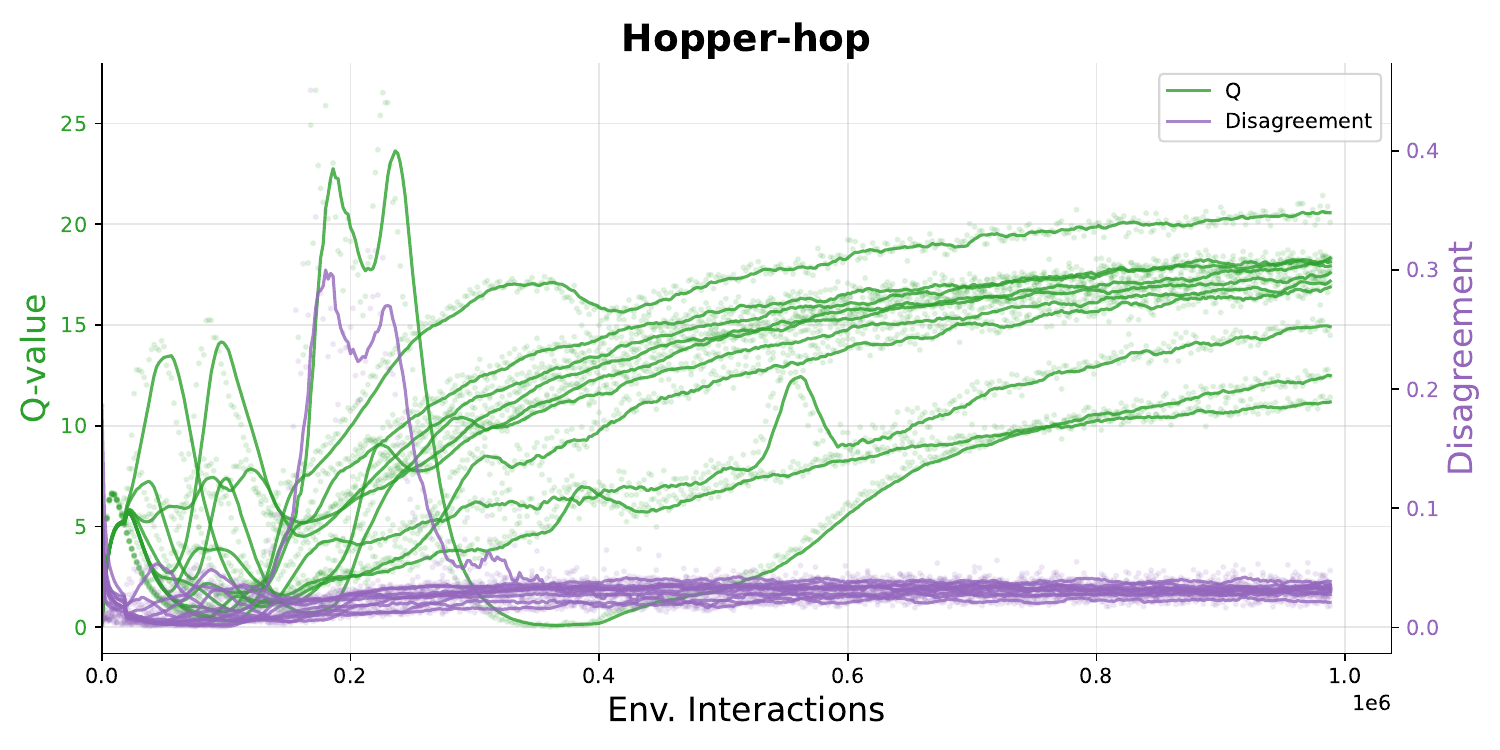}
\includegraphics[width=0.45\linewidth]{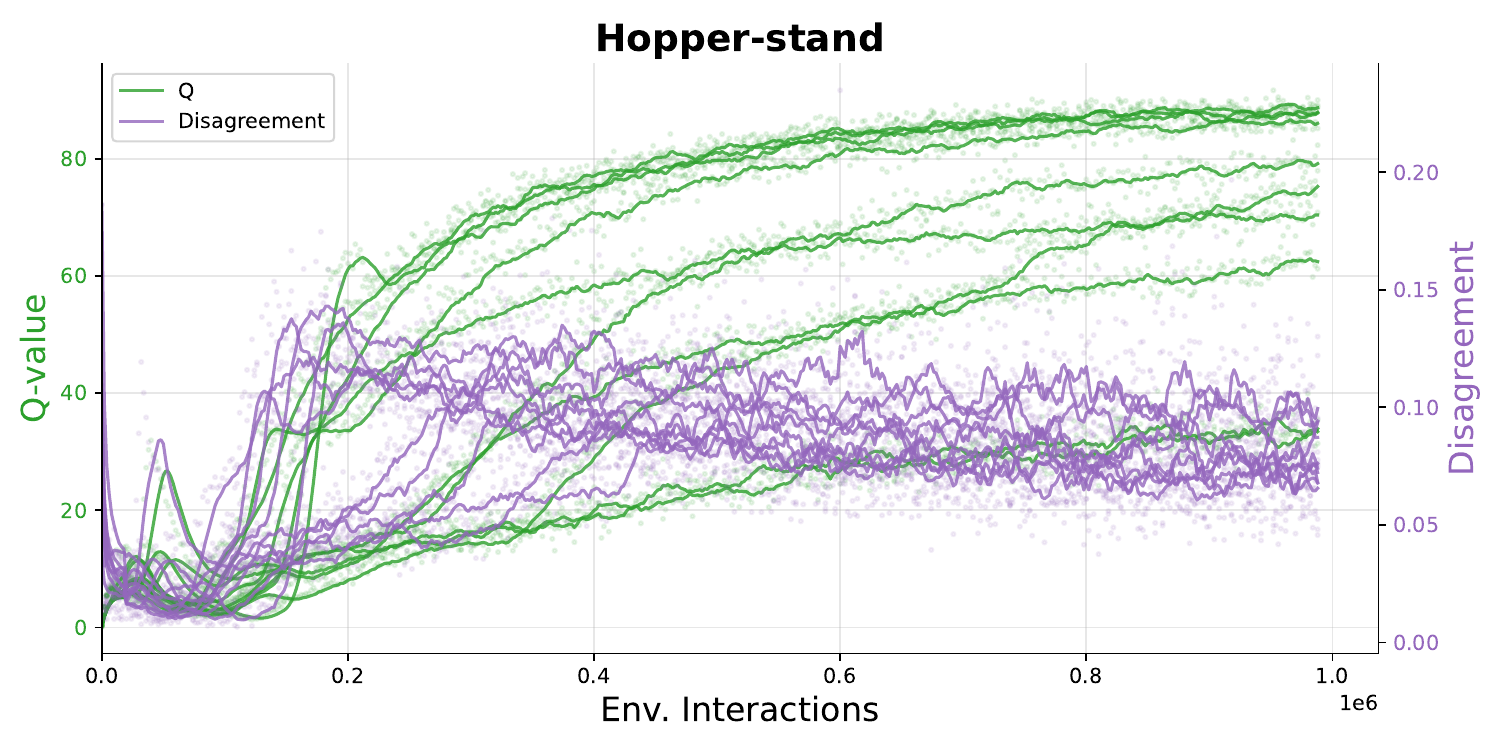}
\includegraphics[width=0.45\linewidth]{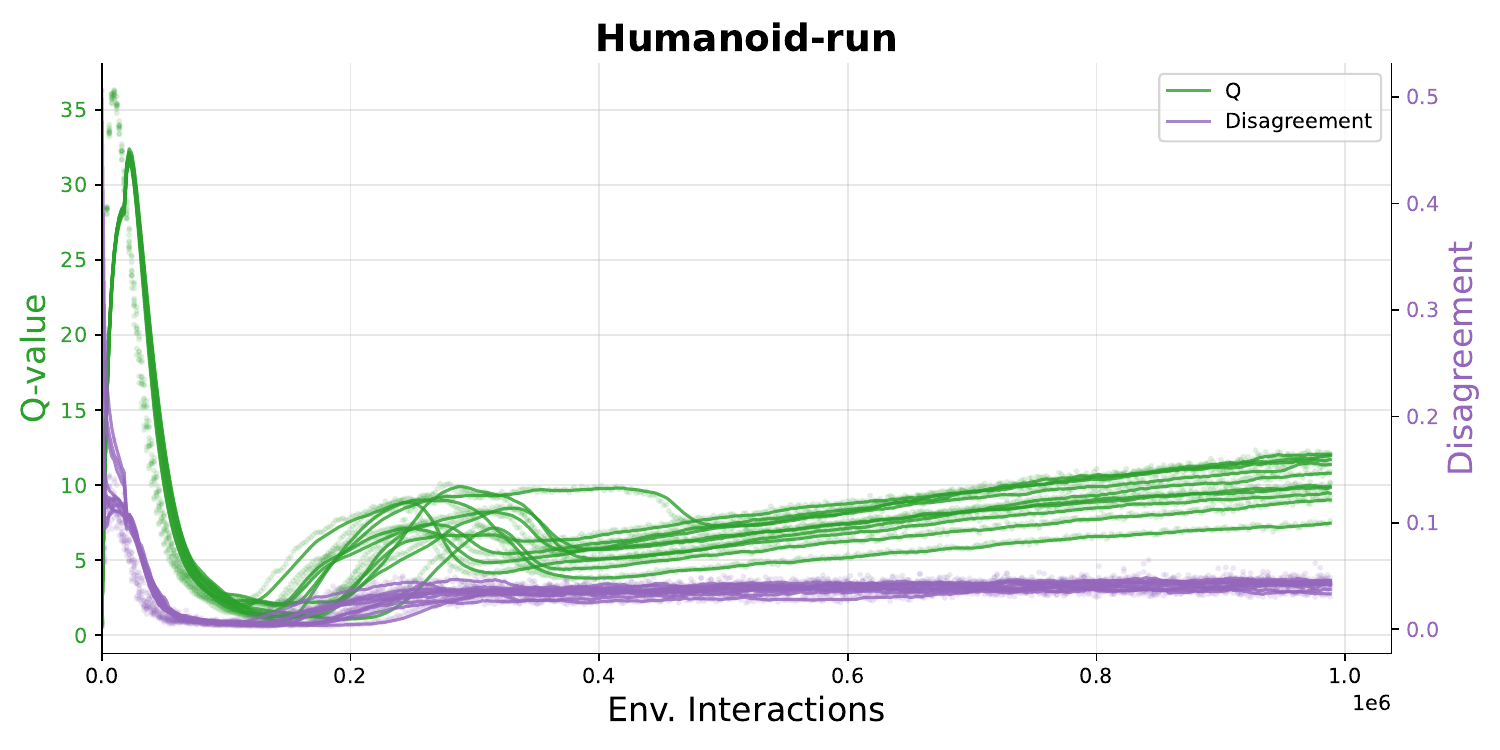}
\includegraphics[width=0.45\linewidth]{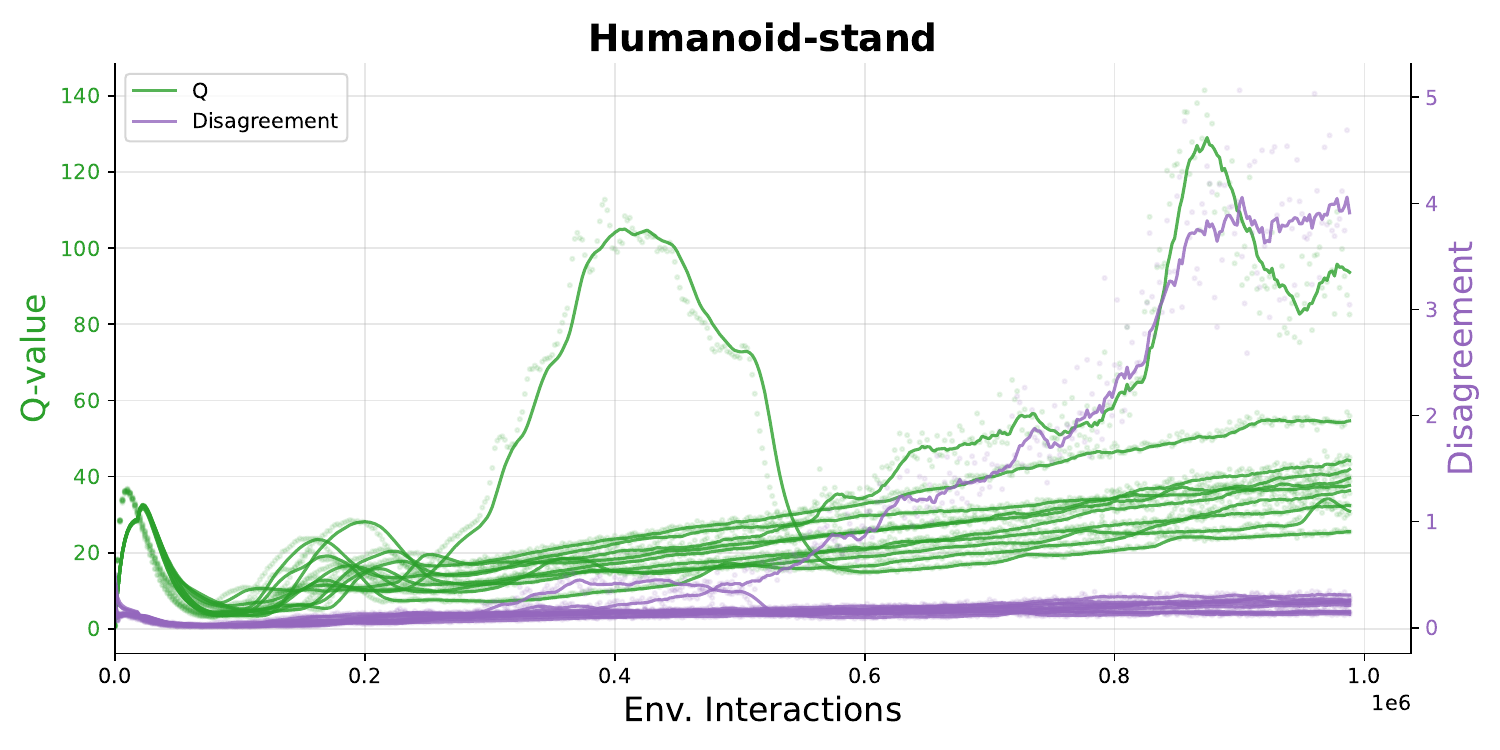}
\includegraphics[width=0.45\linewidth]{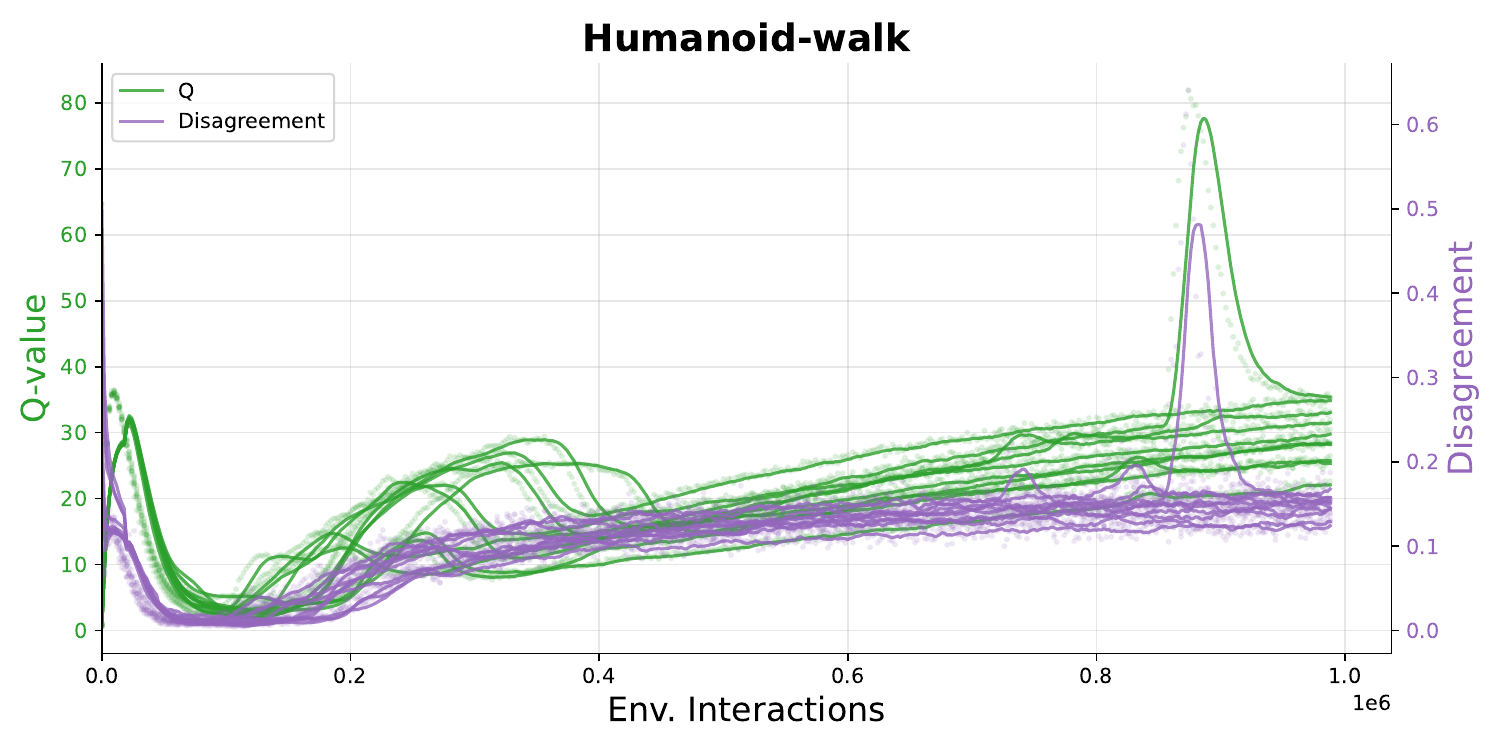}
\includegraphics[width=0.45\linewidth]{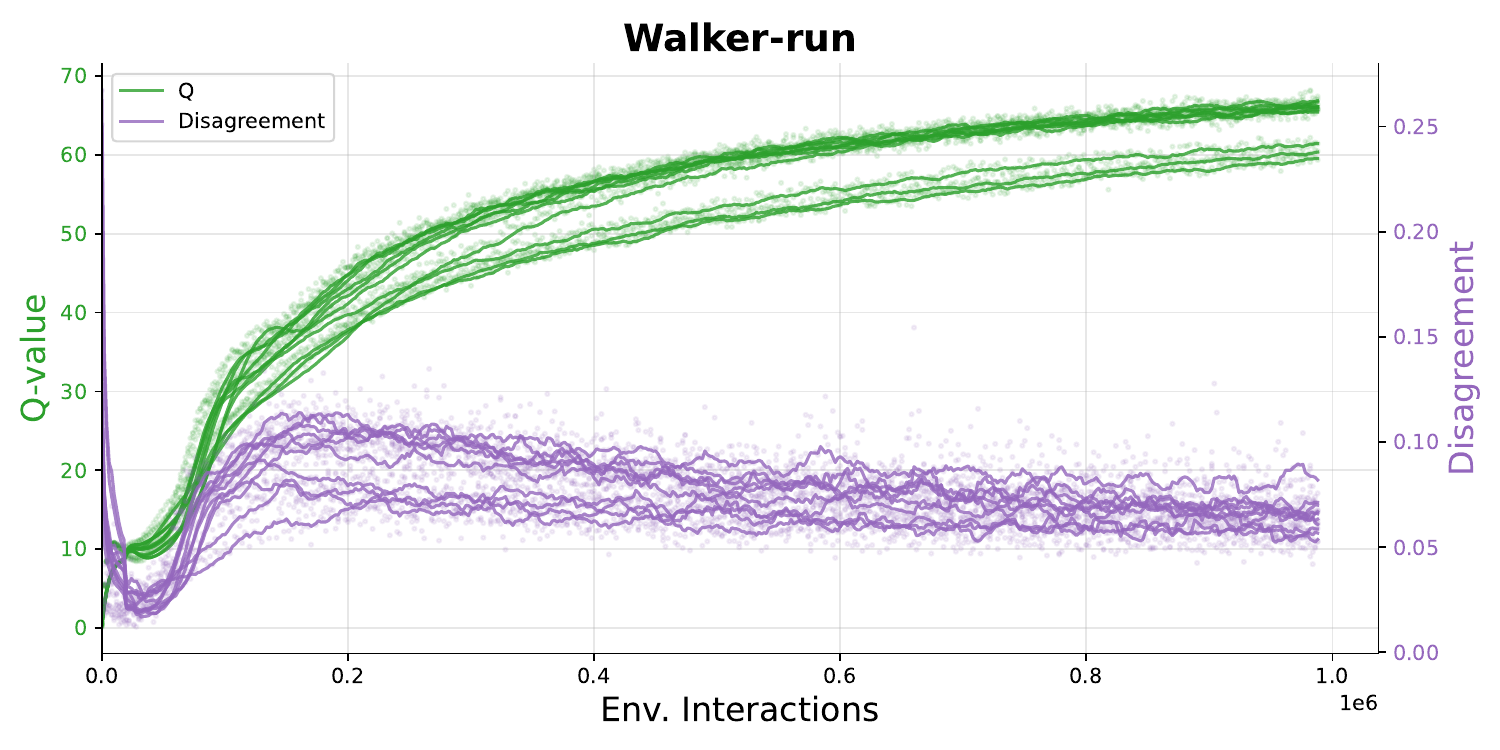}
\caption{Trajectories of $Q$-values and ensemble disagreement in DMC environments under $(N=2, G=1)$. Plotted are the raw values and moving averages per seed.}
\label{fig:dis_trajectory:interactive:dmc}
\end{figure}

\begin{figure}
\centering
\includegraphics[width=0.45\linewidth]{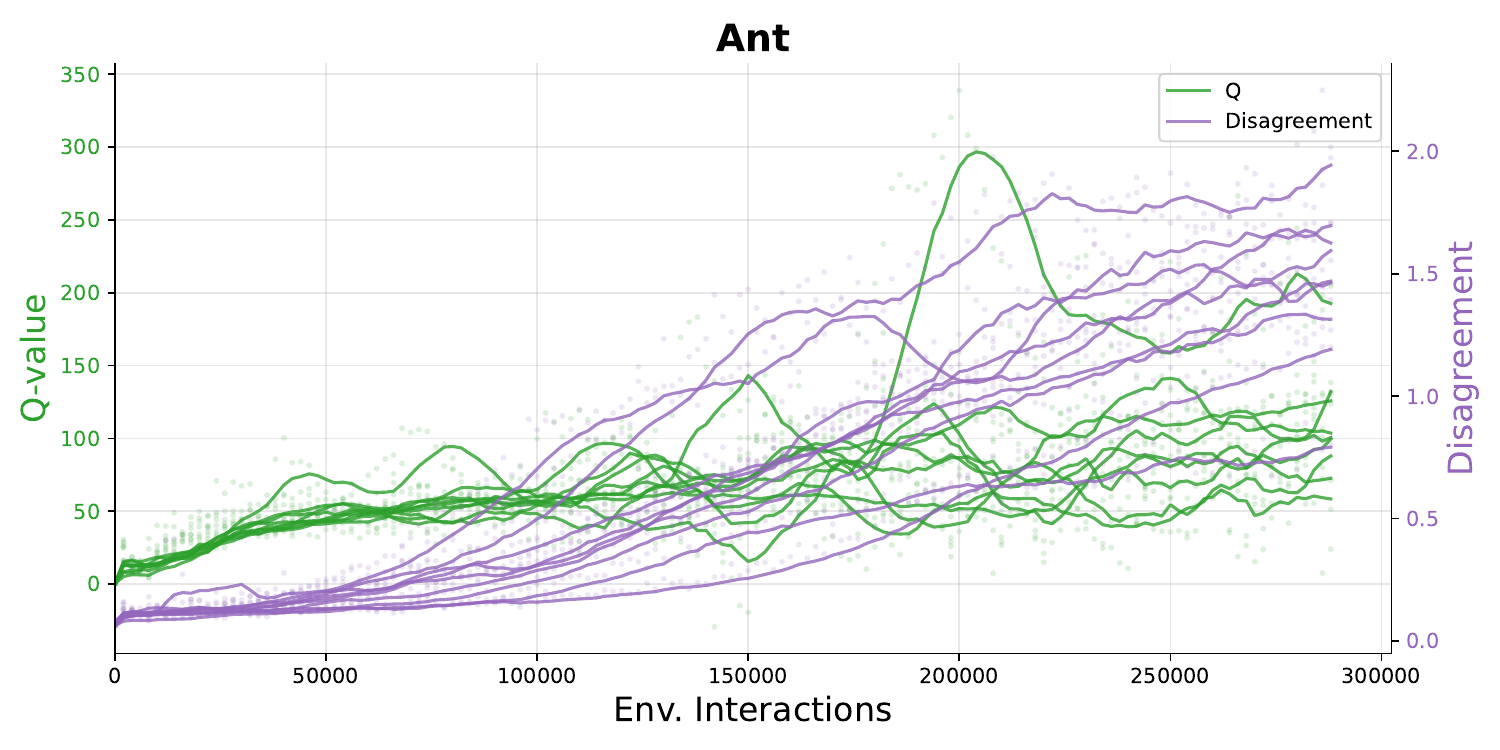}
\includegraphics[width=0.45\linewidth]{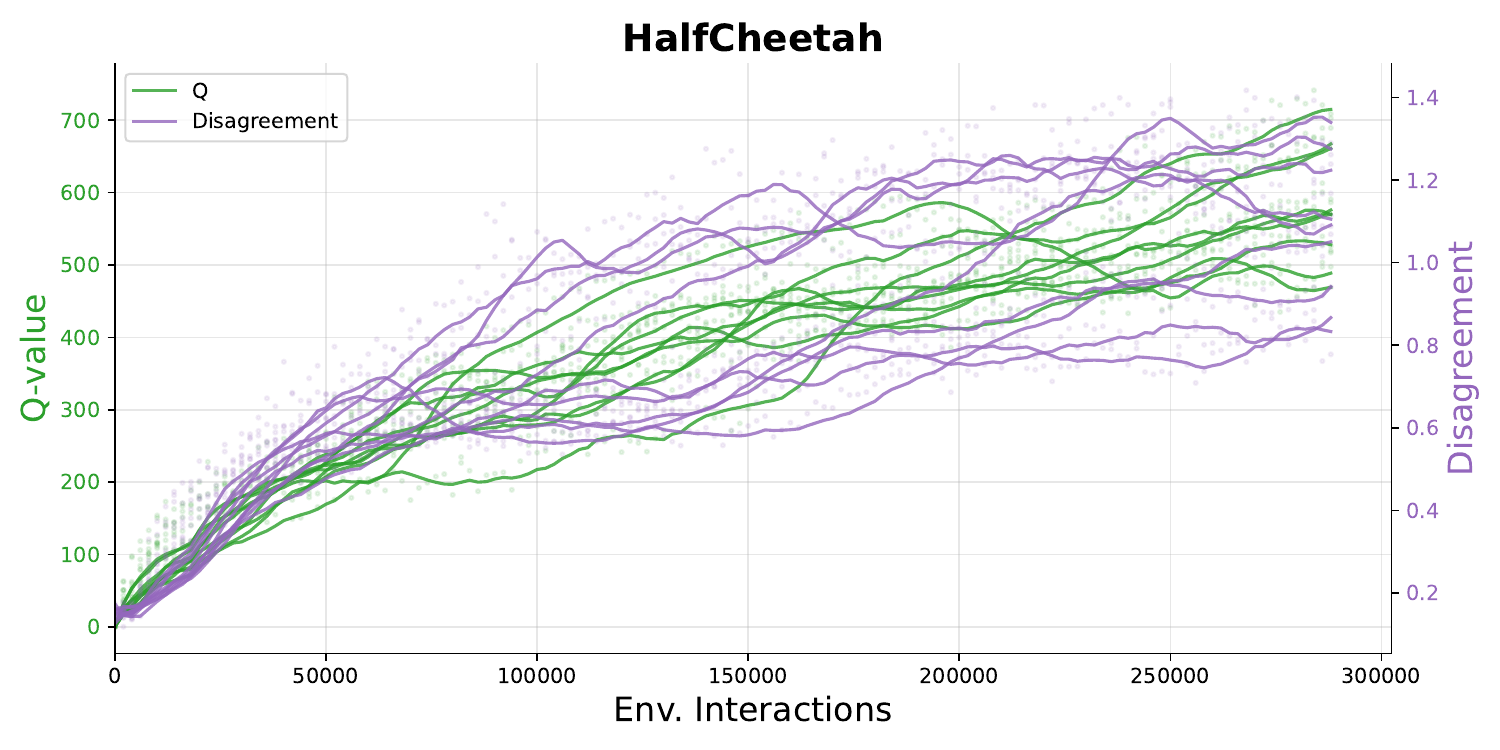}
\includegraphics[width=0.45\linewidth]{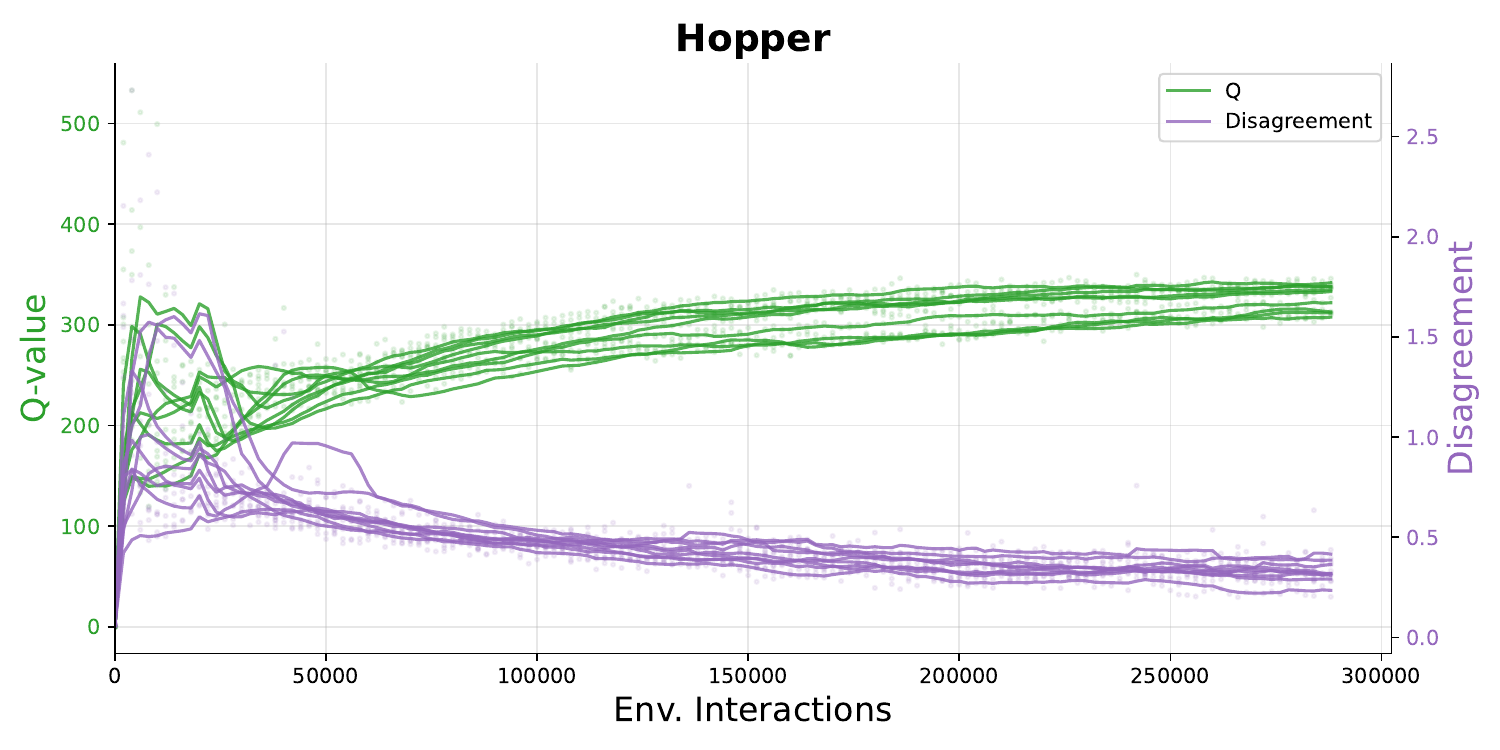}
\includegraphics[width=0.45\linewidth]{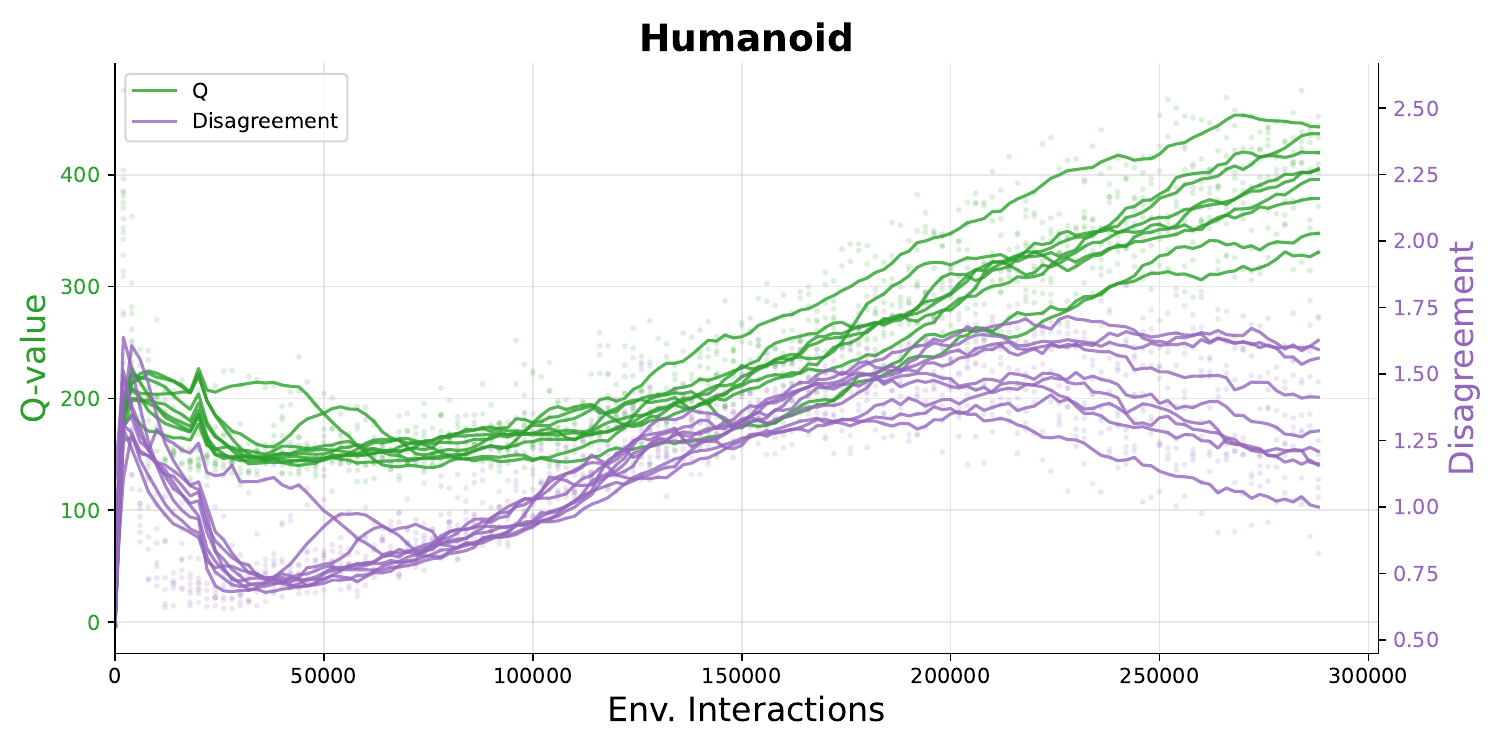}
\includegraphics[width=0.45\linewidth]{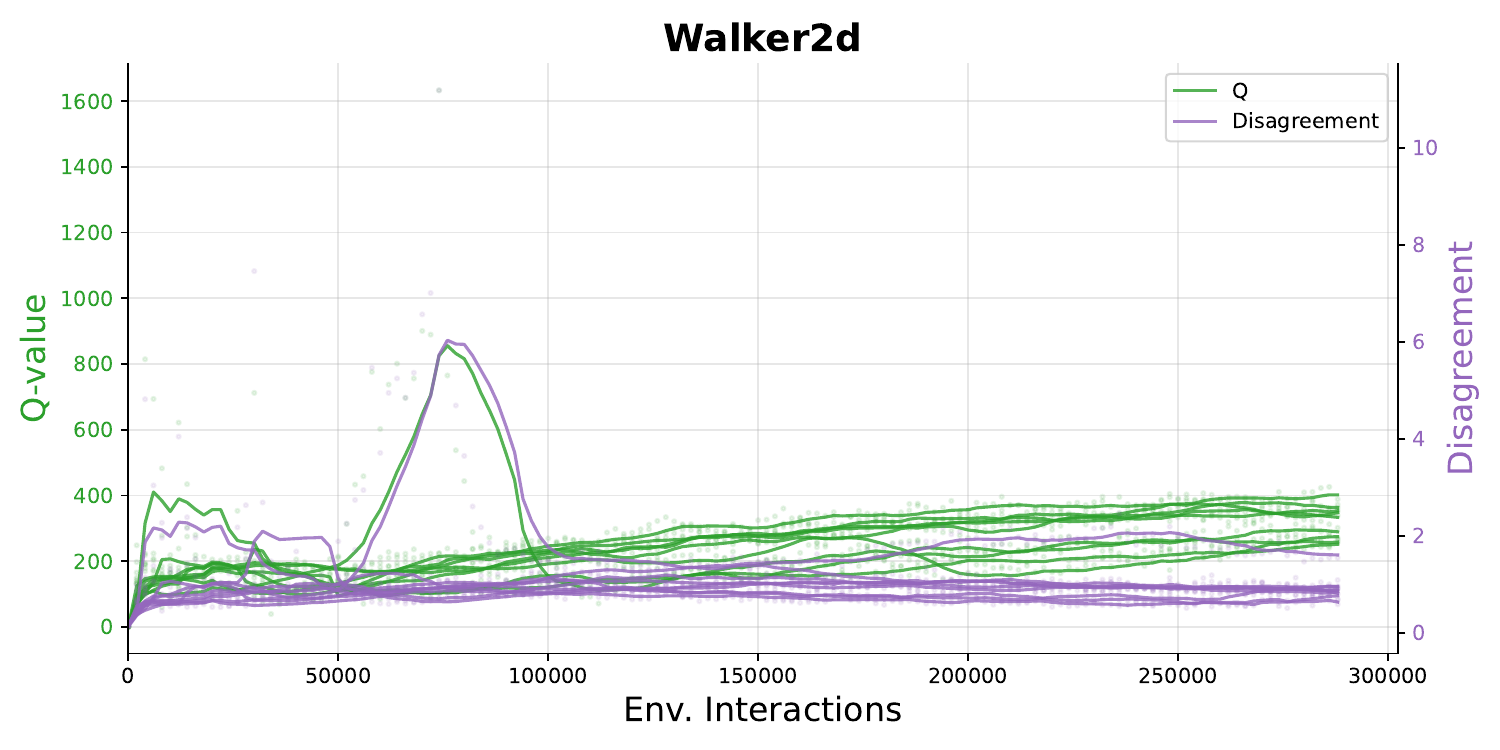}
\caption{Trajectories of $Q$-values and ensemble disagreement for seed 1 in MuJoCo environments under the $(N=10, G=20)$. Plotted are the raw values and moving averages per seed.}
\label{fig:dis_trajectory:sample_efficient:mujoco}
\end{figure}

\begin{figure}
\centering
\includegraphics[width=0.45\linewidth]{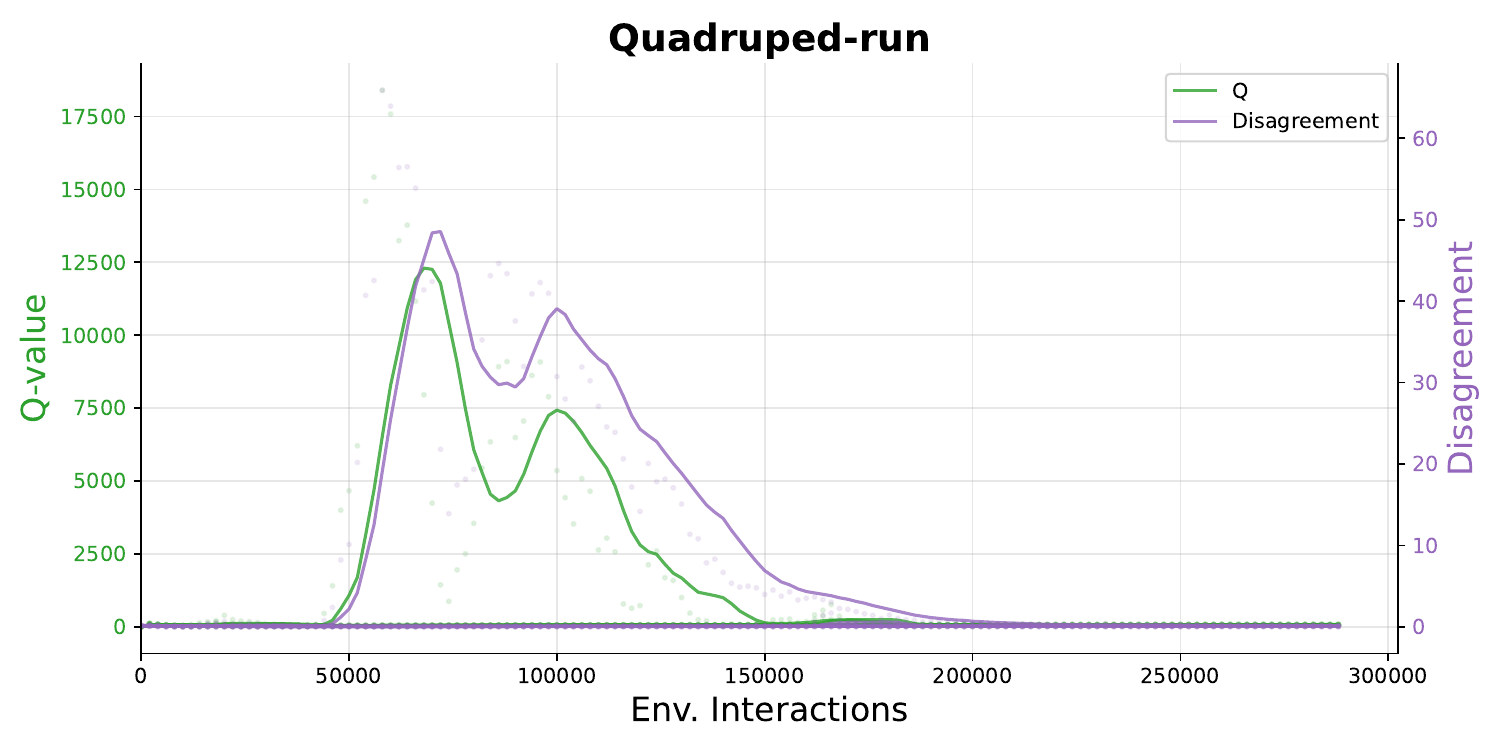}
\includegraphics[width=0.45\linewidth]{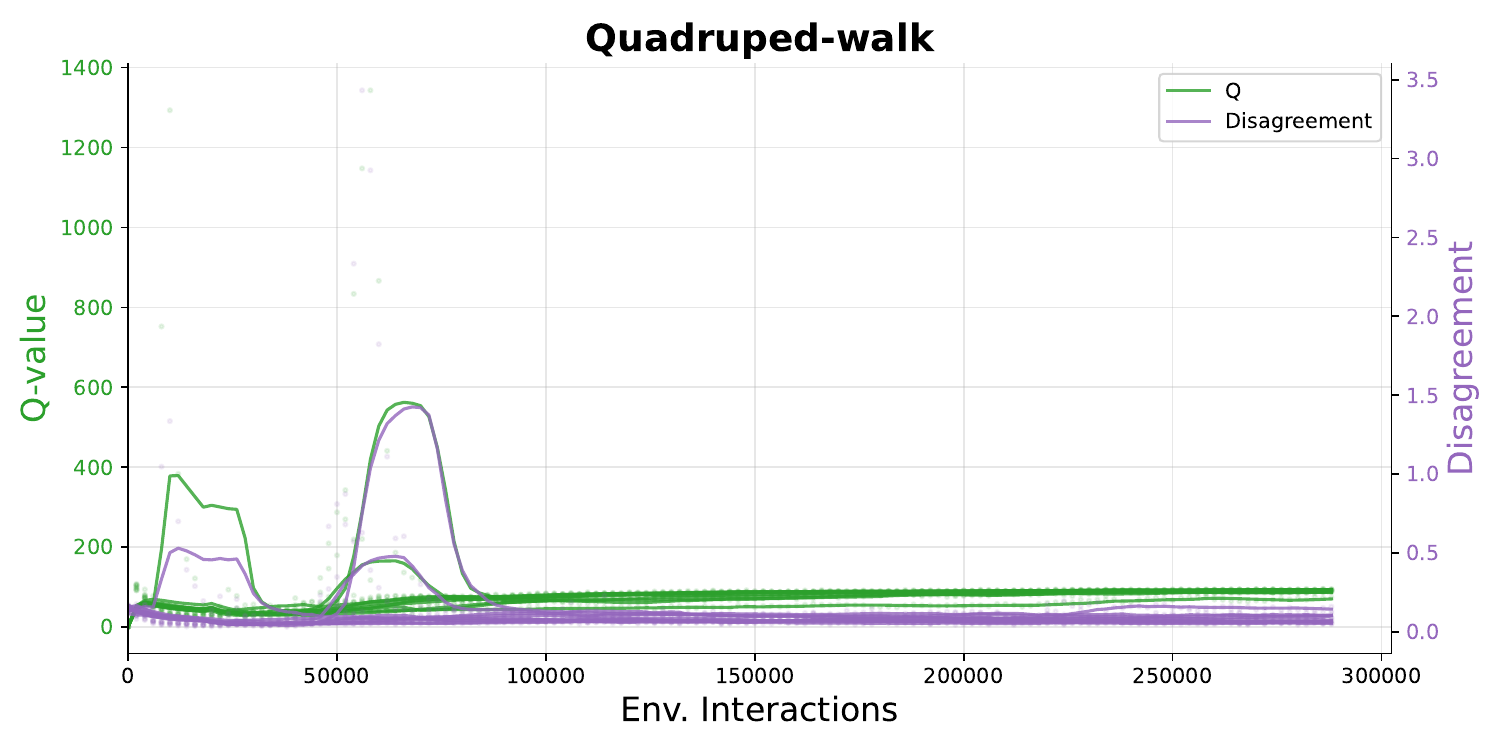}
\includegraphics[width=0.45\linewidth]{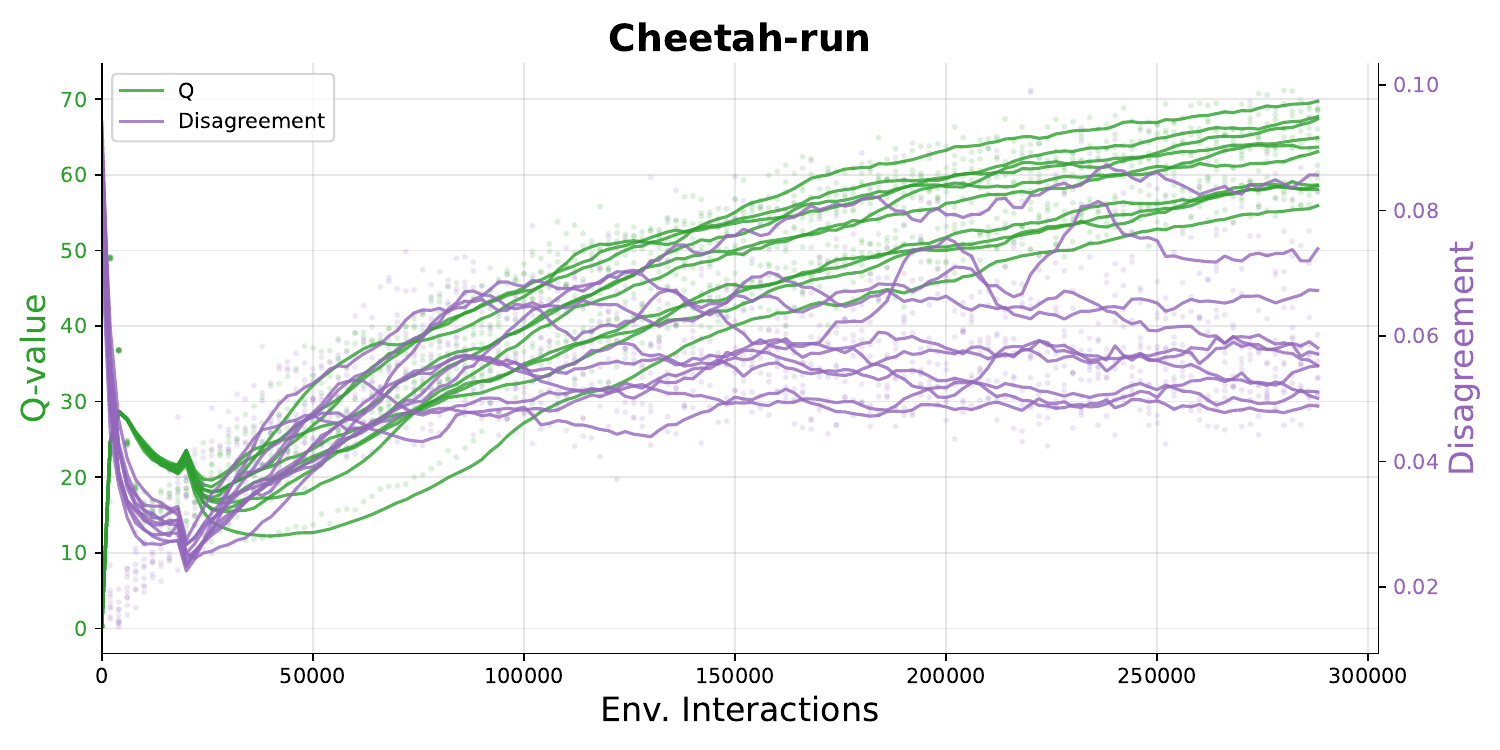}
\includegraphics[width=0.45\linewidth]{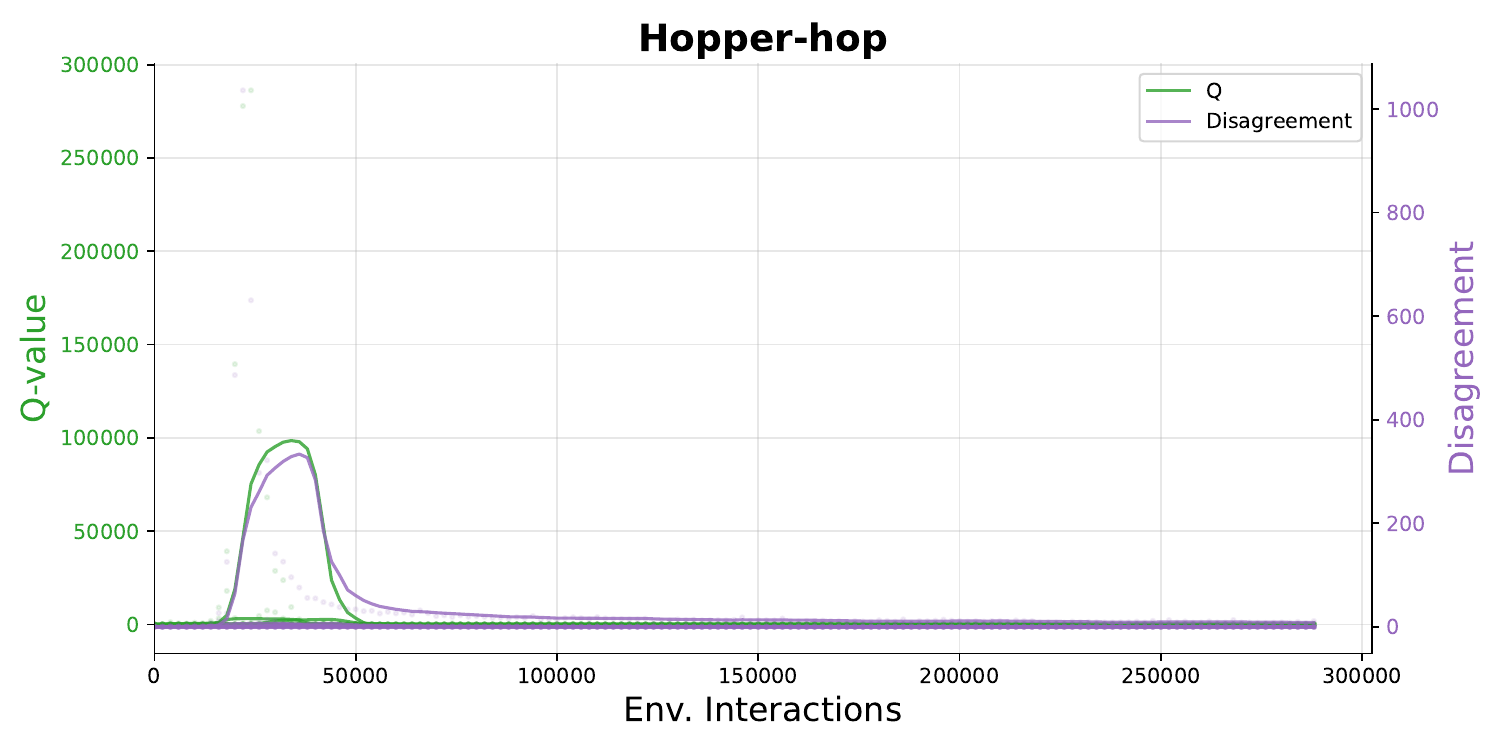}
\includegraphics[width=0.45\linewidth]{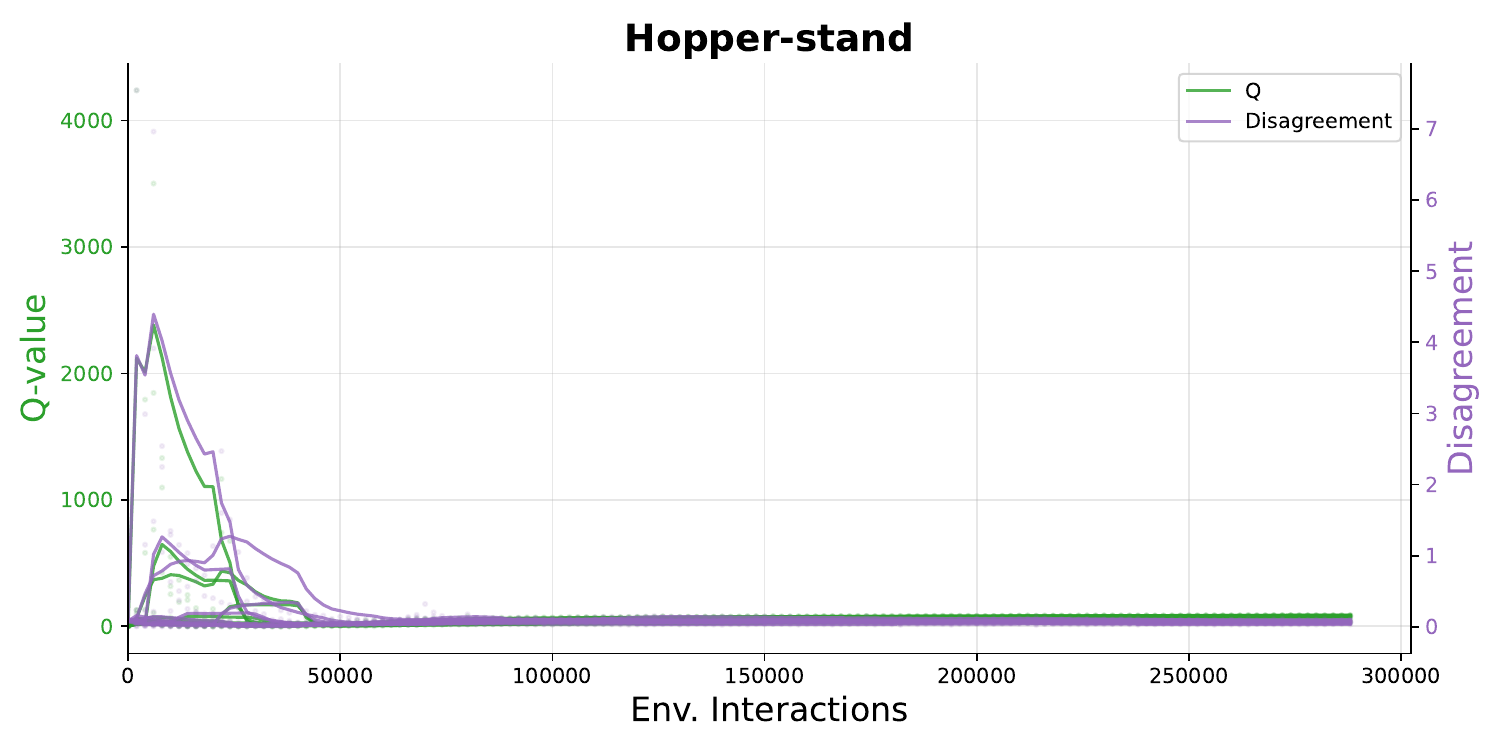}
\includegraphics[width=0.45\linewidth]{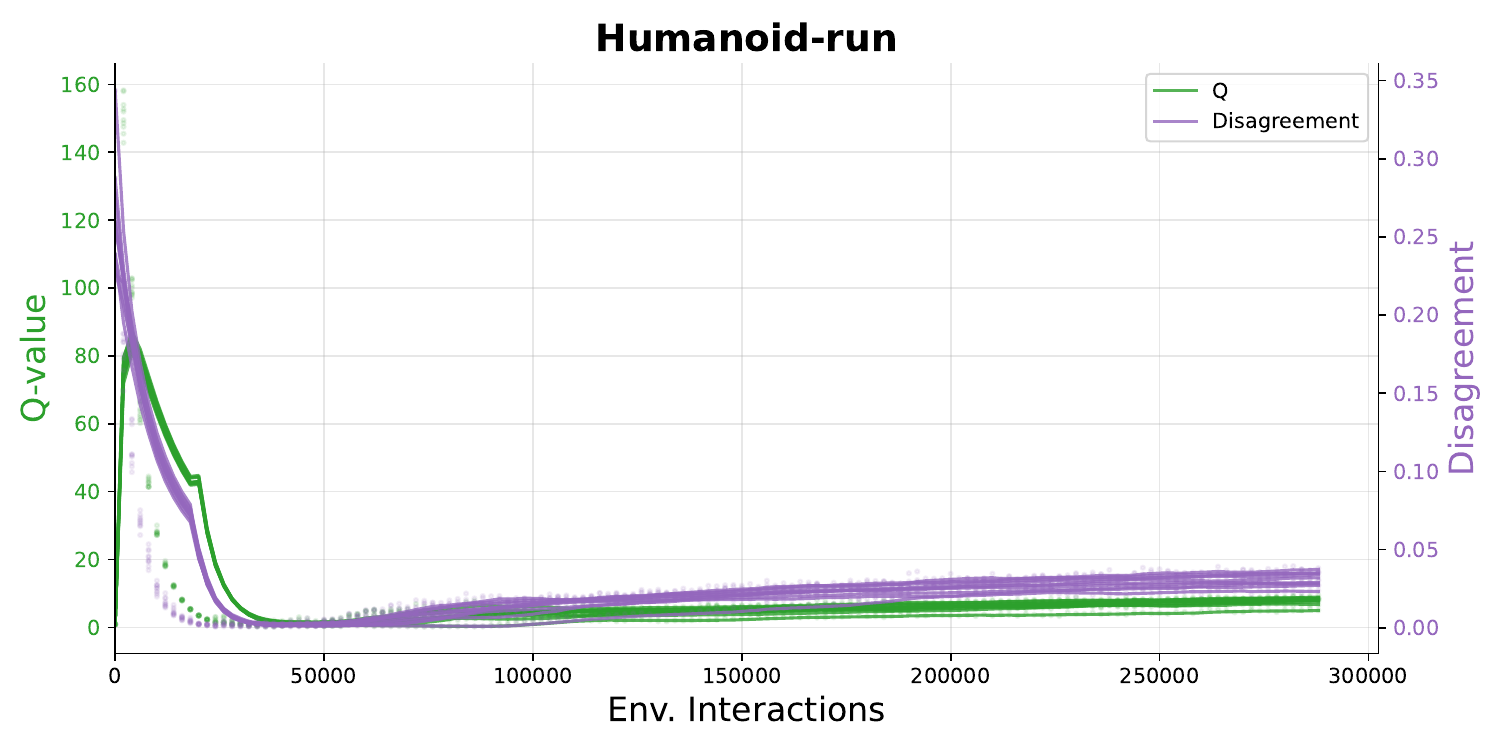}
\includegraphics[width=0.45\linewidth]{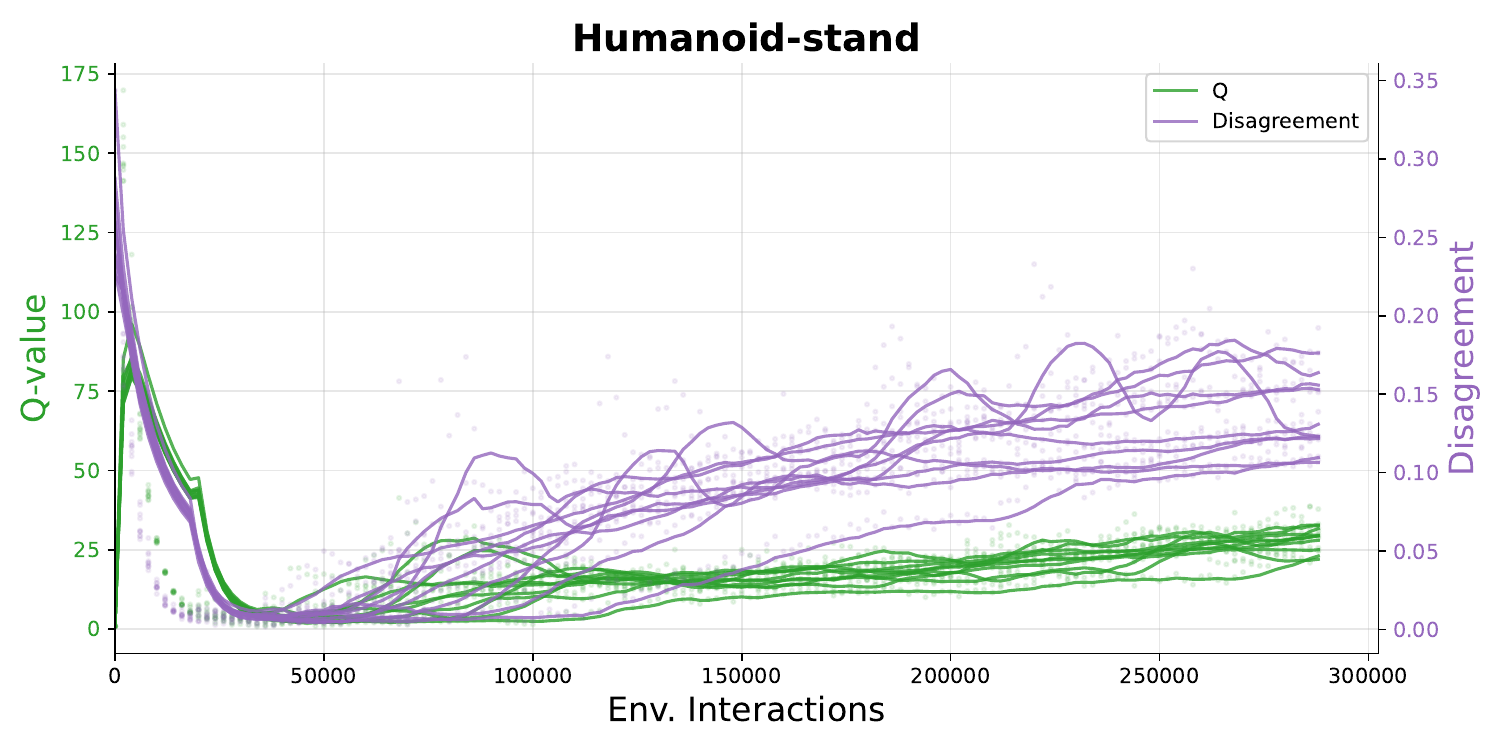}
\includegraphics[width=0.45\linewidth]{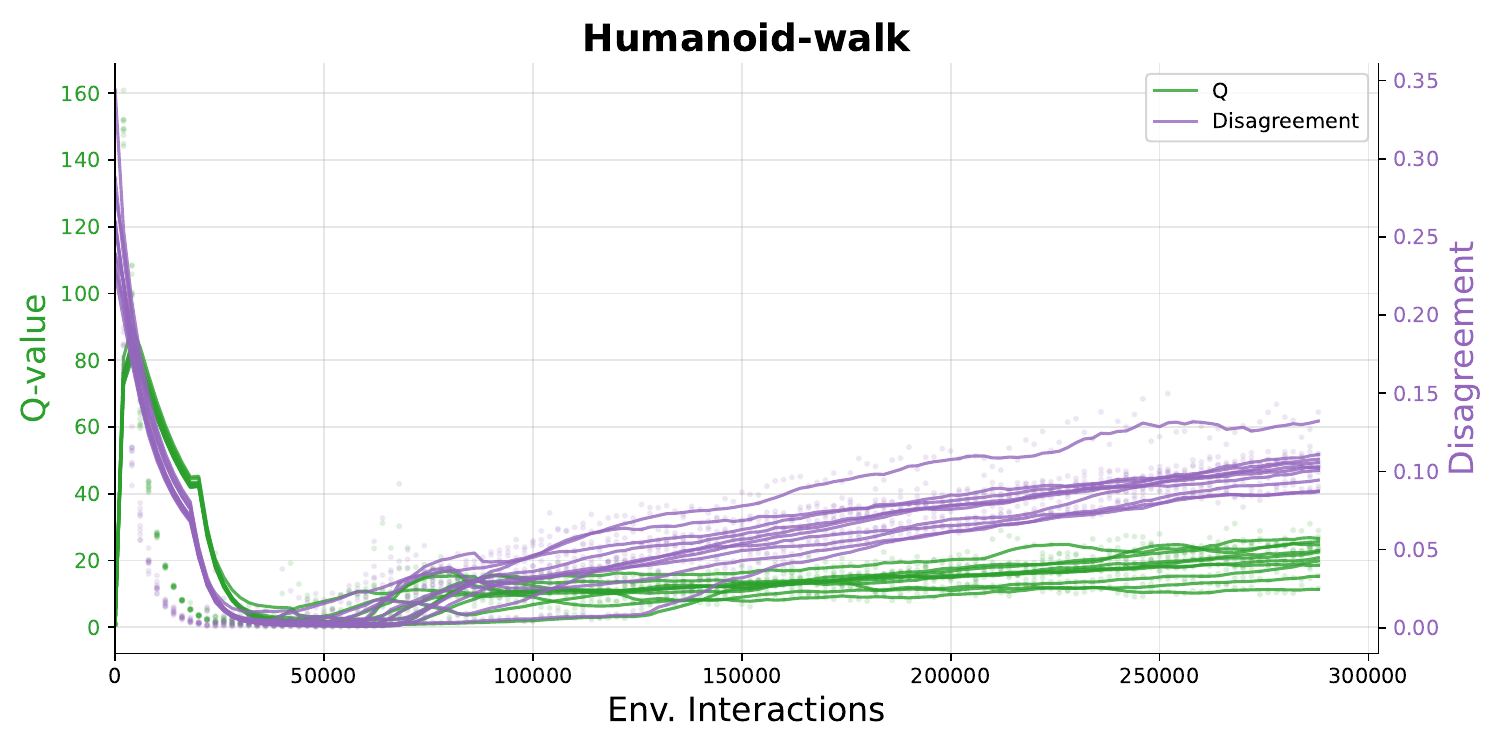}
\includegraphics[width=0.45\linewidth]{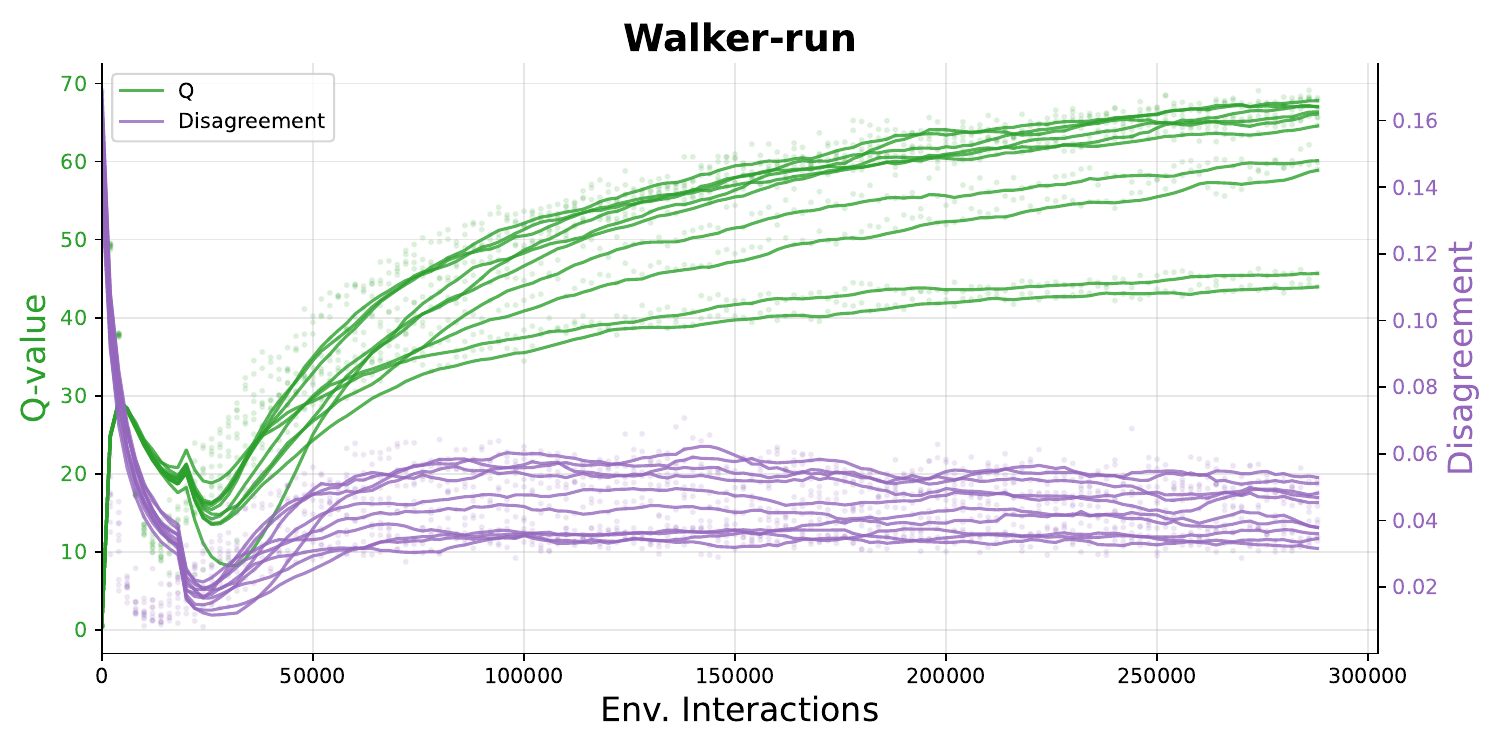}
\caption{Trajectories of $Q$-values and ensemble disagreement in DMC environments under $(N=10, G=20)$. Plotted are the raw values and moving averages per seed.}
\label{fig:dis_trajectory:sample_efficient:dmc}
\end{figure}

\paragraph{Ensemble disagreement trajectories.} 
\Cref{fig:dis_trajectory:interactive:mujoco,fig:dis_trajectory:sample_efficient:mujoco,fig:dis_trajectory:interactive:dmc,fig:dis_trajectory:sample_efficient:dmc} show the evolution of ensemble disagreement in the critic throughout training across both learning regimes. Early in training, limited data and untrained critics result in high disagreement across the ensemble. As training progresses and the critics become more aligned, disagreement generally decreases, particularly, when measured relative to the increasing scale of the $Q$-values.

\section{Ablations} \label{appendix:ablations}
We probe two practical aspects of learning decoupled reference constructions: (i) whether performance gains require the \emph{learned} (as opposed to fixed) calibrations for the critic-target and actor-policy references, and (ii) how sensitive learning is to the initialization of the critic-target calibration. The two learning regimes in the main paper already ablate the ensemble size and update-to-data ratio (UTD) and thus serve as a first check of regime transfer under substantially different training conditions.

\paragraph{Fixed (non-learned) reference constructions.}
A natural sanity check is to remove the learning of the calibration parameters and instead fix them to constants. This recovers standard special cases where the constructions of $\overline Q_{\overline \kappa}$ and $\widetilde Q_\kappa$ are pre-specified. For example, coupling both references to a conservative construction corresponds to SAC-style behavior, while using mean-like constructions corresponds to REDQ-style behavior in the large-ensemble setting. These fixed variants eliminate the possibility of \emph{decoupled adaptation} across tasks, stages, and regimes: the critic-target and actor-policy references can no longer adjust their relative conservatism or optimism based on training dynamics. Empirically, this removal of adaptivity consistently reduces performance relative to learning the calibrations (see main results), which supports the central premise that \emph{learning} how to construct $\overline Q_{\overline \kappa}$ and $\widetilde Q_\kappa$ matters beyond any single hand-designed aggregation rule, and that without learning we cannot adapt across tasks, stages, and regimes without pre-specified hyperparameters.

\paragraph{Initialization sensitivity of the critic-target calibration.}
We next study sensitivity to the initialization of the critic-target calibration, focusing on the parameter $\overline{\kappa}$. We vary the initial value of $\overline{\kappa}$ while keeping the actor-policy initialization fixed at $\kappa = 0$ (neutral with respect to conservatism or optimism). This isolates the effect of the critic-target initialization and avoids confounding interactions between the two calibrations at the start of training.

All experiments use the same protocol and hyperparameters as in \cref{sec::experiments} (see also \cref{appendix:experimental_details}) and are reported for seed 1 on MuJoCo for illustrative purposes. We observed similar results with other seeds.

Two qualitative patterns emerge. First, the effect of initialization is more pronounced in the sample-efficient regime: with high UTD, many critic updates occur before new experience arrives, so the early calibration of $\overline Q_{\overline \kappa}$ has a larger downstream impact on learning. Second, across all tested initializations, learning remains stable (no divergence or collapse). Even when the initial calibration is not well matched to a task, the learned update adjusts it during training, and many initializations achieve performance competitive with, and in several cases better than, SAC, DRND, DSAC, and REDQ under the same regime.

If initialization sensitivity is a practical concern in a particular setting, there are two simple mitigations. First, one can choose a more suitable initialization for $\overline{\kappa}$ (e.g., slightly conservative) based on a small pilot run. Second, increasing the mini-batch size used for updating $\kappa$ and $\overline{\kappa}$ (e.g., from $256$ to $512$) reduces gradient noise in these scalar updates and typically yields smoother calibration trajectories.

\begin{figure}
    \centering
    \includegraphics[width=0.48\linewidth]{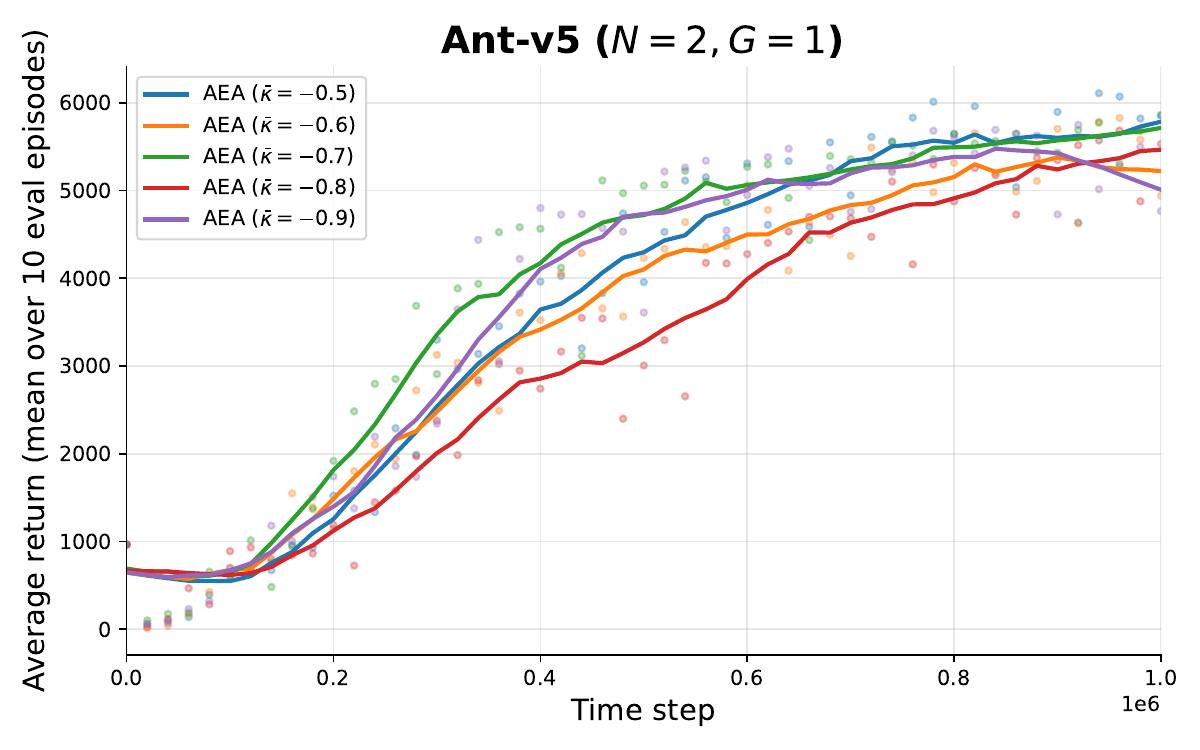}
    \includegraphics[width=0.48\linewidth]{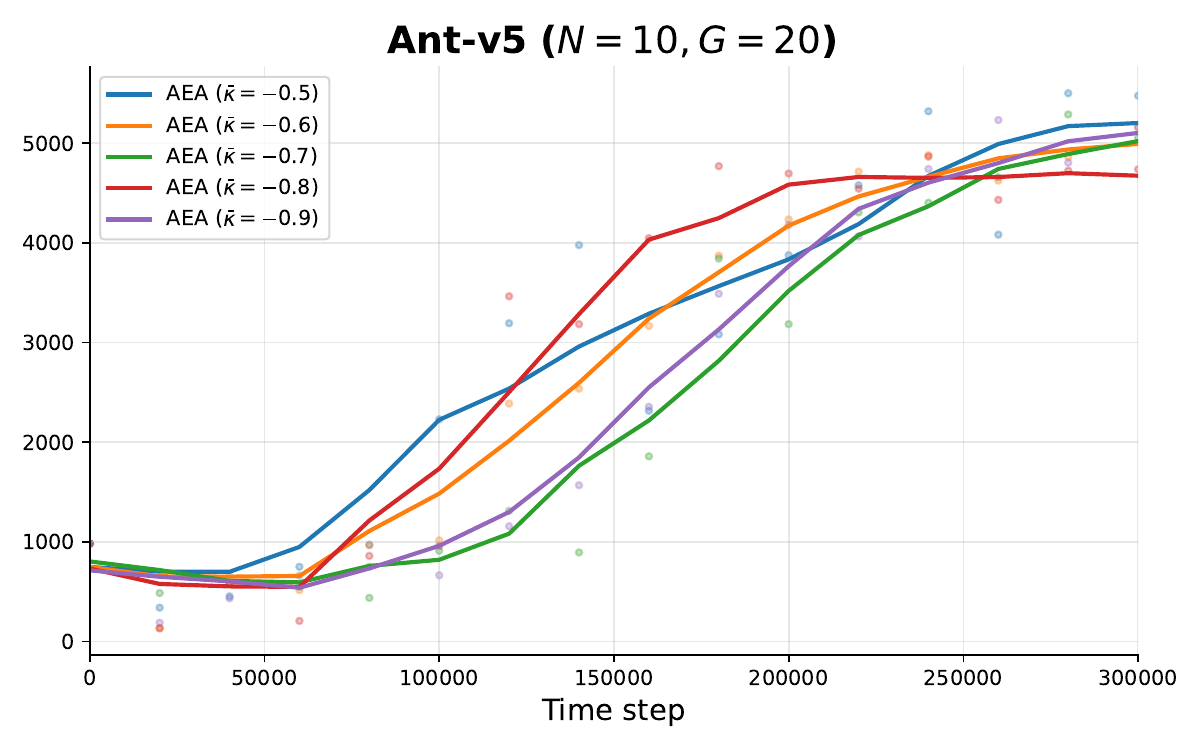}
    \includegraphics[width=0.48\linewidth]{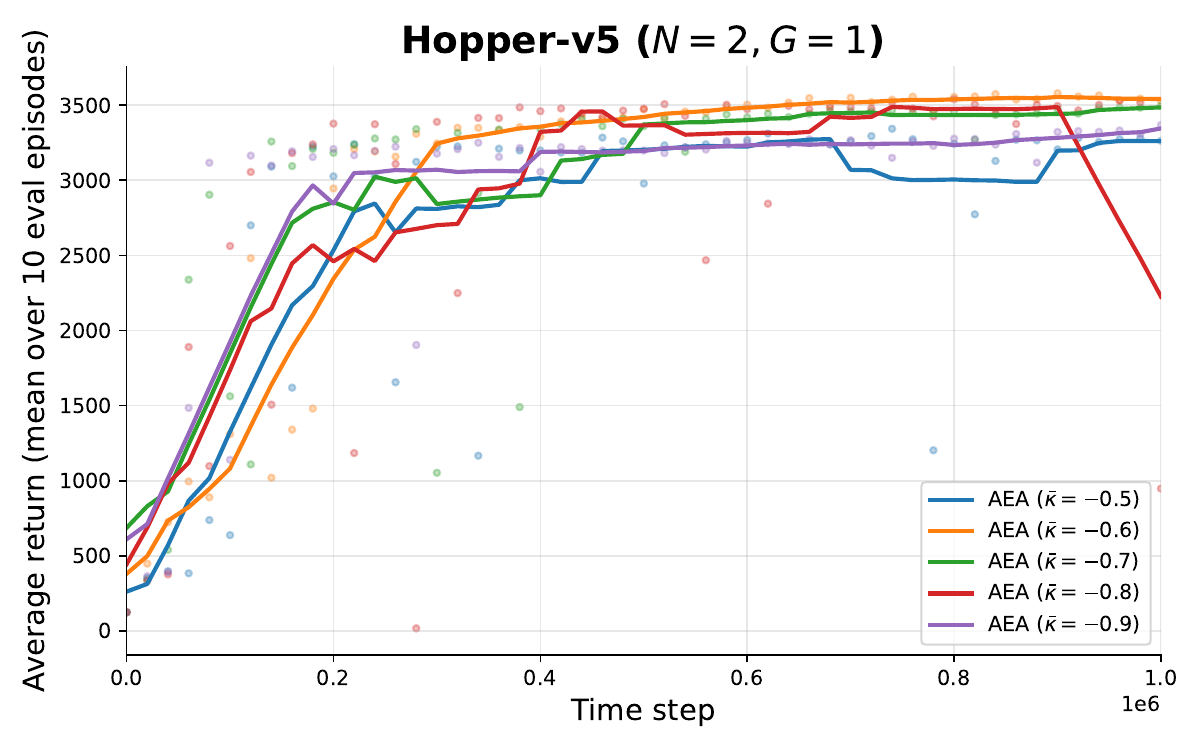}
    \includegraphics[width=0.48\linewidth]{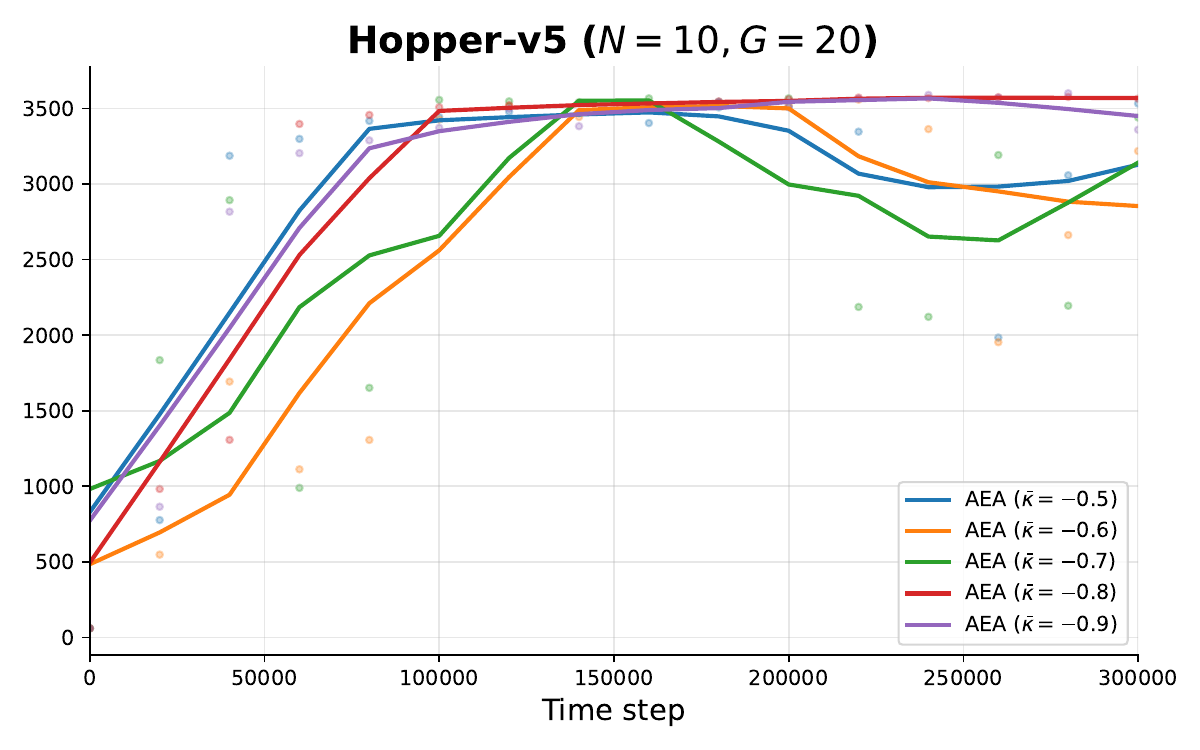}
    \includegraphics[width=0.48\linewidth]{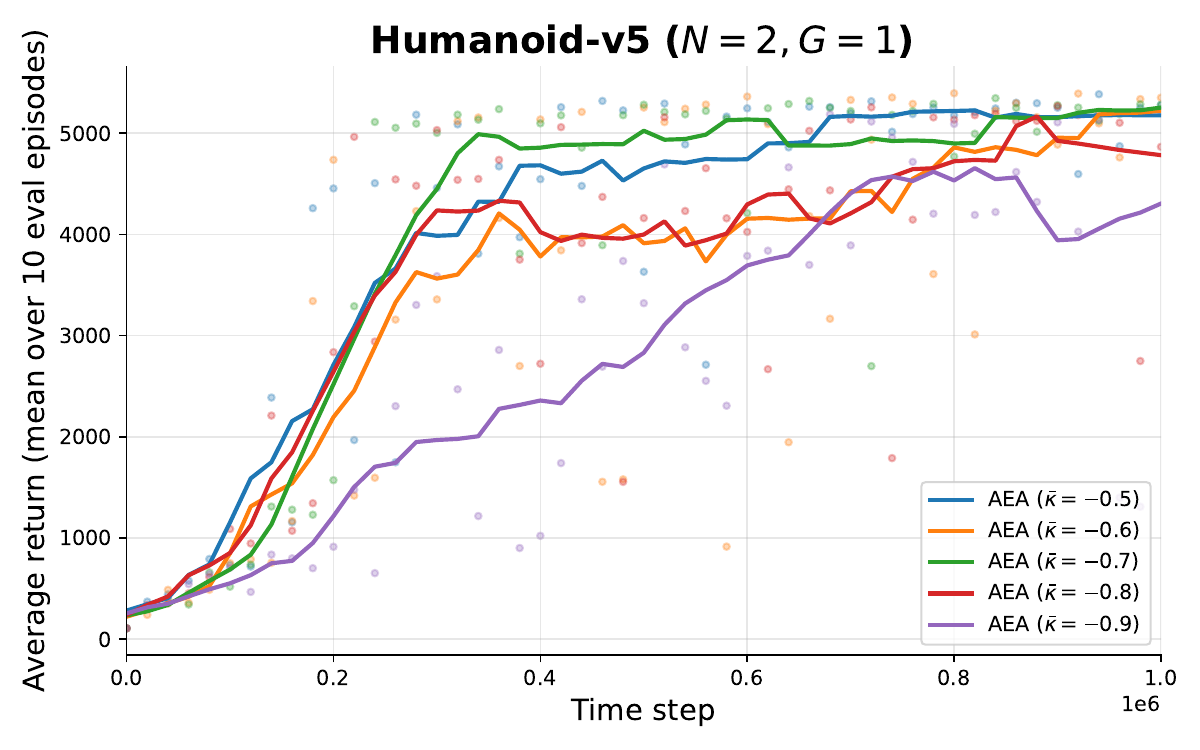}
    \includegraphics[width=0.48\linewidth]{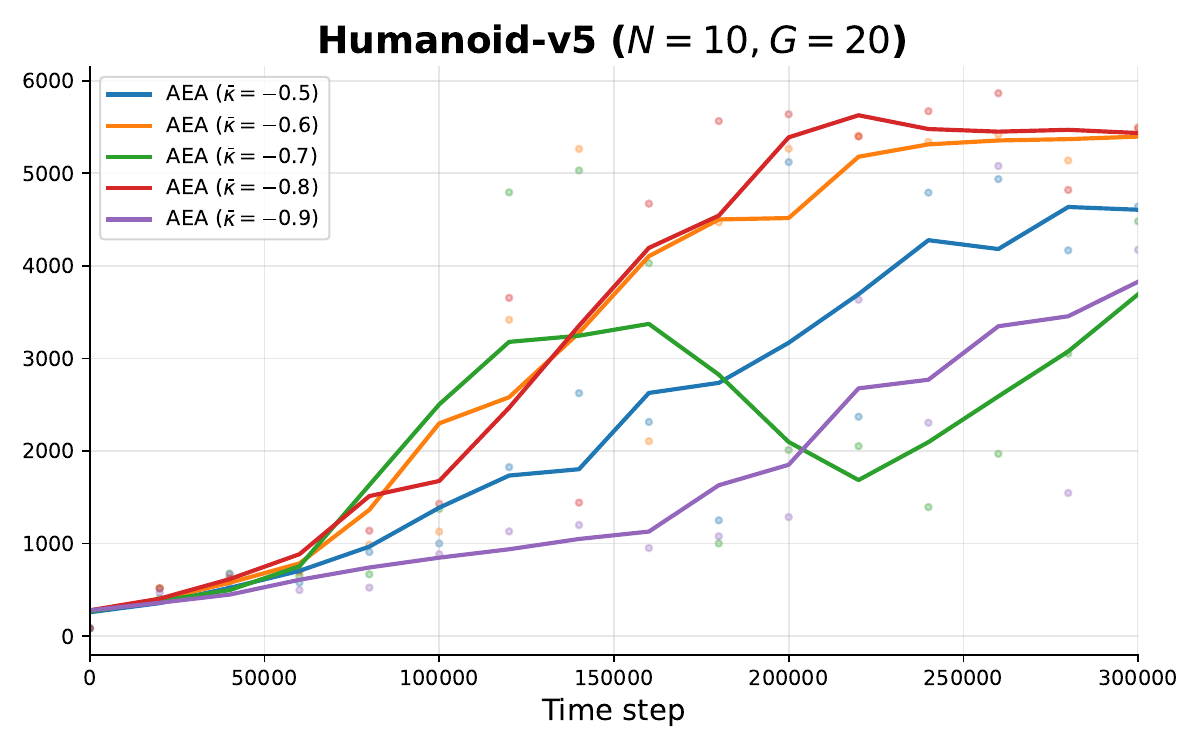}
    \includegraphics[width=0.48\linewidth]{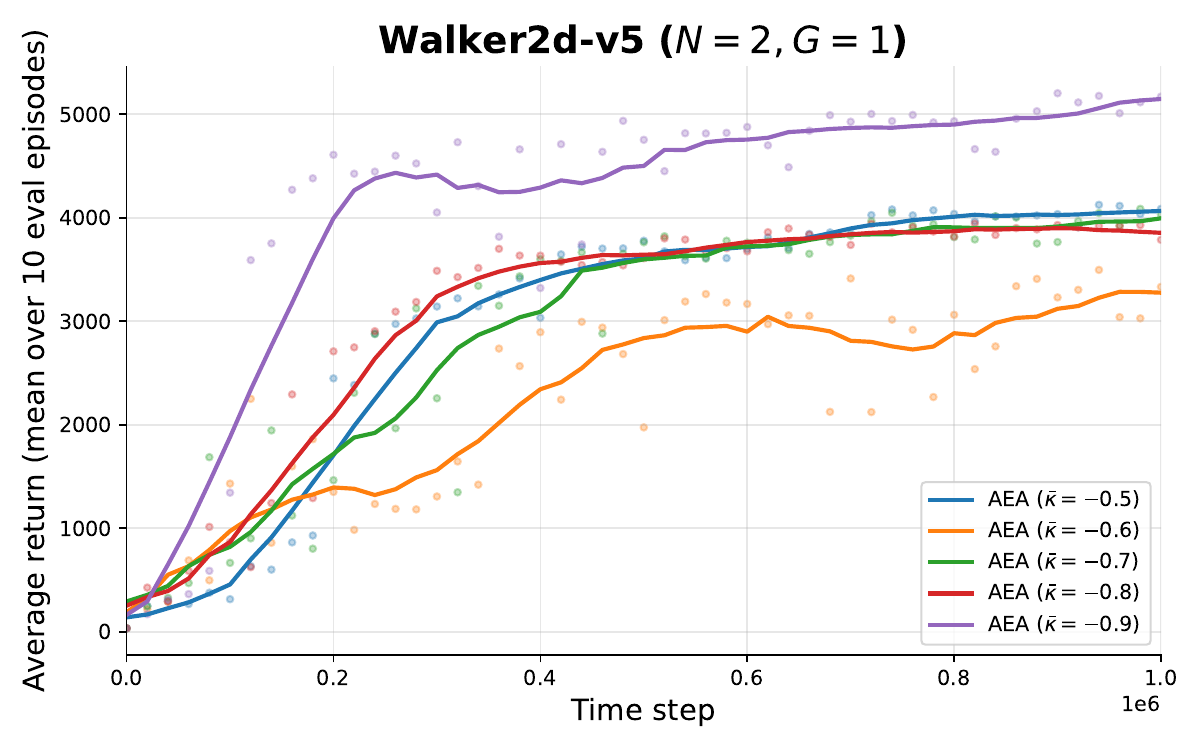}
    \includegraphics[width=0.48\linewidth]{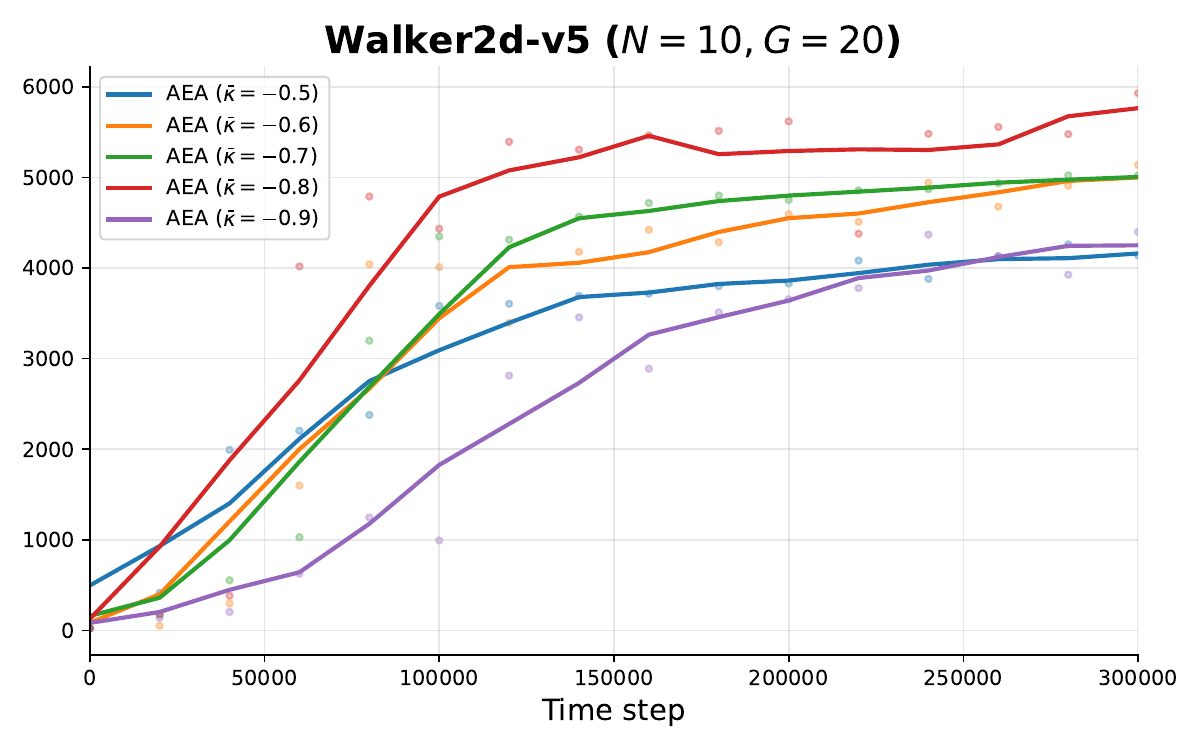}
    \caption{Effect of initializing the critic-side ensemble aggregation parameter $\overline{\kappa}$ on AEA's performance across MuJoCo environments. The left column shows results under $(N=2,G=1)$, while the right column corresponds to $(N=10, G=20)$. Each line is a moving average over the returns of a single seed.}\label{fig:ablation:init:combined}
    \label{fig:placeholder}
\end{figure}

\clearpage

\newpage
\section*{NeurIPS Paper Checklist}

\begin{enumerate}

\item {\bf Claims}
    \item[] Question: Do the main claims made in the abstract and introduction accurately reflect the paper's contributions and scope?
    \item[] Answer: \answerYes{}
    \item[] Justification: Yes, our four claims (i) convergence to an equilibrium; (ii) shrinkage; (iii) online adaptation without manual tuning; (iv) performance improvements are stated and proven/empirically validated in \cref{sec:theory,sec::experiments}.

\item {\bf Limitations}
    \item[] Question: Does the paper discuss the limitations of the work performed by the authors?
    \item[] Answer: \answerYes{}
    \item[] Justification: These are discussed at the end of \cref{sec:theory} and \cref{sec:impact}.

\item {\bf Theory assumptions and proofs}
    \item[] Question: For each theoretical result, does the paper provide the full set of assumptions and a complete (and correct) proof?
    \item[] Answer: \answerYes{}
    \item[] Justification: See \cref{sec:theory,appsec:prior_results,appsec:proofs}.

    \item {\bf Experimental result reproducibility}
    \item[] Question: Does the paper fully disclose all the information needed to reproduce the main experimental results of the paper to the extent that it affects the main claims and/or conclusions of the paper (regardless of whether the code and data are provided or not)?
    \item[] Answer: \answerYes{}
    \item[] Justification: See \cref{alg:AEA} for the pseudocode and \cref{appendix:experimental_details} for experimental details.

\item {\bf Open access to data and code}
    \item[] Question: Does the paper provide open access to the data and code, with sufficient instructions to faithfully reproduce the main experimental results, as described in supplemental material?
    \item[] Answer: \answerNo{}
    \item[] Justification: The experimental environments are based on public data, but a public implementation will only be provided for the camera-ready version of the paper.

\item {\bf Experimental setting/details}
    \item[] Question: Does the paper specify all the training and test details (e.g., data splits, hyperparameters, how they were chosen, type of optimizer) necessary to understand the results?
    \item[] Answer: \answerYes{}
    \item[] Justification: See \cref{alg:AEA} for the pseudocode and \cref{appendix:experimental_details} for all relevant details.

\item {\bf Experiment statistical significance}
    \item[] Question: Does the paper report error bars suitably and correctly defined or other appropriate information about the statistical significance of the experiments?
    \item[] Answer: \answerYes{}
    \item[] Justification: All results are over ten random seeds with standard deviations reported in each case.

\item {\bf Experiments compute resources}
    \item[] Question: For each experiment, does the paper provide sufficient information on the computer resources (type of compute workers, memory, time of execution) needed to reproduce the experiments?
    \item[] Answer: \answerYes{}
    \item[] Justification: See \cref{appendix:experimental_details}.
    
\item {\bf Code of ethics}
    \item[] Question: Does the research conducted in the paper conform, in every respect, with the NeurIPS Code of Ethics \url{https://neurips.cc/public/EthicsGuidelines}?
    \item[] Answer: \answerYes{}
    \item[] Justification: The paper confirms to to the code of ethics.

\item {\bf Broader impacts}
    \item[] Question: Does the paper discuss both potential positive societal impacts and negative societal impacts of the work performed?
    \item[] Answer: \answerYes{} %
    \item[] Justification: See \cref{sec:impact}.
    
\item {\bf Safeguards}
    \item[] Question: Does the paper describe safeguards that have been put in place for responsible release of data or models that have a high risk for misuse (e.g., pre-trained language models, image generators, or scraped datasets)?
    \item[] Answer: \answerNA{} %
    \item[] Justification: \answerNA{}

\item {\bf Licenses for existing assets}
    \item[] Question: Are the creators or original owners of assets (e.g., code, data, models), used in the paper, properly credited and are the license and terms of use explicitly mentioned and properly respected?
    \item[] Answer: \answerYes{} %
    \item[] Justification: All existing libraries are cited whenever they are first mentioned.

\item {\bf New assets}
    \item[] Question: Are new assets introduced in the paper well documented and is the documentation provided alongside the assets?
    \item[] Answer: \answerNA{} %
    \item[] Justification: \answerNA{}

\item {\bf Crowdsourcing and research with human subjects}
    \item[] Question: For crowdsourcing experiments and research with human subjects, does the paper include the full text of instructions given to participants and screenshots, if applicable, as well as details about compensation (if any)? 
    \item[] Answer: \answerNA{}
    \item[] Justification: \answerNA{}

\item {\bf Institutional review board (IRB) approvals or equivalent for research with human subjects}
    \item[] Question: Does the paper describe potential risks incurred by study participants, whether such risks were disclosed to the subjects, and whether Institutional Review Board (IRB) approvals (or an equivalent approval/review based on the requirements of your country or institution) were obtained?
    \item[] Answer: \answerNA{}
    \item[] Justification: \answerNA{}

\item {\bf Declaration of LLM usage}
    \item[] Question: Does the paper describe the usage of LLMs if it is an important, original, or non-standard component of the core methods in this research? Note that if the LLM is used only for writing, editing, or formatting purposes and does \emph{not} impact the core methodology, scientific rigor, or originality of the research, declaration is not required.
    \item[] Answer: \answerNA{}
    \item[] Justification: \answerNA{}

\end{enumerate}

\end{document}